\documentclass{article}
\usepackage[T1]{fontenc}
\usepackage[utf8]{inputenc}
\usepackage{preprint}
\usepackage{amsmath,amssymb}
\usepackage{booktabs}
\usepackage{multirow}
\usepackage{graphicx}
\usepackage[dvipsnames]{xcolor}
\usepackage[font=footnotesize,skip=3pt,subrefformat=parens]{subcaption}
\usepackage[format=plain,labelformat=simple,labelsep=period,font=small,compatibility=false]{caption}
\usepackage[shortlabels,inline]{enumitem}
\setlist[itemize]{noitemsep,leftmargin=*,topsep=0em}
\setlist[enumerate]{noitemsep,leftmargin=*,topsep=0em}
\usepackage{microtype}
\usepackage{soul}
\usepackage{xspace}
\usepackage{fontawesome5}
\usepackage[numbers,sort&compress]{natbib}
\usepackage[hyphens]{url}
\definecolor{linkblue}{rgb}{0.21,0.49,0.74}
\usepackage[breaklinks,colorlinks,allcolors=linkblue]{hyperref}
\usepackage[capitalize]{cleveref}
\crefname{section}{Sec.}{Secs.}
\Crefname{section}{Section}{Sections}
\crefname{table}{Tab.}{Tabs.}
\Crefname{table}{Table}{Tables}

\makeatletter
\DeclareRobustCommand\onedot{\futurelet\@let@token\@onedot}
\def\@onedot{\ifx\@let@token.\else.\null\fi\xspace}

\makeatother

\usepackage{wrapfig}
\makeatletter  
\newcommand{\needlines}[1]{\par\begingroup\dimen@=#1\baselineskip\dimen@ii=\pagegoal\advance\dimen@ii-\pagetotal
  \ifdim\dimen@>\dimen@ii\ifdim\dimen@ii>\z@\vfil\fi\break\fi\endgroup}
\makeatother
\usepackage{tcolorbox}
\definecolor{keyqback}{rgb}{0.95,0.95,1}
\definecolor{keyqframe}{rgb}{0,0,0.75}
\newtcolorbox{keyq}{colback=keyqback,colframe=keyqframe,boxrule=0.5pt,arc=2pt,
  left=6pt,right=6pt,top=3pt,bottom=3pt,before skip=6pt,after skip=6pt,fontupper=\itshape}

\newcommand{\nope}{\textsc{NoClock}}
\newcommand{\clk}{\textsc{Clock}}

\title{Is Your Robot's Memory Just a Clock?\\Temporal Shortcuts in Memory-Augmented VLA Policies}

\author{%
\begin{minipage}{0.96\textwidth}\centering\normalfont\normalsize
{\bfseries Chubin Zhang \qquad Bo An$^{\dagger}$}\\[0.3em]
Nanyang Technological University\\[0.4em]
{\small \faEnvelope\ \href{mailto:chubinzhang23@gmail.com}{\nolinkurl{chubinzhang23@gmail.com}}}\\[0.1em]
{\small \faGithub\ \url{https://github.com/bennidict23/memory-clock}}
\end{minipage}%
}

\begin{document}
\raggedbottom
\maketitle
\begingroup
\renewcommand{\thefootnote}{\ensuremath{\dagger}}
\footnotetext[1]{Corresponding author.}
\endgroup
\begin{center}
\begin{minipage}{\linewidth}
\centering
\includegraphics[width=0.86\linewidth]{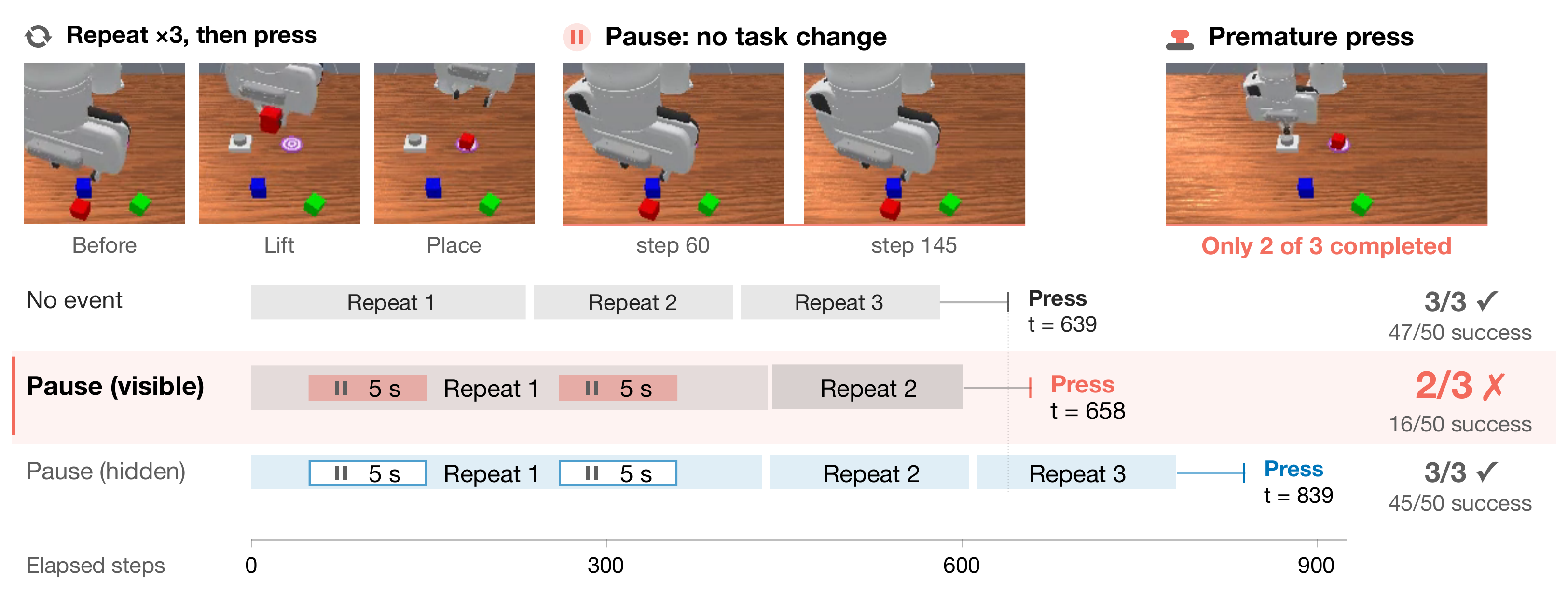}
\captionof{figure}{\textbf{A pause that changes nothing is counted as progress.}
One PickXtimes episode of a released memory VLA (``pick up the cube and place it on the target 3 times, then press the button''), run three times along a shared axis of environment steps.
Two 5\,s pauses leave the world unchanged (top middle: first and last frame of the first pause).
Yet when the pause frames enter memory, the policy presses the button after only 2 of 3 repetitions, at almost the same step as without pauses (dotted line), as if it had counted elapsed time rather than repetitions.
We call this failure \emph{phantom progress}.
When the same pauses are hidden from memory (outlined boxes), the policy completes all 3.
Right: successes over 50 paired episodes.
Repetition boundaries are read from video ($\pm$10 steps), whereas press times are exact.}
\label{fig:teaser}
\end{minipage}
\end{center}
\vspace{0.6em}

\begin{abstract}
Memory-augmented vision-language-action (VLA) policies are meant to remember what happened.
Many of them, however, fill memory by time: they sample frames uniformly across the episode and stamp each frame with its time step.
Because time passes even when nothing happens, such a memory may not tell waiting from working.
We show that it does not.
On the counting tasks of RoboMME, we make the robot briefly pause or fidget its wrist.
Such a \textbf{null event} changes nothing the task depends on, yet a released memory VLA counts the interval as work done and stops early.
We call this failure \textbf{phantom progress}.
It cuts success from 47 to 16 of 50 on one task and from 44 to 10 on the other.
When the same frames are withheld from memory, success returns, so the cause lies in memory rather than in the interruption itself.
We trace the failure to a \textbf{temporal shortcut}: because demonstrations proceed at a steady pace, elapsed time predicts progress.
Indeed, rescaling the time stamps alone makes the policy stop early when the clock runs fast and repeat when it runs slow.
Surprisingly, removing the time encoding does not cure the failure, because the frames of a null event still occupy slots of the time-sampled memory.
Across RoboMME's released designs, vulnerability follows how memory is indexed.
Of the fixes the diagnosis suggests, training with injected pauses does not remove the failure and impairs counting, whereas indexing memory by events recovers half or more of the loss without retraining.
A memory meant to remember what happened should count events rather than clock ticks.
\end{abstract}

\section{Introduction}
\label{sec:intro}

Many manipulation tasks cannot be solved from the current observation alone.
A robot asked to pick up a cube three times and then press a button sees much the same scene before its second pick as before its third, so it has to remember how many picks it has made.
Memory-augmented vision-language-action (VLA) policies address this by conditioning each action on a memory of the episode so far~\cite{shi2026memoryvla,sridhar2026memer,torne2026mem,dai2026robomme}.
A simple and widely used way to fill such a memory is by time: the policy samples past frames uniformly over the episode and stamps each frame with its time step.
This design covers the whole episode at a fixed cost, and on RoboMME~\cite{dai2026robomme}, a benchmark built to test memory, the strongest released policy uses exactly this design.

However, time passes even when nothing happens.
Real robots often wait for a slow planner, for an operator who has paused them, or for a safety monitor to clear.
They also hesitate, or reach for an object and come back empty-handed.
We call such an interval a \textbf{null event}.
It takes time, and the robot may even move, yet nothing the task depends on changes.
A memory that records what happened should therefore pass over a null event as if it had not occurred.
A memory filled by time cannot do so, because the interval receives its share of sampled frames and time stamps like any other stretch of the episode.
This raises a simple question:
\begin{keyq}
Can a memory filled by time tell waiting from working?
\end{keyq}

We find that it cannot (\cref{fig:teaser}).
On the two counting tasks of RoboMME, we make the released policy pause twice for 5 seconds, or roll its wrist back and forth for as long.
Success falls from 47 to 16 of 50 on PickXtimes and from 44 to 10 on SwingXtimes.
Almost every failure is an early stop, in which the policy presses the button before it has made the required number of repetitions.
The robot thus perceives progress that never happened.
We call this failure \textbf{phantom progress}.
One might suspect that the interruption itself disturbs the robot, for instance by leaving its arm in an unusual pose.
We therefore repeat each event with its frames withheld from memory.
The robot pauses or fidgets exactly as before, but its memory never sees the interval, and success returns to within three episodes of the baseline.
The failure therefore lies in what the policy remembers.

Why would a policy mistake waiting for working?
We trace the failure to a \textbf{temporal shortcut}.
Demonstrations proceed at a steady pace, so elapsed time alone predicts how many repetitions are done 89\% and 80\% of the time on the two tasks, and over 92\% on regular demonstrations from two real teleoperation datasets.
Since demonstrations contain almost no idle stretches, imitation learning never penalises a policy for reading progress from time.
If the policy indeed reads progress from time, rescaling the time stamps alone should shift its failures in a predictable direction.
It does: with no null event and the frames unchanged, a faster clock makes the policy stop early, and a slower one makes it repeat (\cref{sec:clock}).
The obvious remedy is then to remove the time encoding.
Surprisingly, a policy trained from scratch without it still fails under pauses (21 and 14 of 50), because the frames of a null event still occupy slots of a memory sampled by time (\cref{sec:noclock}).
\textbf{The shortcut thus lies in how memory is indexed, and the time stamps are only one of its carriers.}
Consistently, among RoboMME's released designs, every memory indexed by time collapses under null events, a memory that selects tokens by pixel change is hurt less by pauses than by fidgets, and memories indexed by sub-goals withstand both null events on PickXtimes (\cref{sec:designs}).

The diagnosis also suggests where to fix the failure.
Training with injected pauses changes the data but keeps the index, so at best it can teach the policy to discount the null events that we know how to synthesise.
It does not even achieve this: pause augmentation does not remove the failure, and it impairs counting when no null event occurs.
Changing the index should work instead.
Indexing memory by events collapses every stretch in which the world returns to a state it was already in.
Applied at test time without retraining, it recovers half or more of the lost success under pauses, fidgets and air-grasps alike (\cref{sec:fixes}).

\noindent We summarize our contributions as follows:
\begin{itemize}
\item \textbf{Phantom progress.} Brief null events make memory VLAs that fill memory by time stop early, and the damage passes through memory alone. Comparing visible with hidden null events, together with rescaling the time stamps, gives a protocol for testing whether any memory policy tracks events or elapsed time (\cref{sec:phenomenon}).
\item \textbf{A shortcut in the index.} We trace the failure to a temporal shortcut that lives in how memory is indexed, so that removing the time stamps does not remove it, and we show that vulnerability follows the index across RoboMME's released designs (\cref{sec:mechanism}).
\item \textbf{Fixes suggested by the diagnosis.} Pause augmentation does not remove the failure, whereas indexing memory by events recovers much of the loss without retraining (\cref{sec:fixes}).
\end{itemize}

\needlines{6}
\section{Related Work}
\label{sec:related}

\paragraph{Memory in robot policies.}
Most VLA policies act on the current observation or a short window~\cite{black2025pi0,black2025pi05,zhao2023act,chi2023diffusion}.
A growing line adds memory: history-aware transformers~\cite{guhur2022hiveformer,li2024roboflamingo,wu2024gr1,koo2026hamlet}, perceptual memory banks~\cite{shi2026memoryvla,fang2025sam2act}, keyframe and event selection~\cite{sridhar2026memer,mark2026bpp,zeng2026kemo,yang2026eventvla}, recurrent and test-time-training memories~\cite{bulatov2022rmt,sun2025ttt,liu2026t2mem}, and mixed video--language memory~\cite{torne2026mem}.
RoboMME~\cite{dai2026robomme} standardises their evaluation and releases the MME-VLA family we study, and MIKASA-Robo~\cite{cherepanov2025mikasa} and MemoryBench~\cite{fang2025sam2act} are related suites.
Such benchmarks ask which memory scores highest.
A smaller set of works asks what memory actually does.
History can be encoded yet barely used~\cite{liao2026present}, and long or uncurated histories invite spurious correlations~\cite{mark2026bpp,wu2026imprint,torne2025pasttoken}.
These works treat the problem as a shift in the distribution of histories and remedy it by choosing which frames to keep.
We instead identify one specific correlate, elapsed time, show by intervention that the policy reads it, and find that selecting memory content by appearance does not remove it.
Closest to us, AGM~\cite{gao2026agm} shows that memories logging attempted actions as progress turn a failed grasp into a persistent error, and RoboMME-Interference~\cite{rathi2026interference} shows that unrelated sessions crowd frame-sampled memories.
Concurrent event-driven keyframe memories~\cite{zeng2026kemo,yang2026eventvla} report that selecting frames by events beats uniform sampling.
Our analysis explains why such memories should resist null events, and our event index (\cref{sec:fixes}) is a minimal intervention that tests this account rather than a new architecture.

\paragraph{Shortcuts, time, and position.}
Shortcut learning~\cite{geirhos2020shortcut} is well documented in imitation learning, as causal confusion~\cite{dehaan2019causal} and copycat behaviour of history-conditioned policies~\cite{wen2020copycat,chuang2022residual}.
Elapsed time is a special case: it is legitimately part of the state under time limits~\cite{pardo2018time}, many controllers are driven by a phase variable~\cite{ijspeert2013dmp,peng2018deepmimic}, and progress estimators are tempted by temporal position, which is why GVL shuffles frames~\cite{ma2025gvl}.
In sequence models, positional encodings shape counting and length generalisation~\cite{press2022alibi,anil2022length,kazemnejad2023nope,ruoss2023randomized}, and video moment-retrieval models can answer from the temporal position of a moment rather than the video content~\cite{otani2020uncovering,yuan2021closer}.
We show that in memory VLAs the time position becomes a substitute for content, and that removing the explicit encoding does not remove it.

\paragraph{Events and redundancy in video memory.}
Video models compress long inputs by merging redundant frames or tokens~\cite{bolya2023tome,song2024moviechat,he2024malmm}, and event segmentation~\cite{zacks2007event,shou2021gebd,fountas2025emllm} groups frames into meaningful units.
Such methods judge redundancy by appearance.
A null event need not look redundant: a fidgeting wrist changes every frame.
Our event-indexed memory instead uses whether the world returns to a previous state, in the spirit of loop closure in SLAM~\cite{cadena2016slam}.

\paragraph{Robustness of robot policies.}
Robustness benchmarks perturb what the robot sees or is told: objects, textures, lighting, camera, language, and initial state~\cite{pumacay2024colosseum,fei2025liberoplus,zhou2025liberopro,wang2025vlatest}.
Timing has been studied as inference latency~\cite{black2025rtc,liu2025bid}.
To our knowledge, no prior work injects test-time null events into the rollouts of memory-augmented policies.

\needlines{6}
\section{Setup}
\label{sec:setup}

\paragraph{Tasks and policies.}
RoboMME~\cite{dai2026robomme} provides memory-dependent manipulation tasks in simulation.
We focus on its two counting tasks, where memory is needed to decide when the task is done: in \emph{PickXtimes} the robot picks and places a cube $K\in\{1,\dots,5\}$ times and then presses a button; in \emph{SwingXtimes} it carries a cube back and forth between two targets $K$ times.
We study the released MME-VLA policies~\cite{dai2026robomme}, built on $\pi_{0.5}$~\cite{black2025pi05}.
Their memory, FrameSamp, keeps 32 frames sampled \emph{uniformly in time} from the whole episode and stamps each with a sinusoidal encoding of its absolute time step.
The three variants read this memory through cross-attention in the action expert (``Modul''), through a separate memory expert (``Expert''), or in context (``Context'').
TokenDrop instead keeps tokens selected by pixel change, with the same time encoding.
For comparison across memory designs (\cref{sec:designs}) we also evaluate the released MemER-style keyframe memory~\cite{sridhar2026memer}, a text sub-goal memory, a recurrent test-time-training memory, and $\pi_{0.5}$ without memory.
Unless stated otherwise, results use FrameSamp-Modul, the strongest released policy.

\paragraph{Models trained from scratch.}
To test whether the failure needs the time encoding, we train two models that differ only in it.
Both start from the $\pi_{0.5}$ base checkpoint and follow the released recipe on the RoboMME training data for 30k steps, but \clk{} keeps the time encoding while \nope{} removes it.
Without null events both come close to the released model (\cref{supp:noclock}).

\paragraph{Null events.}
At action-chunk boundaries after policy steps 48 and 160 we inject one of three events that leave unchanged what the task depends on (\cref{fig:phenomenon}a):
a \emph{pause}, in which the robot holds still for 100 environment steps (5\,s; 30 and 300 steps in dose studies);
a \emph{wrist fidget}, in which the wrist rolls $\pm0.3$\,rad around its held target for 100 steps;
and an \emph{air-grasp}, in which the motion planner executes a grasp above a resting object and returns to the pre-event pose.
A fidget can nudge a held cube, but the damage persists in episodes where it moved nothing (\cref{supp:validity}).

\begin{wraptable}{r}{0.56\linewidth}
\centering
\small
\setlength{\tabcolsep}{4pt}
\begin{tabular}{llccccc}
\toprule
Task & Condition & $K{=}1$ & 2 & 3 & 4 & 5 \\
\midrule
\multirow{4}{*}{PickX.} & No event & 16/16 & 10/10 & 11/12 & 6/6 & 4/6 \\
 & Pause & 16/16 & 0/10 & 0/12 & 0/6 & 0/6 \\
 & Fidget & 16/16 & 0/10 & 0/12 & 0/6 & 0/6 \\
 & Pause, hidden & 16/16 & 10/10 & 11/12 & 5/6 & 3/6 \\
\midrule
\multirow{4}{*}{SwingX.} & No event & 14/15 & 10/12 & 20/23 & -- & -- \\
 & Pause & 10/15 & 0/12 & 0/23 & -- & -- \\
 & Fidget & 10/15 & 0/12 & 0/23 & -- & -- \\
 & Pause, hidden & 13/15 & 12/12 & 22/23 & -- & -- \\
\bottomrule
\end{tabular}
\caption{\textbf{Null events break the policy exactly where it must decide when to stop.}
Successes / episodes by the number of repetitions $K$ ($K{\le}3$ in SwingXtimes), released FrameSamp-Modul, two 5-second events (the episodes of \cref{fig:phenomenon}).}
\label{tab:byK}
\end{wraptable}
\paragraph{Visible versus hidden.}
In the \emph{visible} condition, null frames enter memory like any other frame.
In the \emph{hidden} condition, the same event happens physically, but its frames are withheld from memory and do not advance the time stamps.
The hidden condition is the reference for a memory that is indifferent to null events.

\paragraph{Protocol.}
Each cell is 50 test episodes. All comparisons pair the same episodes (environment seeds and policy noise) and use an exact two-sided McNemar test. Evaluation is not exactly deterministic: two runs of one condition differed by 8 of 50 (\cref{supp:clock}), so we read differences under about 10 episodes with care.
We classify failures as \emph{early} (the policy pressed the button or submitted before completing $K$ repetitions) or \emph{extra} (it performed more than $K$). Per-cell intervals, Holm-corrected tests and results by $K$ are in \cref{supp:cells}.

\needlines{6}
\section{Null Events Break Memory}
\label{sec:phenomenon}

\begin{wrapfigure}{r}{0.48\linewidth}
\centering
\includegraphics[width=\linewidth]{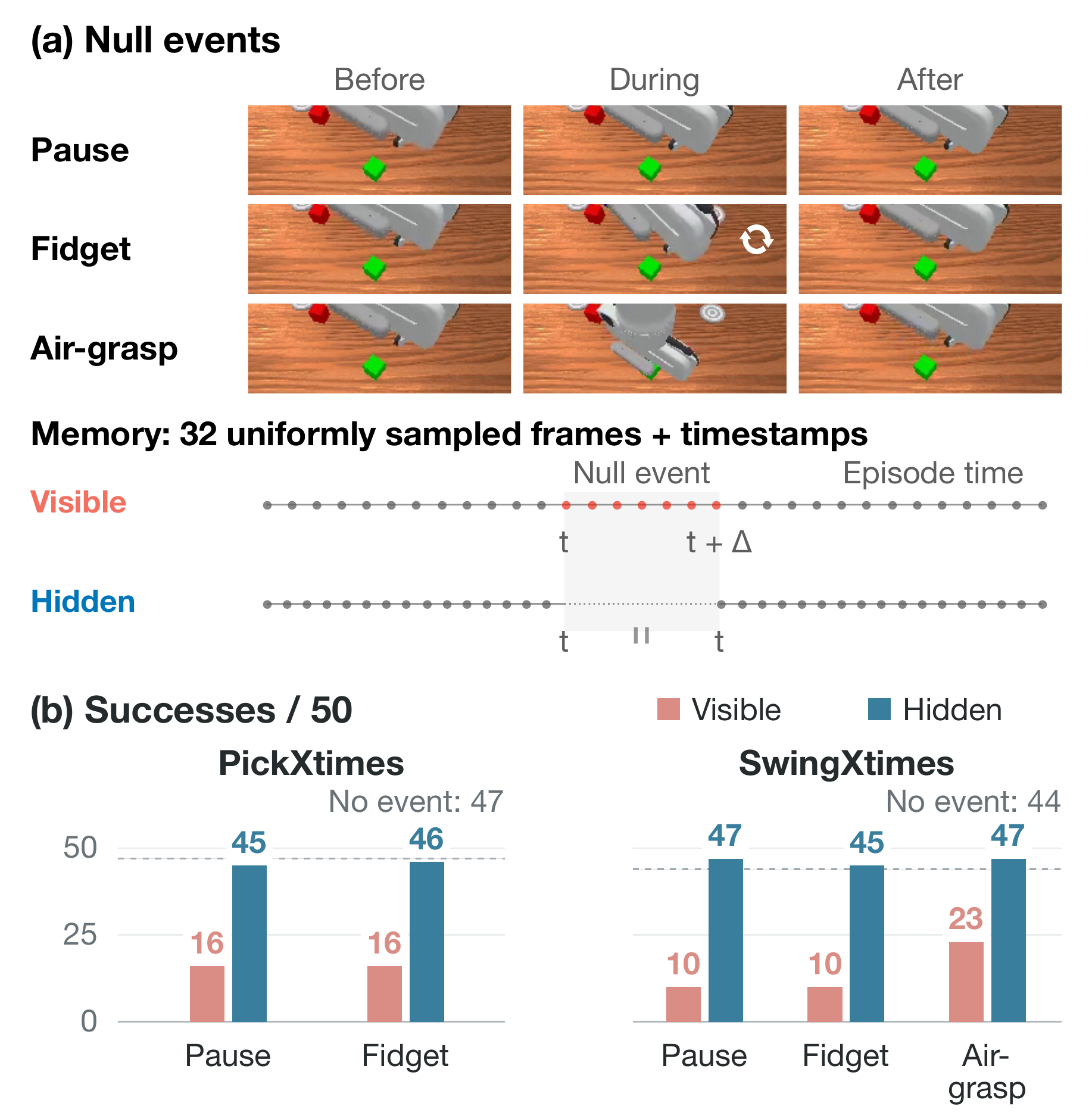}
\caption{\textbf{Null events break memory, and only through memory.}
(a) Three null events in one SwingXtimes episode (front camera; first event of three separate rollouts): the robot pauses, rolls its wrist, or grasps the air above the cube and returns, and no object moves.
Memory keeps 32 frames sampled uniformly over the episode, each stamped with its time step. In the visible condition the frames of the event enter memory and the stamps advance over it, whereas in the hidden condition neither happens.
(b) Successes over 50 paired episodes of the released FrameSamp-Modul policy; dashed line: no event. Air-grasps are not used on PickXtimes, where the returning gripper often re-closes around the cube.}
\label{fig:phenomenon}
\end{wrapfigure}
\Cref{fig:phenomenon} applies the three null events of \cref{sec:setup} to the released policy, and every one of them is damaging.
Two 5-second pauses reduce success from 47 to 16 of 50 on PickXtimes and from 44 to 10 on SwingXtimes.
Wrist fidgets, in which the robot keeps moving, do as much damage, and air-grasps that return the arm to where it started still halve success on SwingXtimes.
Nor do the events need to be long, since pauses of 1.5 seconds already cause most of the loss (\cref{supp:robust}).

How the policy fails is as telling as how often it fails.
A disturbance that merely confused the policy would cause failures of many kinds, such as dropped cubes or extra repetitions.
Under pauses and fidgets, however, nearly all failures are early stops, in which the policy presses the button before the $K$-th repetition (\cref{supp:cells}).
The policy thus overestimates how far it has come.
\Cref{tab:byK} shows where this matters.
With $K{=}1$, the first natural moment to press, right after the first repetition, is already the correct one, and most of these episodes survive.
With $K{\ge}2$, in contrast, not a single episode succeeds under pauses or fidgets, while hiding the pauses restores nearly all of them.

Such early stopping could stem from the interruption itself or from what the policy remembers of it.
The hidden condition separates the two, because it keeps the physical event exactly as it was and changes only what enters memory.
The robot pauses, fidgets or grasps at air as before, but the frames of the event are withheld and the time stamps do not advance over it.
In every cell, success then returns to the level without any event (\cref{fig:phenomenon}b).
\textbf{What breaks the policy is its memory of an interval in which nothing happened.}

\paragraph{Beyond counting.}
Null events also hurt some tasks that open with a demonstration video, through a different consequence of the same time-filled memory.
On RouteStick, for instance, two 15-second pauses reduce success from 33 to 7, against 36 when hidden.
The video shares the 32 time-sampled slots with execution, so the null frames push it out of memory.
Indeed, giving the video only the slots it keeps under the pause reproduces the damage without any pause, and protecting its slots during the pause restores success (\cref{supp:tasks}).
Whether it adds progress that never happened or pushes out a cue the task needs, the null event acts through the room it takes in memory.

\needlines{6}
\section{Why: A Memory Filled by Time Counts Null Events}
\label{sec:mechanism}

\begin{wrapfigure}{r}{0.48\linewidth}
\centering
\includegraphics[width=\linewidth]{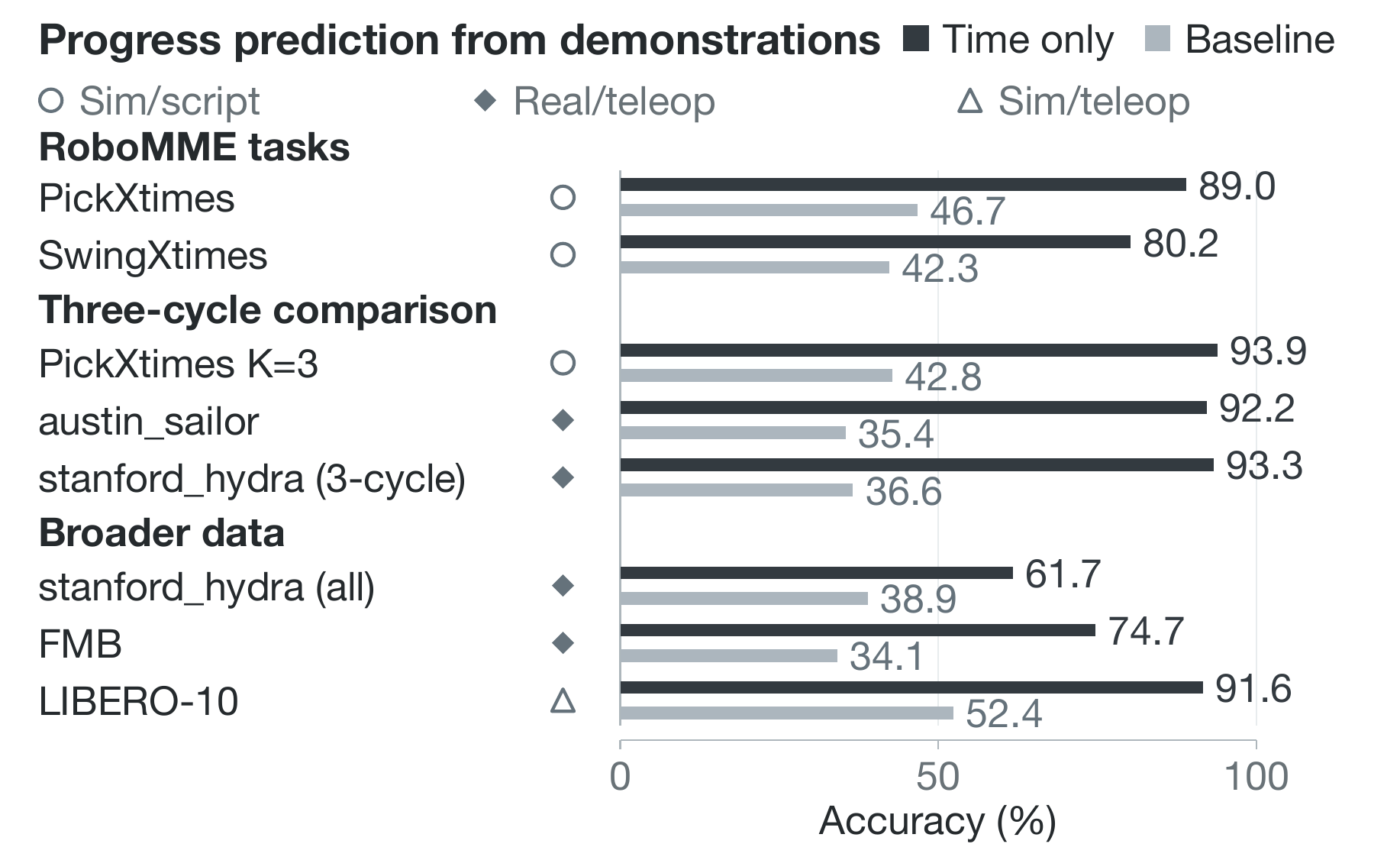}
\caption{\textbf{In demonstrations, elapsed time alone predicts progress.}
Accuracy of predicting the number of completed repetitions from the time step alone (5-fold cross-validation over demonstrations), against always predicting the most frequent count (baseline).
Three-cycle comparison: demonstrations with three repetitions, in simulation and in two real teleoperation datasets~\cite{nasiriany2022sailor,belkhale2023hydra}.
Broader data: all stanford\_hydra tasks, the annotated primitives of FMB~\cite{luo2024fmb}, and the two-object tasks of LIBERO-10~\cite{liu2023libero} (\cref{supp:realdata}).}
\label{fig:realdata}
\end{wrapfigure}
\Cref{sec:phenomenon} showed that, after an interval in which nothing happened, the policy overestimates how far it has come.
We argue that it judges progress from what time puts into its memory rather than from what happened.
A null interval enters a time-filled memory through two routes.
First, the time stamps advance over it and thereby record how long it lasted.
Second, its frames occupy slots of the memory, which registers that something happened.
Below we show why a policy learns to read time (\cref{sec:data}), that it follows the stamps (\cref{sec:clock}), that the slots alone suffice once the stamps are gone (\cref{sec:noclock}), and that vulnerability tracks the index across designs (\cref{sec:designs}).

\needlines{6}
\subsection{In demonstrations, time predicts progress}
\label{sec:data}

Demonstrations proceed at a steady pace, so the time elapsed in an episode reveals how far it has progressed.
To measure this, we predict at every step of a demonstration how many repetitions have been completed, from elapsed time alone, using the most frequent count at that time in held-out demonstrations (\cref{supp:realdata}).
This predictor is right 89\% of the time on PickXtimes and 80\% on SwingXtimes, roughly twice as often as always guessing the most common count.
Nor is this an artefact of scripted simulation data (\cref{fig:realdata}).
On demonstrations with three repetitions, the predictor is as accurate on two real teleoperation datasets from Open X-Embodiment~\cite{oxe2024} as on PickXtimes.
On broader real data, where the pace is less regular, it is less accurate but still far above the baseline.
Moreover, demonstrations contain almost no idle stretches, so nothing in training ever penalises a policy that reads progress from time.
For imitation learning, which rewards any feature that predicts the next action, elapsed time is therefore a cheap and reliable cue.

\needlines{6}
\subsection{Rescaling time shifts the failure}
\label{sec:clock}

\begin{wrapfigure}{r}{0.48\linewidth}
\centering
\includegraphics[width=\linewidth]{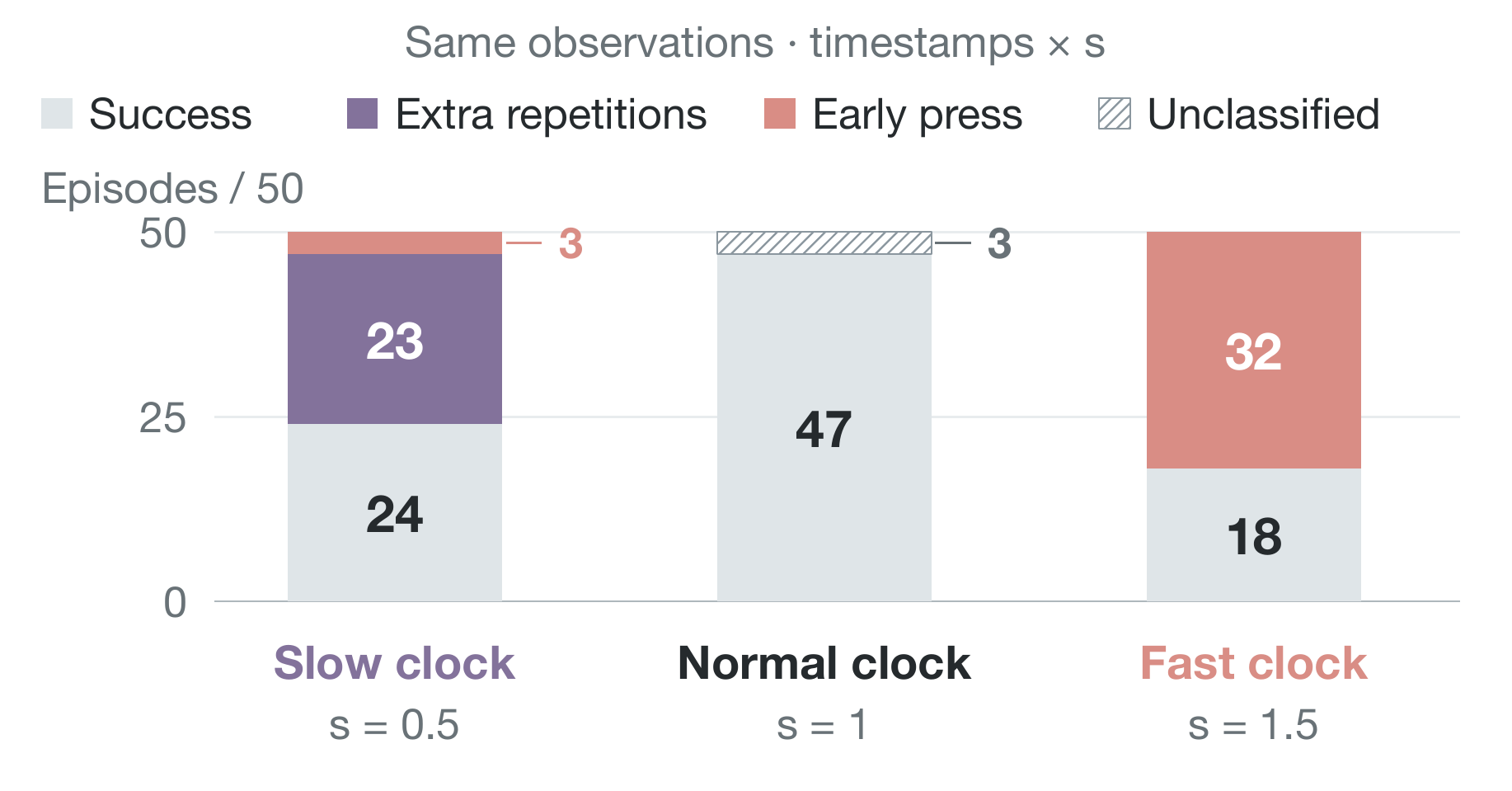}
\caption{\textbf{Rescaling the time stamps shifts the failure.}
No null event is injected and the frames are unchanged. Only the time stamps of the memory frames are multiplied by $s$.
Released FrameSamp-Modul on PickXtimes, 50 paired episodes.
A slower clock ($s{=}0.5$) causes extra repetitions, a faster one ($s{=}1.5$) early presses. Hatched: failures whose type was not logged.}
\label{fig:clock}
\end{wrapfigure}
If the policy reads progress from time, then altering the time stamps alone, without any null event, should shift its estimate of progress in a predictable direction.
We therefore multiply the time stamps of all memory frames by a factor $s$ and leave the frames themselves untouched (\cref{fig:clock}).
When time appears to run faster ($s{=}1.5$), success falls from 47 to 18, and every failure is an early press.
When time appears to run slower ($s{=}0.5$), the policy makes extra repetitions instead.
The direction of the failure thus follows the direction of the clock.
The flip is weaker on SwingXtimes, but it recurs in FrameSamp-Expert, which reads the same memory through a different pathway, and it persists when the stamps keep the in-distribution layout of a training episode (\cref{supp:clock}).

\begin{wrapfigure}{r}{0.55\linewidth}
\centering
\includegraphics[width=\linewidth]{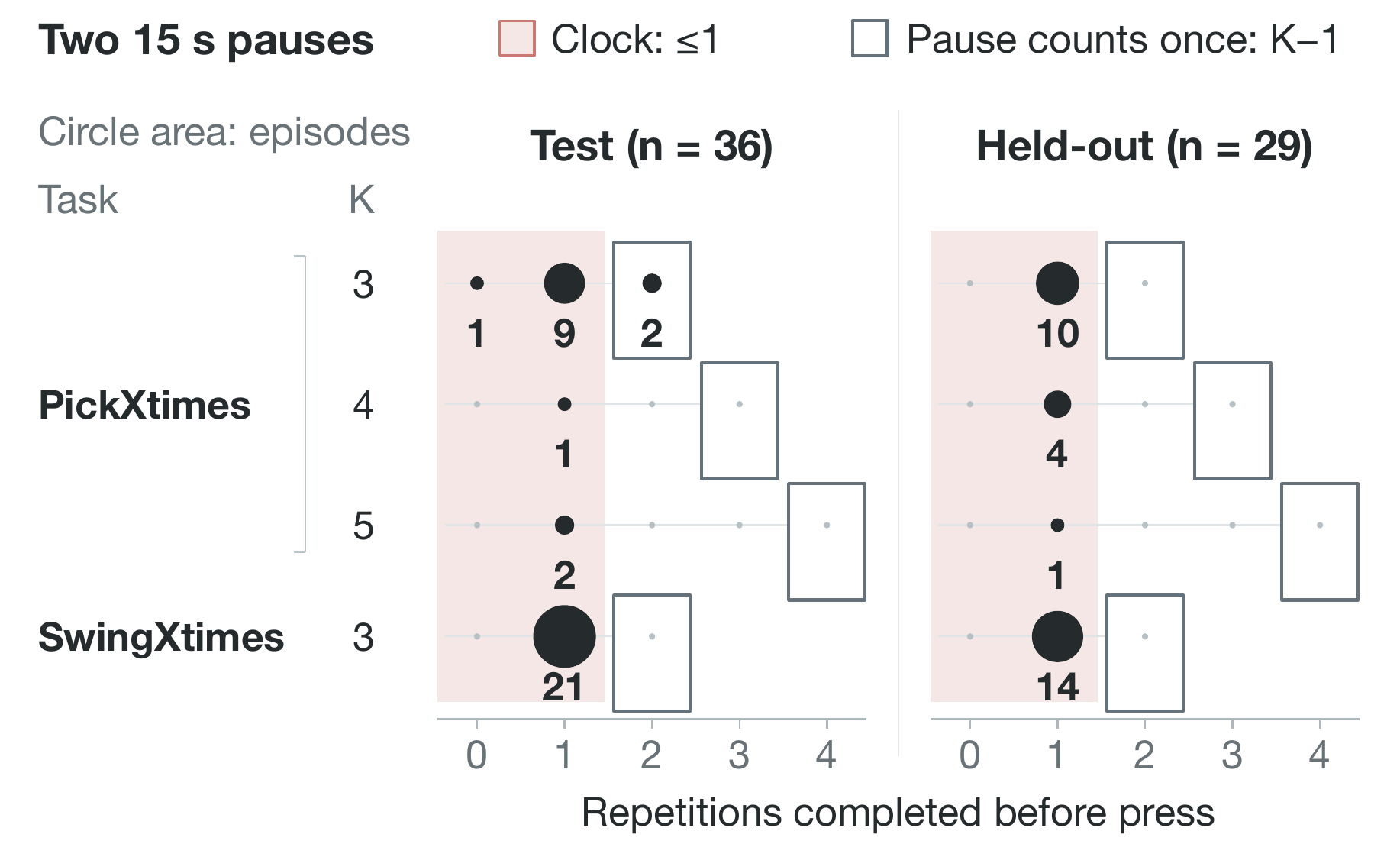}
\caption{\textbf{The policy reads how long a pause lasts.}
Repetitions completed before an early press under two 15-second pauses, for \clk{} on episodes with $K{\ge}3$ that succeed without the pauses.
Reading the clock predicts a press after at most one repetition (shaded), whereas counting the pauses as one repetition predicts $K{-}1$ (boxes).
Held-out: episodes not used for any design decision, with the criterion fixed before the run.
For $K{\le}2$ the two readings predict the same press and are omitted.}
\label{fig:clockcount}
\end{wrapfigure}

A null event thus acts like a local acceleration of this clock, since time advances while the content stands still.
If so, the policy should respond to how long a null event lasts.
A simpler reading would have it count each null interval as one extra repetition, whatever its length.
Long pauses separate the two readings (\cref{fig:clockcount}).
With two 15-second pauses, the clock passes its usual press time after one more repetition, so a policy that reads it should press after at most one repetition, whatever $K$.
A policy that counts the pauses instead should press after $K{-}1$.
On \clk{}, 34 of 36 presses come after at most one repetition, and on held-out episodes, with this criterion fixed in advance, all 29 do.
\textbf{The stamps therefore carry how long a null event lasts, and the policy reads it.}
Yet the stamps are not the only route.
Withholding the pause frames while advancing the stamps over them reproduces the full damage on PickXtimes, but storing the frames while freezing the stamps also hurts (\cref{supp:clock:route}).
The frames of a null event must thus carry part of the damage on their own.

\needlines{6}
\subsection{Removing the clock does not remove the failure}
\label{sec:noclock}

\begin{figure}[b]
\centering
\includegraphics[width=\linewidth]{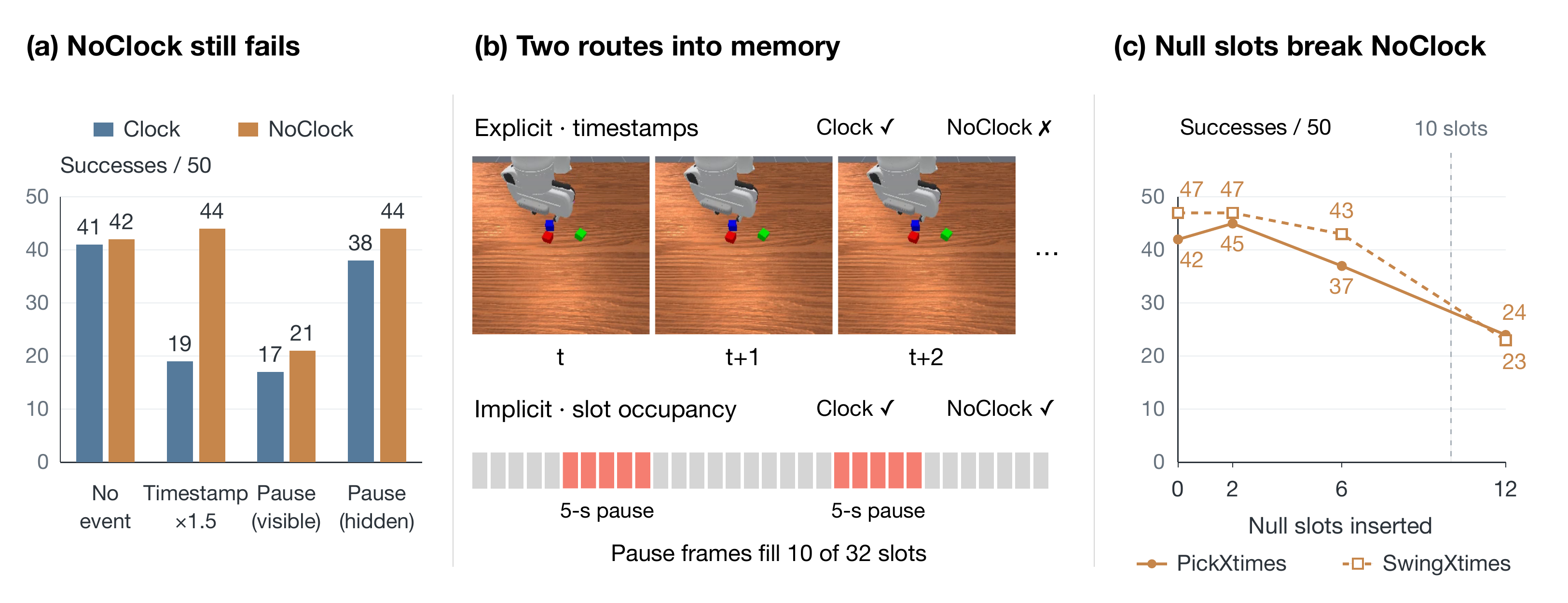}
\caption{\textbf{Removing the clock is not enough.}
(a)~\nope{}, trained without time encoding, ignores rescaled time stamps, yet two 5-second pauses still break it; hiding them restores both models (PickXtimes, successes of 50 paired episodes; both tasks in \cref{supp:noclock}).
(b)~A pause enters a time-sampled memory by two routes: time stamps keep advancing over unchanged frames, which only \clk{} receives, and pause frames take a share of the 32 slots, which both models receive (10 slots in the episode shown, about 450 working steps; frames from its first pause).
(c)~Without any pause, blocks of null slots alone degrade \nope{} (\cref{supp:noclock:dose}). The dashed line marks the share two 5-second pauses take.}
\label{fig:twist}
\end{figure}
If the stamps were the whole story, removing them would remove the failure.
To test this, we trained \nope{} from scratch without any time encoding (\cref{fig:twist}a; \cref{supp:noclock}).
As intended, it ignores rescaled stamps, and it counts as well as \clk{}.
Yet pauses still cut its success to 21 and 14 of 50, and hiding them restores it, so a policy that never sees a time stamp still counts a pause as progress.

The reason is the second route (\cref{fig:twist}b).
Without stamps, a memory sampled uniformly in time still fills a share of its 32 slots with frames of an interval in which nothing happened.
Two controls show that the harm comes from these uneventful slots themselves, rather than from what the null frames depict or from the real frames they displace.
Replacing every null frame with the last frame before the pause leaves the damage unchanged, while filling the same slots with copies of real frames, without any pause, leaves success intact.
Conversely, inserting blocks of identical frames without any pause degrades \nope{} more and more as the blocks grow (\cref{fig:twist}c).
\textbf{The problem therefore lies deeper than the stamps: in a memory filled by time, an interval in which nothing happens still takes up room.}

\needlines{6}
\subsection{Vulnerability tracks the index across RoboMME's released designs}
\label{sec:designs}

\begin{wrapfigure}{r}{0.5\linewidth}
\centering
\includegraphics[width=\linewidth]{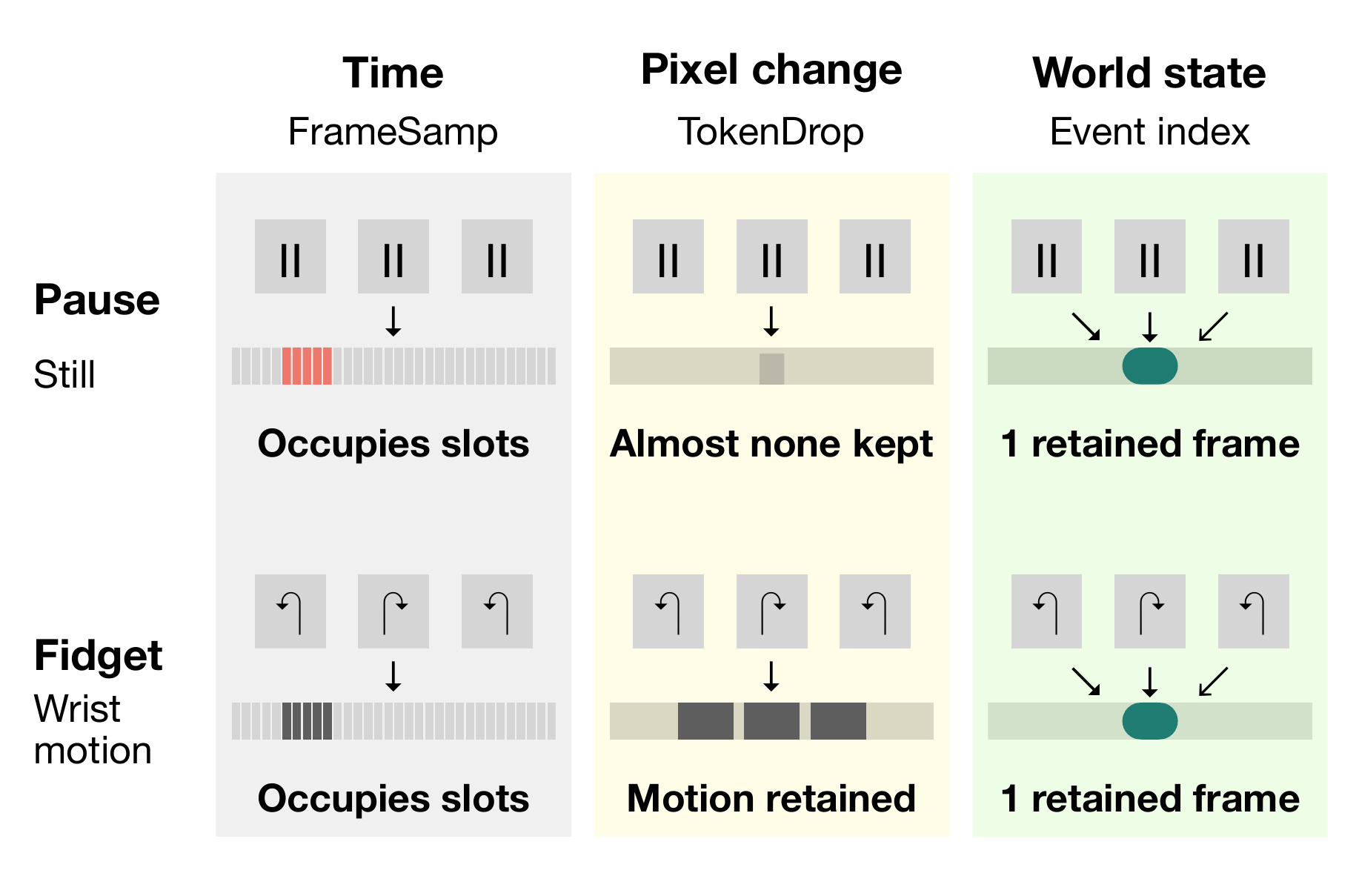}
\caption{\textbf{What a null event leaves in memory depends on what the memory counts.}
A memory indexed by time gives a null event slots in proportion to its duration, whether the robot is still or moving.
Selecting by pixel change keeps little of a pause but retains a fidget, which changes the pixels without changing the world.
The event index of \cref{sec:fixes} advances only when the world changes and reduces either to a single retained frame.}
\label{fig:criterion}
\end{wrapfigure}

\begin{wraptable}{r}{0.55\linewidth}
\centering
\small
\setlength{\tabcolsep}{3pt}
\begin{tabular}{llccc}
\toprule
Memory & Index & \multicolumn{2}{c}{PickXtimes} & SwingX. \\
\cmidrule(lr){3-4}\cmidrule(lr){5-5}
 & & Pause & Fidget & Pause \\
\midrule
FrameSamp-Modul & time & 47$\to$16 & 47$\to$16 & 44$\to$12 \\
FrameSamp-Expert & time & 44$\to$14 & 44$\to$17 & 49$\to$11 \\
FrameSamp-Context & time & 33$\to$11 & 33$\to$20 & 43$\to$10 \\
TokenDrop-Modul & pixel change & 39$\to$30 & 39$\to$21 & 45$\to$27 \\
\midrule
MemER-style & sub-goal & 38$\to$39 & 38$\to$38 & 24$\to$23 \\
Text sub-goals & sub-goal & 46$\to$48 & 46$\to$47 & -- \\
\midrule
TTT & recency & 23$\to$22 & 23$\to$25 & 17$\to$16 \\
$\pi_{0.5}$ & none & 18$\to$19 & 18$\to$20 & 15$\to$13 \\
\midrule
SimpleMemVLA$^\dagger$ & every step & 50$\to$47 & 50$\to$38 & 48$\to$36 \\
LTM-$\pi^\dagger$ & every step & 50$\to$7 & 50$\to$15 & 50$\to$9 \\
\bottomrule
\end{tabular}
\caption{\textbf{Vulnerability tracks the index among RoboMME's released designs.}
Success without and with two null events (pauses of 300 steps, against 100 in \cref{fig:phenomenon}; fidgets of 100 steps), 50 paired episodes.
Frame-sampled memories indexed by time collapse, selection by pixel change is hurt less by still frames than by moving ones, and memories indexed by sub-goals are unaffected.
Text sub-goals on SwingXtimes are at floor (1/50).
$^\dagger$Independent policies on other backbones (\cref{sec:designs}): SimpleMemVLA keeps every frame with its time as text, and LTM-$\pi$ adds one motion token per step. Events of 100 steps, against the same events hidden from memory (SimpleMemVLA) or no event (LTM-$\pi$).}
\label{tab:designs}
\end{wraptable}
If the damage comes from indexing memory by time, it should depend on how a memory is indexed rather than on whether a policy has memory at all.
\Cref{tab:designs} tests this by applying the same null events to every released RoboMME memory policy, without retraining (\cref{supp:designs}).
The three FrameSamp variants, whose slots are sampled and stamped by time, all collapse.
TokenDrop selects tokens by pixel change, so still frames rarely enter its memory, and pauses cost it much less.
A fidget, however, changes the pixels without changing the world, so it passes this filter and costs TokenDrop far more than a pause.
Memories indexed by sub-goals, a MemER-style keyframe memory~\cite{sridhar2026memer} and a text memory, record progress only when a sub-goal is reached, and neither null event affects them.
Each design thus fails where its index lets a null event in (\cref{fig:criterion}).
Sub-goal memories pay for this elsewhere, as they count less well without null events, and the keyframe memory is misled by air-grasps (\cref{supp:designs}).

Two independently developed policies on other backbones fit this picture less neatly.
Both add to memory at every step and both read fidgets as progress, mostly through early presses.
The same pauses, however, barely affect SimpleMemVLA~\cite{yin2026simplememvla} on PickXtimes, whereas they collapse LTM-$\pi$~\cite{sun2026ltm} (\cref{supp:smv,supp:ltm}).

If the index decides which null events reach memory, changing the index rather than retraining the policy should keep them out, which \cref{sec:fixes} tests.

\needlines{6}
\section{What the Diagnosis Predicts for Fixes}
\label{sec:fixes}

The diagnosis points to two places where the failure could be fixed, and it predicts different outcomes for each.
The first is the data: training with injected null events should break the link between time and progress, but only for null events that can be faithfully synthesised.
A held frame is a valid pause, whereas a moving null event cannot be made by replaying recorded frames of real-robot data, because the replay would move the objects as well and so show a real repetition.
The second is the index: if memory advanced with events instead of time, null events would leave no trace, whether or not the robot moves.

\begin{wrapfigure}{r}{0.54\linewidth}
\centering
\includegraphics[width=\linewidth]{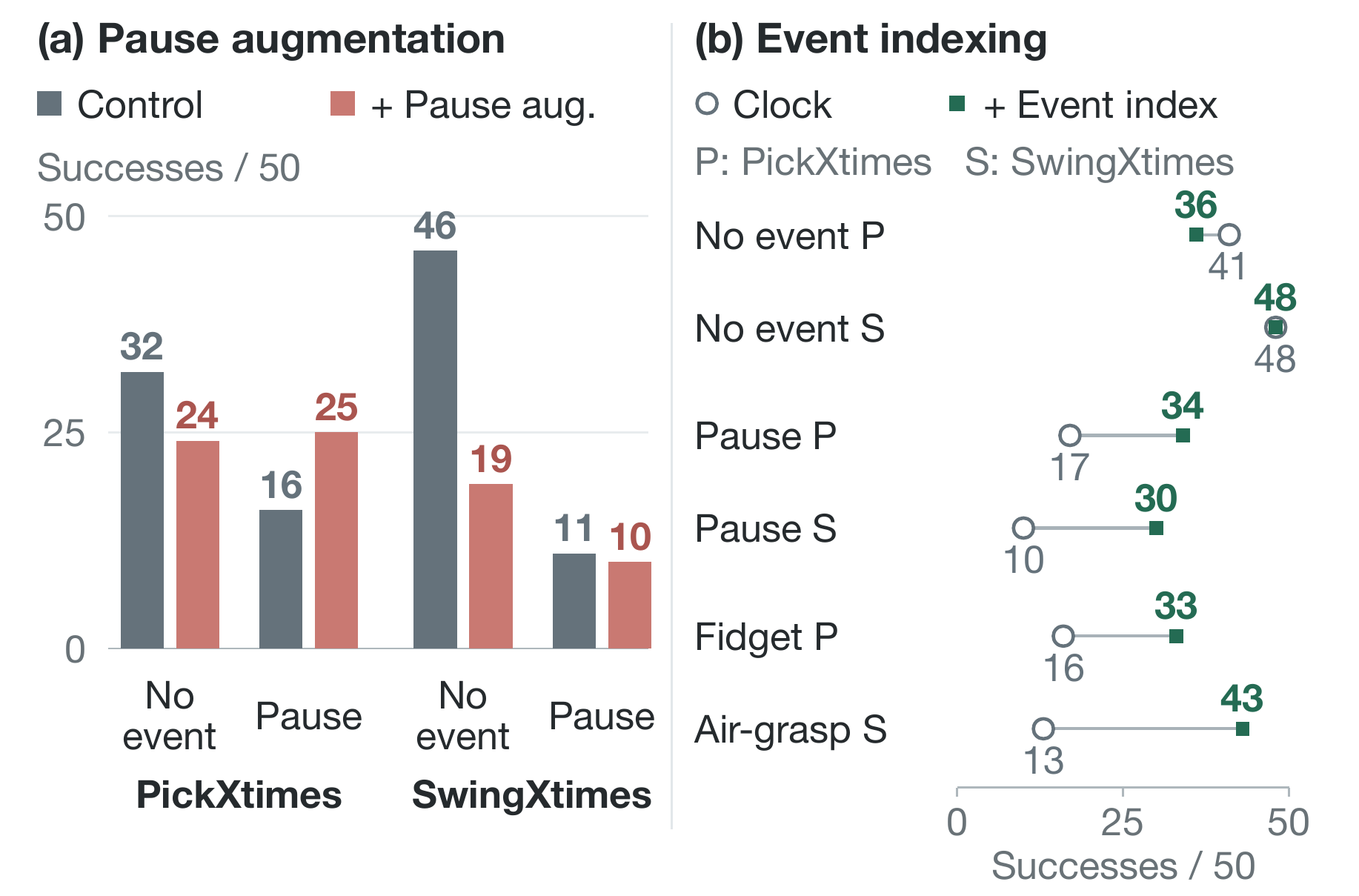}
\caption{\textbf{Pause augmentation does not fix the failure; changing the index does.}
(a) Pause augmentation, trained on a second machine and compared with a retrain without it on that machine (Control): on PickXtimes pauses stop hurting only because success without any null event falls.
(b) Event indexing at test time on the same \clk{} checkpoint; hiding the pause frames entirely gives 38 and 38.
Successes of 50 paired episodes, test split; full numbers, a weaker augmentation and a held-out replication in \cref{supp:eventindex}.}
\label{fig:fixes}
\end{wrapfigure}

\paragraph{Changing the data.}
We retrained \clk{} with pauses injected into its demonstrations and compared it with a retrain without them on the same machine (\cref{fig:fixes}a).
The prediction failed: on PickXtimes, pauses no longer cost the augmented policy anything, but only because its success without any null event has already fallen, and on SwingXtimes it does no better under pauses than the retrain without augmentation.
The augmented policy still solves single repetitions, yet it loses most episodes that require counting further.
We did not predict this cost, and a post hoc reading is that its counting leaned on time and lost that support, as a faster clock no longer hurts it, although crowded memory or slower learning may contribute.
A weaker dose costs less counting but helps no more under pauses (\cref{supp:pauseonly}).

\paragraph{Changing the index.}
We implement the event index at test time, on the same \clk{} checkpoint and without retraining.
A stretch of the episode counts as a null event if the world returns to a state it was already in.
Concretely, the robot's configuration comes back to that of a kept frame, the scene outside the robot is unchanged, and no contact was made or released in between.
Each such stretch is collapsed to a single frame, as is any prolonged stillness, and the time positions are recomputed over the kept frames, so that memory advances only with events.
The detector reads the robot's silhouette, the gripper width and the external force on the arm from the simulator, with tolerances fixed across all 16 RoboMME tasks (\cref{supp:eventindex}).
Since its criterion mirrors how our null events are built, the experiment tests whether taking null frames out of the index suffices, rather than offering a deployable detector.

Taking null frames out of the index does suffice (\cref{fig:fixes}b).
Event indexing at least doubles success under all three null events, including the two in which the robot moves, and under air-grasps it lifts SwingXtimes from 13 to 43 of 50.
The gain replicates on held-out episodes and on the released policy (\cref{supp:eventindex}).
Under pauses on SwingXtimes it still falls short of the level without events, so part of the damage survives the index.

\paragraph{Where event indexing can fail.}
The criterion that makes the index work also sets its blind spot.
On a counting task a true repetition also returns the world to a previous state, since the robot releases the cube and comes back to it.
The index can therefore mistake a real repetition for nothing.
On PickXtimes without null events it costs some success, little on \clk{} but significantly on the released policy, whereas on the other 15 RoboMME tasks it costs nothing overall (\cref{supp:le-cost}).
Two kinds of null event escape the criterion in part, namely a grasp that ends elsewhere, which never returns to a previous state, and brief, randomly timed micro-stops, which are neither still nor closed loops.
Of the damage done by micro-stops, the index recovers little (\cref{supp:eventindex}).
Finally, where time itself is the task, as in TimedTransfer, collapsing a pause is wrong by construction.
An event index is thus the right default for memories that must track what happened, while tasks that depend on duration still need a clock.

\needlines{6}
\section{Discussion and Limitations}
\label{sec:discussion}

\paragraph{Benchmarks may measure clock reading.}
Counting tasks are meant to test memory of events, but when demonstrations proceed at a steady pace, a policy can pass them by reading the clock.
The shortcut is unlikely to be specific to RoboMME.
In an exploratory study on MIKASA-Robo~\cite{cherepanov2025mikasa}, a policy appears to infer how many times a light blinked from how long the cue lasted (\cref{supp:mikasa}).
Null events, clock rescaling and fixed-duration cues are cheap probes that any memory benchmark can add to tell event memory from time keeping.

\paragraph{What should a memory index?}
Our results suggest a principle for memory design.
Memory should advance when the world changes, because that is what progress on a task consists of.
Elapsed time is a convenient proxy, since demonstrations proceed at a steady pace, yet it also advances when nothing happens.
Pixel change filters out stillness, yet a robot can move without changing anything.
Changes in the world itself, such as an object moved or a contact made, track progress and are exactly what null events lack.
Time keeps a role where duration is part of the task, as in TimedTransfer, but there it should enter as an input to the policy, while what memory keeps should still follow events.
Detecting such changes from real sensors is harder than in simulation, and we see it as the main open problem for memories that track what happened.

\paragraph{Limitations.}
Our causal analysis covers one family of memory policies (MME-VLA) and models retrained with its architecture, and two independent policies are probed without dissection.
All experiments are in simulation and every null event is scripted, so we do not measure how often unannounced null events occur in deployment, although the correlation behind the shortcut is as strong in regular real teleoperation demonstrations.
The event detector, too, uses simulator signals.
A second seed of \clk{} and \nope{} met every rule fixed for it (\cref{supp:seed}), but the other models, including each augmentation dose, have one seed.
Since training reruns of \clk{} differ by up to 12 episodes on PickXtimes without events (\cref{supp:stats}), we compare separately trained models only where the gap is larger.
Finally, without time stamps it remains open which feature of the occupied slots the policy reads.

\paragraph{Conclusion.}
Memory VLAs whose memory is filled by time record that time passed as well as what happened, and they read the former as progress.
Null events expose this shortcut, rescaling the stamps alone reproduces it, and removing the clock does not cure it, because null frames still occupy memory.
Indexing memory by events recovers much of the loss, within limits set by its criterion.
A memory meant to track what happened should therefore advance with events rather than with the clock.

{
    \small
    \bibliographystyle{ieeenat_fullname}
    \bibliography{main}

\begin{thebibliography}{59}
\providecommand{\natexlab}[1]{#1}
\providecommand{\url}[1]{\texttt{#1}}
\expandafter\ifx\csname urlstyle\endcsname\relax
  \providecommand{\doi}[1]{doi: #1}\else
  \providecommand{\doi}{doi: \begingroup \urlstyle{rm}\Url}\fi

\bibitem[Anil et~al.(2022)Anil, Wu, Andreassen, Lewkowycz, Misra, Ramasesh, Slone, Gur-Ari, Dyer, and Neyshabur]{anil2022length}
Cem Anil, Yuhuai Wu, Anders Andreassen, Aitor Lewkowycz, Vedant Misra, Vinay Ramasesh, Ambrose Slone, Guy Gur-Ari, Ethan Dyer, and Behnam Neyshabur.
\newblock {Exploring Length Generalization in Large Language Models}.
\newblock In \emph{Advances in Neural Information Processing Systems (NeurIPS)}, 2022.

\bibitem[Belkhale et~al.(2023)Belkhale, Cui, and Sadigh]{belkhale2023hydra}
Suneel Belkhale, Yuchen Cui, and Dorsa Sadigh.
\newblock {HYDRA: Hybrid Robot Actions for Imitation Learning}.
\newblock In \emph{Conference on Robot Learning (CoRL)}, 2023.

\bibitem[Black et~al.(2025{\natexlab{a}})Black, Brown, Driess, Esmail, Equi, Finn, Fusai, Groom, Hausman, Ichter, Jakubczak, Jones, Ke, Levine, Li-Bell, Mothukuri, Nair, Pertsch, Shi, Tanner, Vuong, Walling, Wang, and Zhilinsky]{black2025pi0}
Kevin Black, Noah Brown, Danny Driess, Adnan Esmail, Michael Equi, Chelsea Finn, Niccolo Fusai, Lachy Groom, Karol Hausman, Brian Ichter, Szymon Jakubczak, Tim Jones, Liyiming Ke, Sergey Levine, Adrian Li-Bell, Mohith Mothukuri, Suraj Nair, Karl Pertsch, Lucy~Xiaoyang Shi, James Tanner, Quan Vuong, Anna Walling, Haohuan Wang, and Ury Zhilinsky.
\newblock {$\pi_0$: A Vision-Language-Action Flow Model for General Robot Control}.
\newblock In \emph{Robotics: Science and Systems (RSS)}, 2025{\natexlab{a}}.

\bibitem[Black et~al.(2025{\natexlab{b}})Black, Galliker, and Levine]{black2025rtc}
Kevin Black, Manuel~Y. Galliker, and Sergey Levine.
\newblock {Real-Time Execution of Action Chunking Flow Policies}.
\newblock In \emph{Advances in Neural Information Processing Systems (NeurIPS)}, 2025{\natexlab{b}}.

\bibitem[Bolya et~al.(2023)Bolya, Fu, Dai, Zhang, Feichtenhofer, and Hoffman]{bolya2023tome}
Daniel Bolya, Cheng-Yang Fu, Xiaoliang Dai, Peizhao Zhang, Christoph Feichtenhofer, and Judy Hoffman.
\newblock {Token Merging: Your ViT But Faster}.
\newblock In \emph{International Conference on Learning Representations (ICLR)}, 2023.

\bibitem[Bulatov et~al.(2022)Bulatov, Kuratov, and Burtsev]{bulatov2022rmt}
Aydar Bulatov, Yuri Kuratov, and Mikhail~S. Burtsev.
\newblock {Recurrent Memory Transformer}.
\newblock In \emph{Advances in Neural Information Processing Systems (NeurIPS)}, 2022.

\bibitem[Cadena et~al.(2016)Cadena, Carlone, Carrillo, Latif, Scaramuzza, Neira, Reid, and Leonard]{cadena2016slam}
Cesar Cadena, Luca Carlone, Henry Carrillo, Yasir Latif, Davide Scaramuzza, Jose Neira, Ian Reid, and John~J. Leonard.
\newblock {Past, Present, and Future of Simultaneous Localization And Mapping: Towards the Robust-Perception Age}.
\newblock \emph{IEEE Transactions on Robotics}, 32\penalty0 (6):\penalty0 1309--1332, 2016.

\bibitem[Cherepanov et~al.(2025)Cherepanov, Kachaev, Kovalev, and Panov]{cherepanov2025mikasa}
Egor Cherepanov, Nikita Kachaev, Alexey~K. Kovalev, and Aleksandr~I. Panov.
\newblock {Memory, Benchmark \& Robots: A Benchmark for Solving Complex Tasks with Reinforcement Learning}.
\newblock \emph{arXiv preprint arXiv:2502.10550}, 2025.

\bibitem[Chi et~al.(2023)Chi, Xu, Feng, Cousineau, Du, Burchfiel, Tedrake, and Song]{chi2023diffusion}
Cheng Chi, Zhenjia Xu, Siyuan Feng, Eric Cousineau, Yilun Du, Benjamin Burchfiel, Russ Tedrake, and Shuran Song.
\newblock {Diffusion Policy: Visuomotor Policy Learning via Action Diffusion}.
\newblock In \emph{Robotics: Science and Systems (RSS)}, 2023.

\bibitem[Chuang et~al.(2022)Chuang, Yang, Wen, and Gao]{chuang2022residual}
Chia-Chi Chuang, Donglin Yang, Chuan Wen, and Yang Gao.
\newblock {Resolving Copycat Problems in Visual Imitation Learning via Residual Action Prediction}.
\newblock In \emph{European Conference on Computer Vision (ECCV)}, 2022.

\bibitem[Dai et~al.(2026)Dai, Fu, Lee, Liu, Zhang, Yang, Finn, Fazeli, and Chai]{dai2026robomme}
Yinpei Dai, Hongze Fu, Jayjun Lee, Yuejiang Liu, Haoran Zhang, Jianing Yang, Chelsea Finn, Nima Fazeli, and Joyce Chai.
\newblock {RoboMME: Benchmarking and Understanding Memory for Robotic Generalist Policies}.
\newblock In \emph{International Conference on Machine Learning (ICML)}, 2026.

\bibitem[de~Haan et~al.(2019)de~Haan, Jayaraman, and Levine]{dehaan2019causal}
Pim de Haan, Dinesh Jayaraman, and Sergey Levine.
\newblock {Causal Confusion in Imitation Learning}.
\newblock In \emph{Advances in Neural Information Processing Systems (NeurIPS)}, 2019.

\bibitem[Fang et~al.(2025)Fang, Grotz, Pumacay, Wang, Fox, Krishna, and Duan]{fang2025sam2act}
Haoquan Fang, Markus Grotz, Wilbert Pumacay, Yi~Ru Wang, Dieter Fox, Ranjay Krishna, and Jiafei Duan.
\newblock {SAM2Act: Integrating Visual Foundation Model with A Memory Architecture for Robotic Manipulation}.
\newblock In \emph{International Conference on Machine Learning (ICML)}, 2025.

\bibitem[Fei et~al.(2025)Fei, Wang, Shi, Dai, Cai, Qian, Ji, He, Zhang, Fei, Fu, Gong, and Qiu]{fei2025liberoplus}
Senyu Fei, Siyin Wang, Junhao Shi, Zihao Dai, Jikun Cai, Pengfang Qian, Li Ji, Xinzhe He, Shiduo Zhang, Zhaoye Fei, Jinlan Fu, Jingjing Gong, and Xipeng Qiu.
\newblock {LIBERO-Plus: In-depth Robustness Analysis of Vision-Language-Action Models}.
\newblock \emph{arXiv preprint arXiv:2510.13626}, 2025.

\bibitem[Fountas et~al.(2025)Fountas, Benfeghoul, Oomerjee, Christopoulou, Lampouras, Bou-Ammar, and Wang]{fountas2025emllm}
Zafeirios Fountas, Martin~A Benfeghoul, Adnan Oomerjee, Fenia Christopoulou, Gerasimos Lampouras, Haitham Bou-Ammar, and Jun Wang.
\newblock {Human-inspired Episodic Memory for Infinite Context LLMs}.
\newblock In \emph{International Conference on Learning Representations (ICLR)}, 2025.

\bibitem[Gao et~al.(2026)Gao, Ni, Wen, Xu, Li, Xu, and Wang]{gao2026agm}
Hongbo Gao, Zeyu Ni, Xin Wen, Siyu Xu, Ruifeng Li, Rui Xu, and Yunke Wang.
\newblock {AGM: Achievement-Grounded Memory for Closed-Loop Agents with Frozen VLA Policies}.
\newblock \emph{arXiv preprint arXiv:2608.29537}, 2026.

\bibitem[Geirhos et~al.(2020)Geirhos, Jacobsen, Michaelis, Zemel, Brendel, Bethge, and Wichmann]{geirhos2020shortcut}
Robert Geirhos, J{\"o}rn-Henrik Jacobsen, Claudio Michaelis, Richard Zemel, Wieland Brendel, Matthias Bethge, and Felix~A. Wichmann.
\newblock {Shortcut Learning in Deep Neural Networks}.
\newblock \emph{Nature Machine Intelligence}, 2\penalty0 (11):\penalty0 665--673, 2020.

\bibitem[Guhur et~al.(2022)Guhur, Chen, Garcia, Tapaswi, Laptev, and Schmid]{guhur2022hiveformer}
Pierre-Louis Guhur, Shizhe Chen, Ricardo Garcia, Makarand Tapaswi, Ivan Laptev, and Cordelia Schmid.
\newblock {Instruction-driven history-aware policies for robotic manipulations}.
\newblock In \emph{Conference on Robot Learning (CoRL)}, 2022.

\bibitem[He et~al.(2024)He, Li, Jang, Jia, Cao, Shah, Shrivastava, and Lim]{he2024malmm}
Bo He, Hengduo Li, Young~Kyun Jang, Menglin Jia, Xuefei Cao, Ashish Shah, Abhinav Shrivastava, and Ser-Nam Lim.
\newblock {MA-LMM: Memory-Augmented Large Multimodal Model for Long-Term Video Understanding}.
\newblock In \emph{IEEE/CVF Conference on Computer Vision and Pattern Recognition (CVPR)}, 2024.

\bibitem[Ijspeert et~al.(2013)Ijspeert, Nakanishi, Hoffmann, Pastor, and Schaal]{ijspeert2013dmp}
Auke~Jan Ijspeert, Jun Nakanishi, Heiko Hoffmann, Peter Pastor, and Stefan Schaal.
\newblock {Dynamical Movement Primitives: Learning Attractor Models for Motor Behaviors}.
\newblock \emph{Neural Computation}, 25\penalty0 (2):\penalty0 328--373, 2013.

\bibitem[Kazemnejad et~al.(2023)Kazemnejad, Padhi, Ramamurthy, Das, and Reddy]{kazemnejad2023nope}
Amirhossein Kazemnejad, Inkit Padhi, Karthikeyan~Natesan Ramamurthy, Payel Das, and Siva Reddy.
\newblock {The Impact of Positional Encoding on Length Generalization in Transformers}.
\newblock In \emph{Advances in Neural Information Processing Systems (NeurIPS)}, 2023.

\bibitem[Koo et~al.(2026)Koo, Choi, Kim, Lee, Kim, Seo, and Shin]{koo2026hamlet}
Myungkyu Koo, Daewon Choi, Taeyoung Kim, Kyungmin Lee, Changyeon Kim, Younggyo Seo, and Jinwoo Shin.
\newblock {HAMLET: Switch your Vision-Language-Action Model into a History-Aware Policy}.
\newblock In \emph{International Conference on Learning Representations (ICLR)}, 2026.

\bibitem[Li et~al.(2024)Li, Liu, Zhang, Yu, Xu, Wu, Cheang, Jing, Zhang, Liu, Li, and Kong]{li2024roboflamingo}
Xinghang Li, Minghuan Liu, Hanbo Zhang, Cunjun Yu, Jie Xu, Hongtao Wu, Chilam Cheang, Ya Jing, Weinan Zhang, Huaping Liu, Hang Li, and Tao Kong.
\newblock {Vision-Language Foundation Models as Effective Robot Imitators}.
\newblock In \emph{International Conference on Learning Representations (ICLR)}, 2024.

\bibitem[Liao and Cao(2026)]{liao2026present}
Chih-Ting Liao and Xin Cao.
\newblock {Present but Not Remembered: Auditing How Frozen VLAs Encode, Deploy, and Steer Visual History}.
\newblock \emph{arXiv preprint arXiv:2607.03372}, 2026.

\bibitem[Liu et~al.(2023)Liu, Zhu, Gao, Feng, Liu, Zhu, and Stone]{liu2023libero}
Bo Liu, Yifeng Zhu, Chongkai Gao, Yihao Feng, Qiang Liu, Yuke Zhu, and Peter Stone.
\newblock {LIBERO: Benchmarking Knowledge Transfer for Lifelong Robot Learning}.
\newblock In \emph{Advances in Neural Information Processing Systems (NeurIPS) Datasets and Benchmarks Track}, 2023.

\bibitem[Liu et~al.(2025)Liu, Hamid, Xie, Lee, Du, and Finn]{liu2025bid}
Yuejiang Liu, Jubayer~Ibn Hamid, Annie Xie, Yoonho Lee, Maximilian Du, and Chelsea Finn.
\newblock {Bidirectional Decoding: Improving Action Chunking via Guided Test-Time Sampling}.
\newblock In \emph{International Conference on Learning Representations (ICLR)}, 2025.

\bibitem[Liu et~al.(2026)Liu, Huang, Hong, Du, Cao, Fei-Fei, and Wu]{liu2026t2mem}
Yize Liu, Huang Huang, Yining Hong, Zijian Du, Zhi Cao, Li Fei-Fei, and Jiajun Wu.
\newblock {T$^2$Mem: Learning Test-Time Memory for Robotics}.
\newblock \emph{arXiv preprint arXiv:2609.36720}, 2026.

\bibitem[Luo et~al.(2024)Luo, Xu, Liu, Tan, Lin, Wu, Abbeel, and Levine]{luo2024fmb}
Jianlan Luo, Charles Xu, Fangchen Liu, Liam Tan, Zipeng Lin, Jeffrey Wu, Pieter Abbeel, and Sergey Levine.
\newblock {FMB: a Functional Manipulation Benchmark for Generalizable Robotic Learning}.
\newblock \emph{The International Journal of Robotics Research}, 2024.

\bibitem[Ma et~al.(2025)Ma, Hejna, Wahid, Fu, Shah, Liang, Xu, Kirmani, Xu, Driess, Xiao, Tompson, Bastani, Jayaraman, Yu, Zhang, Sadigh, and Xia]{ma2025gvl}
Yecheng~Jason Ma, Joey Hejna, Ayzaan Wahid, Chuyuan Fu, Dhruv Shah, Jacky Liang, Zhuo Xu, Sean Kirmani, Peng Xu, Danny Driess, Ted Xiao, Jonathan Tompson, Osbert Bastani, Dinesh Jayaraman, Wenhao Yu, Tingnan Zhang, Dorsa Sadigh, and Fei Xia.
\newblock {Vision Language Models are In-Context Value Learners}.
\newblock In \emph{International Conference on Learning Representations (ICLR)}, 2025.

\bibitem[Mark et~al.(2026)Mark, Liang, Attarian, Fu, Dwibedi, Shah, and Kumar]{mark2026bpp}
Max~Sobol Mark, Jacky Liang, Maria Attarian, Chuyuan Fu, Debidatta Dwibedi, Dhruv Shah, and Aviral Kumar.
\newblock {BPP: Long-Context Robot Imitation Learning by Focusing on Key History Frames}.
\newblock \emph{arXiv preprint arXiv:2602.15010}, 2026.

\bibitem[Nasiriany et~al.(2022)Nasiriany, Gao, Mandlekar, and Zhu]{nasiriany2022sailor}
Soroush Nasiriany, Tian Gao, Ajay Mandlekar, and Yuke Zhu.
\newblock {Learning and Retrieval from Prior Data for Skill-based Imitation Learning}.
\newblock In \emph{Conference on Robot Learning (CoRL)}, 2022.

\bibitem[{Open X-Embodiment Collaboration}(2024)]{oxe2024}
{Open X-Embodiment Collaboration}.
\newblock {Open X-Embodiment: Robotic Learning Datasets and RT-X Models}.
\newblock In \emph{IEEE International Conference on Robotics and Automation (ICRA)}, 2024.

\bibitem[Otani et~al.(2020)Otani, Nakashima, Rahtu, and Heikkil{\"a}]{otani2020uncovering}
Mayu Otani, Yuta Nakashima, Esa Rahtu, and Janne Heikkil{\"a}.
\newblock {Uncovering Hidden Challenges in Query-Based Video Moment Retrieval}.
\newblock In \emph{British Machine Vision Conference (BMVC)}, 2020.

\bibitem[Pardo et~al.(2018)Pardo, Tavakoli, Levdik, and Kormushev]{pardo2018time}
Fabio Pardo, Arash Tavakoli, Vitaly Levdik, and Petar Kormushev.
\newblock {Time Limits in Reinforcement Learning}.
\newblock In \emph{International Conference on Machine Learning (ICML)}, pages 4042--4051, 2018.

\bibitem[Peng et~al.(2018)Peng, Abbeel, Levine, and van~de Panne]{peng2018deepmimic}
Xue~Bin Peng, Pieter Abbeel, Sergey Levine, and Michiel van~de Panne.
\newblock {DeepMimic: Example-Guided Deep Reinforcement Learning of Physics-Based Character Skills}.
\newblock \emph{ACM Transactions on Graphics}, 37\penalty0 (4), 2018.

\bibitem[{Physical Intelligence} et~al.(2025){Physical Intelligence}, Black, Brown, Darpinian, Dhabalia, Driess, Esmail, Equi, Finn, Fusai, Galliker, Ghosh, Groom, Hausman, Ichter, Jakubczak, Jones, Ke, LeBlanc, Levine, Li-Bell, Mothukuri, Nair, Pertsch, Ren, Shi, Smith, Springenberg, Stachowicz, Tanner, Vuong, Walke, Walling, Wang, Yu, and Zhilinsky]{black2025pi05}
{Physical Intelligence}, Kevin Black, Noah Brown, James Darpinian, Karan Dhabalia, Danny Driess, Adnan Esmail, Michael Equi, Chelsea Finn, Niccolo Fusai, Manuel~Y. Galliker, Dibya Ghosh, Lachy Groom, Karol Hausman, Brian Ichter, Szymon Jakubczak, Tim Jones, Liyiming Ke, Devin LeBlanc, Sergey Levine, Adrian Li-Bell, Mohith Mothukuri, Suraj Nair, Karl Pertsch, Allen~Z. Ren, Lucy~Xiaoyang Shi, Laura Smith, Jost~Tobias Springenberg, Kyle Stachowicz, James Tanner, Quan Vuong, Homer Walke, Anna Walling, Haohuan Wang, Lili Yu, and Ury Zhilinsky.
\newblock {$\pi_{0.5}$: a Vision-Language-Action Model with Open-World Generalization}.
\newblock In \emph{Conference on Robot Learning (CoRL)}, 2025.

\bibitem[Press et~al.(2022)Press, Smith, and Lewis]{press2022alibi}
Ofir Press, Noah~A. Smith, and Mike Lewis.
\newblock {Train Short, Test Long: Attention with Linear Biases Enables Input Length Extrapolation}.
\newblock In \emph{International Conference on Learning Representations (ICLR)}, 2022.

\bibitem[Pumacay et~al.(2024)Pumacay, Singh, Duan, Krishna, Thomason, and Fox]{pumacay2024colosseum}
Wilbert Pumacay, Ishika Singh, Jiafei Duan, Ranjay Krishna, Jesse Thomason, and Dieter Fox.
\newblock {THE COLOSSEUM: A Benchmark for Evaluating Generalization for Robotic Manipulation}.
\newblock In \emph{Robotics: Science and Systems (RSS)}, 2024.

\bibitem[Rathi(2026)]{rathi2026interference}
Soumil Rathi.
\newblock {RoboMME-Interference: Benchmarking Robot Memory Under Interference}.
\newblock \emph{arXiv preprint arXiv:2606.22338}, 2026.

\bibitem[Ruoss et~al.(2023)Ruoss, Del{\'e}tang, Genewein, Grau-Moya, Csord{\'a}s, Bennani, Legg, and Veness]{ruoss2023randomized}
Anian Ruoss, Gr{\'e}goire Del{\'e}tang, Tim Genewein, Jordi Grau-Moya, R{\'o}bert Csord{\'a}s, Mehdi Bennani, Shane Legg, and Joel Veness.
\newblock {Randomized Positional Encodings Boost Length Generalization of Transformers}.
\newblock In \emph{Annual Meeting of the Association for Computational Linguistics (ACL)}, 2023.

\bibitem[Shi et~al.(2026)Shi, Xie, Liu, Sun, Liu, Wang, Zhou, Fan, Zhang, and Huang]{shi2026memoryvla}
Hao Shi, Bin Xie, Yingfei Liu, Lin Sun, Fengrong Liu, Tiancai Wang, Erjin Zhou, Haoqiang Fan, Xiangyu Zhang, and Gao Huang.
\newblock {MemoryVLA: Perceptual-Cognitive Memory in Vision-Language-Action Models for Robotic Manipulation}.
\newblock In \emph{International Conference on Learning Representations (ICLR)}, 2026.

\bibitem[Shou et~al.(2021)Shou, Lei, Wang, Ghadiyaram, and Feiszli]{shou2021gebd}
Mike~Zheng Shou, Stan~Weixian Lei, Weiyao Wang, Deepti Ghadiyaram, and Matt Feiszli.
\newblock {Generic Event Boundary Detection: A Benchmark for Event Segmentation}.
\newblock In \emph{IEEE/CVF International Conference on Computer Vision (ICCV)}, 2021.

\bibitem[Song et~al.(2024)Song, Chai, Wang, Zhang, Zhou, Wu, Chi, Guo, Ye, Zhang, Lu, Hwang, and Wang]{song2024moviechat}
Enxin Song, Wenhao Chai, Guanhong Wang, Yucheng Zhang, Haoyang Zhou, Feiyang Wu, Haozhe Chi, Xun Guo, Tian Ye, Yanting Zhang, Yan Lu, Jenq-Neng Hwang, and Gaoang Wang.
\newblock {MovieChat: From Dense Token to Sparse Memory for Long Video Understanding}.
\newblock In \emph{IEEE/CVF Conference on Computer Vision and Pattern Recognition (CVPR)}, 2024.

\bibitem[Sridhar et~al.(2026)Sridhar, Pan, Sharma, and Finn]{sridhar2026memer}
Ajay Sridhar, Jennifer Pan, Satvik Sharma, and Chelsea Finn.
\newblock {Scaling up Memory for Robotic Control via Experience Retrieval}.
\newblock In \emph{International Conference on Learning Representations (ICLR)}, 2026.
\newblock arXiv:2510.20328 (as ``MemER: Scaling Up Memory for Robot Control via Experience Retrieval'').

\bibitem[Sun et~al.(2026)Sun, Gao, Chen, Wang, Liu, and Wang]{sun2026ltm}
Qixin Sun, WenLong Gao, Xiansheng Chen, Zhongyuan Wang, Si Liu, and Pengwei Wang.
\newblock {LTM: Latent Transition Based In-Context Memory for Vision-Language-Action Models}.
\newblock \emph{arXiv preprint arXiv:2610.21362}, 2026.

\bibitem[Sun et~al.(2025)Sun, Li, Dalal, Xu, Vikram, Zhang, Dubois, Chen, Wang, Koyejo, Hashimoto, and Guestrin]{sun2025ttt}
Yu Sun, Xinhao Li, Karan Dalal, Jiarui Xu, Arjun Vikram, Genghan Zhang, Yann Dubois, Xinlei Chen, Xiaolong Wang, Sanmi Koyejo, Tatsunori Hashimoto, and Carlos Guestrin.
\newblock {Learning to (Learn at Test Time): RNNs with Expressive Hidden States}.
\newblock In \emph{International Conference on Machine Learning (ICML)}, 2025.

\bibitem[Torne et~al.(2025)Torne, Tang, Liu, and Finn]{torne2025pasttoken}
Marcel Torne, Andy Tang, Yuejiang Liu, and Chelsea Finn.
\newblock {Learning Long-Context Diffusion Policies via Past-Token Prediction}.
\newblock In \emph{Conference on Robot Learning (CoRL)}, 2025.

\bibitem[Torne et~al.(2026)Torne, Pertsch, Walke, Vedder, Nair, Ichter, Ren, Wang, Tang, Stachowicz, Dhabalia, Equi, Vuong, Springenberg, Levine, Finn, and Driess]{torne2026mem}
Marcel Torne, Karl Pertsch, Homer Walke, Kyle Vedder, Suraj Nair, Brian Ichter, Allen~Z. Ren, Haohuan Wang, Jiaming Tang, Kyle Stachowicz, Karan Dhabalia, Michael Equi, Quan Vuong, Jost~Tobias Springenberg, Sergey Levine, Chelsea Finn, and Danny Driess.
\newblock {MEM: Multi-Scale Embodied Memory for Vision Language Action Models}.
\newblock \emph{arXiv preprint arXiv:2603.03596}, 2026.

\bibitem[Wang et~al.(2025)Wang, Zhou, Song, Huang, Shu, and Ma]{wang2025vlatest}
Zhijie Wang, Zhehua Zhou, Jiayang Song, Yuheng Huang, Zhan Shu, and Lei Ma.
\newblock {VLATest: Testing and Evaluating Vision-Language-Action Models for Robotic Manipulation}.
\newblock \emph{Proceedings of the ACM on Software Engineering (FSE)}, 2\penalty0 (FSE):\penalty0 1615--1638, 2025.

\bibitem[Wen et~al.(2020)Wen, Lin, Darrell, Jayaraman, and Gao]{wen2020copycat}
Chuan Wen, Jierui Lin, Trevor Darrell, Dinesh Jayaraman, and Yang Gao.
\newblock {Fighting Copycat Agents in Behavioral Cloning from Observation Histories}.
\newblock In \emph{Advances in Neural Information Processing Systems (NeurIPS)}, 2020.

\bibitem[Wu et~al.(2024)Wu, Jing, Cheang, Chen, Xu, Li, Liu, Li, and Kong]{wu2024gr1}
Hongtao Wu, Ya Jing, Chilam Cheang, Guangzeng Chen, Jiafeng Xu, Xinghang Li, Minghuan Liu, Hang Li, and Tao Kong.
\newblock {Unleashing Large-Scale Video Generative Pre-training for Visual Robot Manipulation}.
\newblock In \emph{International Conference on Learning Representations (ICLR)}, 2024.

\bibitem[Wu et~al.(2026)Wu, Shah, Huang, Sontakke, Bharadhwaj, Bajcsy, Simchowitz, and Shi]{wu2026imprint}
Yilin Wu, Rutav Shah, Shuaiyi Huang, Sumedh~A. Sontakke, Homanga Bharadhwaj, Andrea Bajcsy, Max Simchowitz, and Guanya Shi.
\newblock {Memory Without Curation: Learning Multi-Stream Latent Memory for Robot Policies}.
\newblock \emph{arXiv preprint arXiv:2610.14406}, 2026.

\bibitem[Yang et~al.(2026)Yang, Tu, Yang, Mao, Dong, Chen, Peng, Xiong, Cao, Dai, Zhou, Mu, and Wang]{yang2026eventvla}
Ganlin Yang, Zhangzheng Tu, Yuqiang Yang, Sitong Mao, Junyi Dong, Tianxing Chen, Jiaqi Peng, Jing Xiong, Jiafei Cao, Jifeng Dai, Wengang Zhou, Yao Mu, and Tai Wang.
\newblock {EventVLA: Event-Driven Visual Evidence Memory for Long-Horizon Vision-Language-Action Policies}.
\newblock \emph{arXiv preprint arXiv:2606.20092}, 2026.

\bibitem[Yin et~al.(2026)Yin, Xu, Yang, Tam, Liu, Yao, Zeng, Cui, Wang, Yin, and Lin]{yin2026simplememvla}
Cheng Yin, Wang Xu, Junpeng Yang, Sikyuen Tam, Hanyu Liu, Yuan Yao, Xiangrui Zeng, Junbo Cui, Yequan Wang, Zhouping Yin, and Yankai Lin.
\newblock {SimpleMemVLA: A Simple but Effective Native-Video Memory for Vision-Language-Action Models}.
\newblock \emph{arXiv preprint arXiv:2609.05533}, 2026.

\bibitem[Yuan et~al.(2021)Yuan, Lan, Wang, Chen, Wang, and Zhu]{yuan2021closer}
Yitian Yuan, Xiaohan Lan, Xin Wang, Long Chen, Zhi Wang, and Wenwu Zhu.
\newblock {A Closer Look at Temporal Sentence Grounding in Videos: Dataset and Metric}.
\newblock In \emph{ACM Multimedia Workshop on Human-centric Multimedia Analysis}, 2021.

\bibitem[Zacks et~al.(2007)Zacks, Speer, Swallow, Braver, and Reynolds]{zacks2007event}
Jeffrey~M. Zacks, Nicole~K. Speer, Khena~M. Swallow, Todd~S. Braver, and Jeremy~R. Reynolds.
\newblock {Event Perception: A Mind-Brain Perspective}.
\newblock \emph{Psychological Bulletin}, 133\penalty0 (2):\penalty0 273--293, 2007.

\bibitem[Zeng et~al.(2026)Zeng, Ye, Chen, Shentu, Wu, Yan, and Li]{zeng2026kemo}
Yihan Zeng, Minghao Ye, Yiyuan Chen, Yide Shentu, Philipp Wu, Zike Yan, and Zhongyu Li.
\newblock {KEMO: Event-Driven Keyframe Memory for Long-Horizon Robot Manipulation with VLA Policies}.
\newblock \emph{arXiv preprint arXiv:2606.23589}, 2026.

\bibitem[Zhao et~al.(2023)Zhao, Kumar, Levine, and Finn]{zhao2023act}
Tony~Z. Zhao, Vikash Kumar, Sergey Levine, and Chelsea Finn.
\newblock {Learning Fine-Grained Bimanual Manipulation with Low-Cost Hardware}.
\newblock In \emph{Robotics: Science and Systems (RSS)}, 2023.

\bibitem[Zhou et~al.(2025)Zhou, Xu, Tie, Chen, Zhang, Chu, Zhou, and Sun]{zhou2025liberopro}
Xueyang Zhou, Yangming Xu, Guiyao Tie, Yongchao Chen, Guowen Zhang, Duanfeng Chu, Pan Zhou, and Lichao Sun.
\newblock {LIBERO-PRO: Towards Robust and Fair Evaluation of Vision-Language-Action Models Beyond Memorization}.
\newblock \emph{arXiv preprint arXiv:2510.03827}, 2025.

\end{thebibliography}
}
\clearpage
\appendix
\begin{center}{\LARGE\bfseries Appendix}\end{center}
\vspace{0.5em}
\newcommand{\suppfit}[1]{\sbox0{#1}\typeout{SUPPFITW:\the\wd0}\ifdim\wd0>\linewidth\resizebox{\linewidth}{!}{\usebox0}\else\usebox0\fi}
\renewcommand{\thesection}{\Alph{section}}
\makeatletter\setlength{\@fptop}{0pt}\setlength{\@fpsep}{20pt plus 4pt}\setlength{\@dblfptop}{0pt}\setlength{\@dblfpsep}{20pt plus 4pt}\makeatother

This appendix follows the order of the main text.
\Cref{supp:tab:roadmap} maps each step of the argument to the section that backs it; \cref{supp:stats} gives the protocol and statistics shared by all sections, \cref{supp:prereg} collects the pre-registered predictions with their outcomes, and \cref{supp:cells} lists every cell of the four models analysed throughout with its interval and test.
Each section opens with its conclusion, and each table states its model and condition.

\begin{center}
\small\setlength{\tabcolsep}{3pt}
\begin{tabular}{@{}p{0.46\linewidth}p{0.51\linewidth}@{}}
\toprule
Main text & Appendix \tabularnewline
\midrule
\raggedright Null events cause early stops, through memory only (\cref{sec:phenomenon}) & \raggedright \ref{supp:robust}: dose, timing, natural stops, event validity; \ref{supp:tasks}: all 16 tasks \tabularnewline
\raggedright Time predicts progress in demonstrations (\cref{sec:data}) & \raggedright \ref{supp:realdata}: predictor, real datasets, pacing \tabularnewline
\raggedright The policy reads the clock (\cref{sec:clock}) & \raggedright \ref{supp:clock}: frames versus stamps, graded rescaling, press timing \tabularnewline
\raggedright Without the clock, the slots carry the failure (\cref{sec:noclock}) & \raggedright \ref{supp:noclock}: controls, press timing, probe, null-slot dose \tabularnewline
\raggedright Vulnerability tracks the index (\cref{sec:designs}) & \raggedright \ref{supp:designs}: SimpleMemVLA, LTM-$\pi$, all designs \tabularnewline
\raggedright Training on null events fails (\cref{sec:fixes}) & \raggedright \ref{supp:data}: pause-only, weaker dose, replay \tabularnewline
\raggedright The event index helps, within limits (\cref{sec:fixes}) & \raggedright \ref{supp:eventindex}: definition, audit, held-out, costs \tabularnewline
\raggedright Beyond RoboMME (\cref{sec:discussion}) & \raggedright \ref{supp:mikasa}: MIKASA-Robo, exploratory \tabularnewline
\bottomrule
\end{tabular}
\captionof{table}{\textbf{Where each claim is supported.} Letters are sections of this appendix.}
\label{supp:tab:roadmap}
\end{center}

\section{Protocol and statistics}
\label{supp:stats}

\paragraph{Episodes.}
Every cell uses the 50 test episodes of RoboMME for the task, with the benchmark's environment seeds and policy noise seed 7, so that all conditions of a model are paired episode by episode.
The required number of repetitions $K$ is set by the benchmark: on PickXtimes 16, 10, 12, 6 and 6 episodes have $K=1,\dots,5$; on SwingXtimes 15, 12 and 23 episodes have $K=1,2,3$.
Held-out confirmations use the benchmark's 50 validation episodes per task (other environment seeds, same difficulty mix).

\begin{figure*}[t]
\centering
\includegraphics[width=\linewidth]{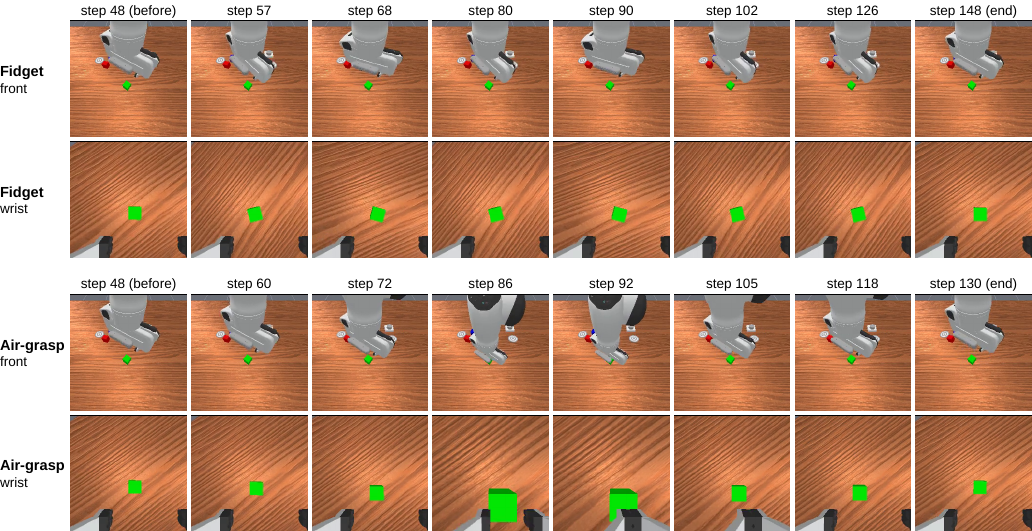}
\caption{\textbf{A fidget and an air-grasp, frame by frame} (SwingXtimes episode 18, released FrameSamp-Modul, visible condition; front and wrist cameras, unedited frames).
Both events start after policy step 48, before the robot has picked up the cube, and the policy is not queried while they run.
The fidget rolls the wrist back and forth around its held target, which the wrist camera shows as a rotating scene; the air-grasp lowers the gripper to the cube, closes it on air, and returns to the starting pose.
Neither moves the cube (0.00\,cm in the evaluator's log), and both end where they started, yet every frame enters memory.
The two rows come from separate rollouts of the same episode; a pause, which leaves the image still, is shown in \cref{fig:phenomenon}.}
\label{supp:fig:film}
\end{figure*}

\paragraph{Null events.}
Events are injected at action-chunk boundaries after policy steps 48 and 160 (other times in \cref{supp:robust}).
During a pause the robot holds its last joint target and the policy is not queried; during a fidget the wrist rolls $\pm0.3$\,rad around the held target; an air-grasp is executed by the motion planner above a resting object and returns to the pre-event pose.
After each event the evaluator checks and logs that no object moved by more than 1\,cm and no button was pressed.
On tasks whose scene draws fading highlights (gripper trail, button glow), the highlight clock is frozen during the pause so that the images after the pause match those before it; on RouteStick, the evaluator's gripper trace is not extended during the pause.
In the hidden condition, the event is executed identically but its frames are not written to memory and the time stamps do not advance over it.
\Cref{supp:fig:film} shows a fidget and an air-grasp frame by frame.

\paragraph{Failure classes.}
From the end-of-episode record we classify a failure as \emph{early} (the policy pressed the button or submitted at a stage before the $K$-th repetition was complete), \emph{extra} (it failed at or after the stopping stage, or timed out, having completed $K$ repetitions), or \emph{other}.
Some runs of the released model predate stage logging; their failures are listed as unclassified (Uncl.) and are not used for any statement about failure direction, unless a stage-logged rerun reproduces the cell episode by episode (\cref{supp:robust}).

\paragraph{Tests.}
All comparisons are paired on episodes and use the exact two-sided McNemar test on the discordant pairs.
\Cref{supp:cells} lists every cell of the four models analysed throughout (released FrameSamp-Modul, \clk{}, \clk{} with the event index, \nope{}) with Wilson 95\% intervals and Holm-corrected $p$-values over these 42 comparisons; every effect there that the main text calls significant remains significant after correction.
\Cref{supp:clock,supp:eventindex} use a broader core family of 59 pre-specified tests (rescaling, route split, designs, event index) and state which effects survive it.
Where a test was pre-registered, the prediction and its rule were logged before the run; \cref{supp:prereg} lists them with their outcomes.

\paragraph{Training variance and a second machine.}
Each from-scratch model was trained with seed 42; \clk{} and \nope{} were also trained with seed 43 (below).
Two further training runs of the \clk{} recipe with the same seed, which differ only through GPU non-determinism, and one on a second machine bound the training-level variation (\cref{supp:tab:reruns}): the pause damage reproduces within 2 episodes, whereas the no-event and hidden cells on PickXtimes vary by up to 12 and 9 episodes.
Intermediate checkpoints of rerun 1 also show that the reliance on the clock grows with the ability to count: from 10k to 30k training steps, success without events rises from 28 to 44 on PickXtimes, while success with time stamps scaled by 1.5 stays at 23--27, and failures under a 0.5 scale turn from early stops into extra repetitions.
The pause-augmentation model and its control are trained and evaluated on a second machine (H200 GPUs of a different variant, different driver and CPU).
Running the released model there reproduces the cluster's results within 3 episodes (PickXtimes 44 against 47, SwingXtimes 46 against 44) with 95 of 100 episodes agreeing.

\begin{wraptable}{r}{0.62\linewidth}
\typeout{WRAPLABEL:supp:tab:reruns}%
\centering\small\setlength{\tabcolsep}{4pt}
\suppfit{\begin{tabular}{lcccccc}
\toprule
& \multicolumn{3}{c}{PickXtimes} & \multicolumn{3}{c}{SwingXtimes} \\
\cmidrule(lr){2-4}\cmidrule(lr){5-7}
Run & None & Pause & Hidden & None & Pause & Hidden \\
\midrule
\clk{} (reported) & 41 & 17 & 38 & 48 & 10 & 38 \\
Rerun 1 & 44 & 16 & 44 & 48 & 12 & 43 \\
Rerun 2 & 40 & -- & 43 & -- & -- & 41 \\
Second machine & 32 & 16 & 35 & 46 & 11 & 40 \\
\bottomrule
\end{tabular}}
\caption{\textbf{Training reruns of \clk{}} (same recipe and seed). Successes of 50; pauses of 5\,s.}
\label{supp:tab:reruns}
\end{wraptable}
\paragraph{A second training seed.}
\label{supp:seed}
We trained \clk{} and \nope{} again with seed 43, on the second machine, with the recipe unchanged, and compare them with each other (same seed, same machine).
Rules, written before training: \nope{} counts as well (no event at least that of \clk{} $-\,6$, both tasks); it does not read the stamps ($\times1.5$ at least its own no event $-\,6$, PickXtimes); pauses still hurt it (visible at most its own hidden $-\,8$, both tasks); pauses hurt \clk{} (visible at most its own no event $-\,15$, both tasks); a fast clock causes early stops in \clk{} ($\times1.5$ at most its own no event $-\,10$, at least 75\% of failures early). A miss would be reported and the claim narrowed. %
All five hold (\cref{tab:supp-seed}): \nope{} 39 and 47 against \clk{} 37 and 44; $\times1.5$ 38 against 39; visible 24 and 15 against hidden 38 and 43; \clk{} 37 to 17 and 44 to 15; \clk{} $\times1.5$ 27 against 37 ($-11/{+}1$, $p{=}.006$), 22 of 23 failures early, at the threshold.
Not registered: on SwingXtimes, among episodes that succeed without the event and stop early with it, \nope{}'s presses with $K{=}3$ under pauses come after two swings in 5 of 8 (2 after one), \clk{}'s after one in 10 of 12, a contrast in the same direction as with seed 42 (14 of 14 and 19 of 21) but less clean. Counting all early presses: 6 of 9 and 10 of 14 (seed 42: 17 of 17 and 21 of 23); presses after one, 2 of 9 against 10 of 14, Fisher $p{=}.04$.

\begin{table}[t]
\centering\small\setlength{\tabcolsep}{3.5pt}
\suppfit{\begin{tabular}{lccccccc}
\toprule
& \multicolumn{4}{c}{PickXtimes} & \multicolumn{3}{c}{SwingXtimes} \\
\cmidrule(lr){2-5}\cmidrule(lr){6-8}
Model, seed & None & $\times1.5$ & Pause & Hidden & None & Pause & Hidden \\
\midrule
\clk{}, 42 & 41 & 19 & 17 & 38 & 48 & 10 & 38 \\
\clk{}, 43 & 37 & 27 & 17 & 38 & 44 & 15 & 35 \\
\nope{}, 42 & 42 & 44 & 21 & 44 & 47 & 14 & 46 \\
\nope{}, 43 & 39 & 38 & 24 & 38 & 47 & 15 & 43 \\
\bottomrule
\end{tabular}}
\caption{\textbf{Second training seed} of \clk{} and \nope{}. Seed 43 is trained and evaluated on the second machine. Successes of 50; pauses of 5\,s.}
\label{tab:supp-seed}
\end{table}

\section{Robustness of the phenomenon}
\label{supp:robust}
The early stops of \cref{sec:phenomenon} do not hinge on one length, time or kind of stop: the damage sets in once a pause lasts 0.5--1.5\,s, appears wherever a decision to stop still lies ahead, and follows from many short random stops as well.
Where a fidget nudges a held cube, the damage is the same as where it moves nothing; the one event that interferes with the task, the air-grasp on PickXtimes, is excluded from the main text.

\Cref{supp:tab:robust} varies how long the null event lasts, when it happens, and whether it is one designed event or many short, randomly timed stops.
All cells use the released FrameSamp-Modul and are paired with the same 50 episodes without an event.
Failure classes for the 5-second pauses come from a rerun with stage logging that reproduces the cell episode by episode (16 and 10 successes; no episode changes outcome): 34 of 34 failures on PickXtimes and 36 of 40 on SwingXtimes are early.

\begin{table}[b]
\centering\small\setlength{\tabcolsep}{3pt}
\suppfit{\begin{tabular}{lcccc}
\toprule
& \multicolumn{2}{c}{PickXtimes} & \multicolumn{2}{c}{SwingXtimes} \\
\cmidrule(lr){2-3}\cmidrule(lr){4-5}
Condition & Succ. & vs none & Succ. & vs none \\
\midrule
No event & 47 & -- & 44 & -- \\
\midrule
Two pauses of 0.5\,s & 43 & $-$5/+1, .22 & 45 & $-$1/+2, 1 \\
Two pauses of 1.5\,s & 24 & $-$23/+0, $<10^{-6}$ & 21 & $-$25/+2, $<10^{-5}$ \\
Two pauses of 5\,s & 16 & $-$31/+0, $<10^{-9}$ & 10 & $-$34/+0, $<10^{-9}$ \\
\quad hidden & 45 & $-$3/+1, .62 & 47 & $-$2/+5, .45 \\
Two pauses of 15\,s & 16 & $-$31/+0, $<10^{-9}$ & 12 & $-$32/+0, $<10^{-9}$ \\
\quad hidden & 44 & $-$4/+1, .38 & 46 & $-$2/+4, .69 \\
\midrule
One 5\,s pause at step 48 & 18 & $-$29/+0, $<10^{-8}$ & 28 & $-$18/+2, $<10^{-3}$ \\
One 5\,s pause at step 300 & 26 & $-$21/+0, $<10^{-6}$ & 36 & $-$8/+0, .008 \\
\midrule
Random micro-stops & 20 & $-$27/+0, $<10^{-7}$ & 17 & $-$29/+2, $<10^{-6}$ \\
\quad hidden & 46 & $-$2/+1, 1 & 45 & $-$1/+2, 1 \\
\bottomrule
\end{tabular}}
\caption{\textbf{Dose, timing, and natural stops} (released FrameSamp-Modul). ``vs none'': episodes lost/gained relative to no event and exact McNemar $p$. Random micro-stops: at each chunk boundary the robot stops with probability 0.3 for 5--40 steps, until a budget of 300 steps; episodes usually end first, after about 7 stops and 157 stopped steps.}
\label{supp:tab:robust}
\end{table}

\paragraph{Dose.}
The damage sets in between 0.5 and 1.5\,s per pause and saturates by 5\,s: two 15-second pauses do no more harm than two 5-second ones, and hiding either restores success.
Saturation is expected if the policy stops at the first decision point after its estimate of progress overshoots; a longer pause cannot make it stop earlier than that.

\paragraph{Timing.}
A single pause hurts wherever it falls, but only in episodes that still have a decision to make after it.
With the pause at step 48, $K=1$ episodes are untouched (16 of 16 on PickXtimes), whereas only 2 of 34 episodes with $K\geq2$ succeed.
With the pause at step 300, PickXtimes episodes with $K\leq2$, which are nearly finished by then, are untouched (25 of 26), whereas those with $K\geq3$ fail (1 of 24).
This rules out an immediate reflex to stop after any pause and matches a stop decision taken when the remembered elapsed time reaches its usual value.

\paragraph{Natural stops.}
Many short stops at random times, as a safety layer or a slow planner might produce, are nearly as harmful as two long pauses, and again only through memory: hidden, they leave success at the no-event level.
They are also a boundary of the event index (\cref{supp:eventindex}).

\paragraph{The PickXtimes air-grasp.}
On PickXtimes the air-grasp that returns to its starting pose leaves 12 of 50 (43 hidden), but its failures do not have the clock signature: they concentrate at $K=1$ (2 of 16 succeed, against 16 of 16 for every other event), where no early stop is possible.
The second air-grasp often happens just after the cube has been released, and the returning open gripper encloses the cube; the policy then grasps it again.
The event is therefore not null on this task and is excluded from the main text; on SwingXtimes the same event behaves as the other null events.

\paragraph{Null-event validity.}
\label{supp:validity}
Every fidget and air-grasp is logged with the largest object displacement and the button state; a pause holds the current joint targets, toward which the arm keeps settling for a median of 14 (PickXtimes) and 16 (SwingXtimes) steps, at most 23, after which no arm joint moves by more than $10^{-5}$\,rad per step (\cref{supp:hidstatic}). The policies do not pause on their own: in 850 rollouts without injected events (PickXtimes and SwingXtimes), the arm never stays below $10^{-4}$\,rad per step for more than 2 consecutive steps, so every null event studied here is injected.
 Air-grasps moved an object by more than 1\,cm in at most 8 of 50 SwingXtimes episodes (none on PickXtimes). A wrist fidget performed while a cube is held can nudge it beyond 1\,cm: in 17--33 of 50 episodes per cell, at similar rates with the frames visible and hidden, because the physical event is the same. We keep these episodes, since the paired comparison holds the physical event fixed. \Cref{tab:supp-validity} shows that the damage does not depend on them.

\begin{table*}[t]\centering\small
\suppfit{%
\begin{tabular}{@{}llcccc@{}}\toprule
 & & \multicolumn{2}{c}{Moved nothing in either} & \multicolumn{2}{c}{Nudged a cube in both} \\
Model & Task & Visible / hidden ($n$) & $p$ & Visible / hidden ($n$) & $p$ \\\midrule
FrameSamp-Modul & PickXtimes & 6 / 17 (17) & .001 & 1 / 8 (10) & .016 \\
 & SwingXtimes & 3 / 11 (11) & .008 & 5 / 17 (20) & .002 \\
TokenDrop-Modul & PickXtimes & 10 / 17 (21) & .016 & 6 / 12 (14) & .031 \\
 & SwingXtimes & 4 / 10 (12) & .031 & 4 / 15 (24) & .001 \\
\nope{} & PickXtimes & 5 / 12 (15) & .016 & 8 / 10 (11) & .50 \\\bottomrule
\end{tabular}}
\caption{\textbf{Fidget damage by event validity}: successes with the fidget visible and hidden among episodes in which the fidget moved no object in either condition, or nudged a cube in both (exact McNemar). SimpleMemVLA: \cref{supp:smv}.}
\label{tab:supp-validity}
\end{table*}

\subsection{Hiding a pause only once the arm is still}
\label{supp:hidstatic}
On SwingXtimes, hiding a pause does not fully restore \clk{} (38 against 48 without an event) or LTM-$\pi$ (31 against 50; \cref{supp:ltm}).
The hidden condition drops every frame of the hold, including the arm's settling motion at its start, which on SwingXtimes interrupts carrying the cube.
\emph{Prediction fixed before the run:} if dropping that motion explains the gap, a condition that sends the hold frames while the arm still moves and hides the rest should reach at least the no-event success minus 3 on SwingXtimes for both models (\clk{} $\geq45$, LTM-$\pi$ $\geq47$), and match the hidden condition on PickXtimes.
The arm counts as still from the first frame at which no arm joint moves by $10^{-5}$\,rad and neither finger by $10^{-5}$\,m; the threshold, about twice the largest per-step change of a settled arm and below every step of 430\,k steps of normal motion, was set from existing joint logs before the run.
It keeps 5--16 of the 100 frames of each hold.
\begin{wraptable}{r}{0.62\linewidth}
\typeout{WRAPLABEL:tab:supp-hidstatic}%
\setlength{\tabcolsep}{4pt}
\suppfit{\begin{tabular}{@{}llcccc@{}}\toprule
 & & No event & Hidden & \begin{tabular}[b]{@{}c@{}}Hidden\\once still\end{tabular} & Visible \\\midrule
\clk & PickXtimes & 41 & 38 & 37 & 17 \\
 & SwingXtimes & 48 & 38 & 27 & 10 \\
LTM-$\pi$ & PickXtimes & 50 & 50 & 49 & 7 \\
 & SwingXtimes & 50 & 31 & 34 & 9 \\\bottomrule
\end{tabular}}
\caption{\textbf{Pauses hidden once the arm is still} (100-step pauses, 50 paired episodes).}
\label{tab:supp-hidstatic}
\end{wraptable}
The prediction fails for both models: on SwingXtimes success reaches 27 (\clk) and 34 (LTM-$\pi$), on PickXtimes it matches the hidden condition (37 and 49).
The settling frames are thus not what the hidden condition is missing; sending them instead turns failures towards early presses (\clk{} 17 of 23, LTM-$\pi$ 11 of 16), as if a few frames of a still interval were already read as progress.
What else separates hidden pauses from no event on SwingXtimes for these two models is open, so all their SwingXtimes comparisons are made against no event, where the visible damage is larger still.

\section{Null events across all 16 RoboMME tasks}
\label{supp:tasks}

\Cref{tab:supp_tasks} applies the protocol of \cref{sec:phenomenon} to every RoboMME task: released FrameSamp-Modul, two 15-second pauses (300 steps, after policy steps 48 and 160), visible or hidden, 50 paired episodes per cell.
Two tasks at floor were not run; tasks with fewer than 15 no-event successes are reported but not interpreted (rule fixed before the runs).

\begin{table*}[t]
\centering
\small
\setlength{\tabcolsep}{3pt}
\suppfit{%
\begin{tabular}{llrcccrl}
\toprule
Suite & Task & Video & No event & Visible & Hidden & \multicolumn{2}{c}{Visible vs hidden} \\
 & & (frames) & & & & lost/gained & $p$ \\
\midrule
\multirow{4}{*}{Counting}
 & BinFill          & --  & 23 & 9  & 21 & $-12/{+0}$ & $<10^{-3}$ \\ %
 & PickXtimes       & --  & 47 & 16 & 44 & $-28/{+0}$ & $<10^{-8}$ \\ %
 & SwingXtimes      & --  & 44 & 12 & 46 & $-36/{+2}$ & $<10^{-8}$ \\ %
 & StopCube         & --  & 24 & 2 & 0 & $-0/{+2}$ & .50 \\ %
\midrule
\multirow{4}{*}{Permanence}
 & VideoUnmask     & 66  & 17 & 16 & 19 & $-3/{+0}$ & .25 \\ %
 & VideoUnmaskSwap$^{\circ}$ & 114 & 14 & 12 & 14 & $-2/{+0}$ & .50 \\ %
 & ButtonUnmask$^{\circ}$    & --  & 11 & 7 & 13 & $-10/{+4}$ & .18 \\ %
 & ButtonUnmaskSwap & --  & 6  & \multicolumn{4}{c}{floor, not run} \\ %
\midrule
\multirow{4}{*}{Reference}
 & PickHighlight$^{\circ}$   & --  & 12 & 13 & 14 & $-7/{+6}$ & 1 \\ %
 & VideoRepick$^{\circ}$     & 281 & 13 & 5 & 15 & $-10/{+0}$ & .002 \\ %
 & VideoPlaceButton$^{\dagger}$ & 754 & 29 & 11 & 29 & $-18/{+0}$ & $<10^{-5}$ \\ %
 & VideoPlaceOrder  & 894 & 18 & 15 & 21 & $-8/{+2}$ & .11 \\ %
\midrule
\multirow{4}{*}{Imitation}
 & MoveCube         & 242 & 39 & 33 & 39 & $-9/{+3}$  & .15 \\ %
 & InsertPeg        & 238 & 5  & \multicolumn{4}{c}{floor, not run} \\ %
 & PatternLock$^{\dagger}$   & 96  & 28 & 19 & 28 & $-9/{+0}$  & .004 \\ %
 & RouteStick$^{\ddagger}$   & 150 & 33 & 7  & 36 & $-29/{+0}$ & $<10^{-8}$ \\ %
\bottomrule
\end{tabular}}
\caption{\textbf{Two 15-second pauses across the 16 RoboMME tasks} (released FrameSamp-Modul, successes of 50 paired episodes).
Video: median length of the demonstration video preceding execution, which shares the 32 memory slots.
$^{\dagger}$Highlight timers (fading after 40 steps) are shifted across the pause so that it does not erase them.
$^{\ddagger}$As $\dagger$; trajectory points recorded during the pause are excluded from the success check.
$^{\circ}$Not interpreted ($<$15 no-event successes).
Only the first pause occurs on RouteStick; on PatternLock it occurs in 34 of 50 episodes.}
\label{tab:supp_tasks}
\end{table*}

\paragraph{Counting tasks: through the clock.}
PickXtimes and SwingXtimes are analysed in \cref{sec:clock,sec:noclock}.
BinFill, which also ends with a button press, behaves the same: 9 visible against 21 hidden.
Advancing only the time stamps over the pause reproduces the damage (time only 7 against 23 without an event, $-16/{+0}$, $p<10^{-4}$); storing only the null frames does not significantly change success (content only 19, $-5/{+1}$, $p{=}.22$).
Rescaling all time stamps does not significantly change BinFill (23 to 18 at $\times0.5$, $p{=}.23$; 23 at $\times1.5$), so here the role of the clock rests on the pause decomposition alone.

\paragraph{Video-conditioned tasks: through eviction of the video.}
Nine tasks open with a demonstration video sampled into the same 32 uniformly spaced slots as the execution frames, so a visible pause pushes video frames out of memory (RouteStick: from about 24 slots to about 10).
\emph{Starvation} injects no pause but gives the video only the slots it keeps under the visible pause; \emph{protection} injects the visible pause but keeps the video's no-event slot count.
Starvation reproduces the visible damage and protection restores the hidden level (starved and protected against visible and hidden):
RouteStick 4 and 32 against 7 and 36;
PatternLock 19 and 28 against 19 and 28;
VideoPlaceButton 14 and 27 against 11 and 29.
Starved never differs significantly from visible ($p\geq.37$), nor protected from hidden ($p\geq.29$); protection beats the visible pause on every task ($p\leq.012$).
The time stamps play no measurable role here: protection restores success although the stamps still advance over the pause, and on RouteStick rescaling them leaves success unchanged (33 at $\times0.5$, 35 at $\times1.5$, against 33).
On MoveCube the drop under eviction is small and not significant (33 against 39 hidden, $p{=}.15$).
Of the remaining tasks, two meet the 15-success threshold besides StopCube, and the pre-registered prediction misses on both.
Harm was predicted to be largest where key information is shown only briefly at the start, and to shrink with video length; instead the short-video VideoUnmask loses only 3 (16 against 19, $p{=}.25$), and the long-video VideoPlaceOrder 6 (15 against 21, $p{=}.11$), neither significant.
A post hoc reading: a pause can only take away what memory contributes, and on these two tasks the released policy barely beats the same backbone without memory ($\pi_{0.5}$: 14 and 14, against 17 and 18).
Where memory matters most, VideoRepick ($\pi_{0.5}$: 0 of 50), the visible pause cuts success from 15 to 5 ($p{=}.002$), but this task is below the threshold and is reported only.

\paragraph{StopCube: time is the task.}
In StopCube elapsed time is legitimate information: in 98 of 100 demonstrations the robot hovers still above the button, for a median of 198 steps, before pressing.
Rescaling the time stamps in either direction lowers success from 24 to 6 ($\times0.5$, $-23/{+5}$, $p<10^{-3}$) and 5 ($\times1.5$, $-22/{+3}$, $p<10^{-3}$).
Because hiding a pause also removes real elapsed time, our prediction, recorded before the run, is the reverse of the other tasks: hiding should not help, and may hurt.
The prediction holds only trivially: both conditions collapse (2 visible, 0 hidden, against 24 without an event), so the comparison is at floor and we do not interpret it; whatever the memory does, a 15-second interruption leaves StopCube nearly unsolvable.

\section{Time predicts progress in demonstrations}
\label{supp:realdata}
In scripted simulation and in real teleoperation alike, the time step alone predicts how many repetitions a demonstration has completed.
This section defines the predictor, lists the data and bounds what the evidence shows.

\paragraph{Predictor.}
At each step $t$ from the first repetition through the stop decision, the label is the number of repetitions completed by $t$.
The predictor sees only $t$ and outputs the most frequent training-fold label in the same 10-step bin (the overall most frequent label for unseen bins); neither $K$ nor the task is given.
Accuracy uses 5-fold cross-validation over demonstrations; the baseline predicts the most frequent label. %
On RoboMME, repetition boundaries come from the recorded sub-goal text; on real data, a gripper cycle (flips under 0.25\,s removed; a closure of at least 0.5\,s followed by a release; the window ends at the last release). %
On RoboMME the two definitions agree (gripper-cycle count equals $K$ in 100 of 100 PickXtimes demonstrations; accuracy 0.894 against 0.890). %
Bins aligned to $\approx$14 per stage instead of 10 frames give results within 0.006. %

\begin{table*}[t]
\centering
\small
\setlength{\tabcolsep}{3pt}
\suppfit{\begin{tabular}{llccc}
\toprule
Dataset & Source & Demos & Time only & Majority \\
\midrule
RoboMME PickXtimes ($K$ 1--5) & sim., scripted & 100 & 0.890 & 0.467 \\ %
RoboMME SwingXtimes ($K$ 1--3) & sim., scripted & 100 & 0.802 & 0.423 \\ %
RoboMME PickXtimes, $K{=}3$ only & sim., scripted & 27 & 0.939 & 0.428 \\ %
\midrule
austin\_sailor (3 cycles) & real, teleop. & 90 & 0.922 & 0.354 \\ %
stanford\_hydra, dishes (3-cycle demos) & real, teleop. & 86 & 0.933 & 0.366 \\ %
stanford\_hydra, dishes (all) & real, teleop. & 110 & 0.824 & 0.309 \\ %
stanford\_hydra, 3 tasks (all) & real, teleop. & 570 & 0.617 & 0.389 \\ %
FMB~\cite{luo2024fmb} (annotated primitives) & real, teleop. & 1804 & 0.747 & 0.341 \\ %
LIBERO-10~\cite{liu2023libero}, two-object tasks & sim., teleop. & 207 & 0.916 & 0.524 \\ %
\bottomrule
\end{tabular}}
\caption{\textbf{Elapsed time alone predicts progress}: accuracy of predicting the number of completed repetitions from the time step, 5-fold cross-validation over demonstrations. At matched $K$, real and scripted data reach the same accuracy (0.92--0.93 against 0.94). Accuracy drops when irregular demonstrations (regrasps, variable object counts) are kept (hydra, all) and for fine-grained primitives (FMB), but stays well above the majority baseline.}
\label{tab:supp-realdata}
\end{table*}

\paragraph{Pacing and idle stretches.}
Within a task and stage, the coefficient of variation of stage durations is 0.11--0.13 in the real datasets and 0.08--0.13 on RoboMME. %
End-effector stillness ($<0.5$\,mm per frame) of at least 2\,s occurs in 0\% (austin\_sailor), 1.8\% (stanford\_hydra) and 1.1\% (LIBERO-10) of demonstrations; 1 of 100 RoboMME PickXtimes demonstrations contains a still stretch (12 steps). %
FMB has one $\approx$1.5\,s hold per demonstration, always at the same phase (grasp settling), so it is part of the regular pace. %

\paragraph{Scope of this evidence.}
The correlation the shortcut exploits is as strong in regular (three-cycle) real teleoperation demonstrations as in RoboMME, and weaker across all tasks (62\% on stanford\_hydra, 75\% on FMB); this does not show that a policy trained on real data uses it, nor that a real robot fails under null events: all causal evidence in this paper comes from simulation.
The real datasets are small (e.g.\ 30 demonstrations per task in austin\_sailor), and public versions may have had idle frames trimmed during conversion, making them look more regular than raw collection; the statistics describe the data that VLA training actually consumes. %

\section{The clock route in detail}
\label{supp:clock}
A time-stamped policy reads a null event partly through its stamps.
On PickXtimes, advancing the stamps over a pause of 5\,s or more without storing its frames reproduces the full damage, rescaling them flips the failure even when their layout stays in distribution, and where the stamps carry the null interval and the two accounts of the press differ, presses follow the clock.

All manipulations are applied at test time to the released policies (FrameSamp-Modul unless noted); 50 paired episodes per cell, exact two-sided McNemar test against the same model without event.
Two null events of $L$ steps follow policy steps 48 and 160 (\cref{sec:setup}); $L{=}30,100,300$ is 1.5, 5 and 15\,s.
Failure types: \textbf{e} early (press before the $K$-th repetition); \textbf{x} extra (failure after $K$ repetitions or time-out, as in \cref{supp:stats}, and on SwingXtimes also a put-down failure without a press, which the tables of \cref{supp:cells} list as other); \textbf{u} unclassified (no stage log).

\subsection{Splitting a null event into frames and time stamps}
\label{supp:clock:route}

A null event puts its frames into the 32 slots and advances the time stamps of all later frames.
In all four conditions the robot pauses physically.
\emph{Visible}: null frames stored, stamps advance.
\emph{Time only}: null frames withheld, stamps as in visible.
\emph{Content only}: null frames stored with the last working stamp.
\emph{Hidden}: null frames withheld, stamps equal work time.
Content only is not stamp-clean: at $L{=}300$ about half of the slots share two repeated stamps and working frames are 2.4$\times$ further apart in stamp than normal, a layout never seen in training.

\begin{table*}[t]
\centering
\small
\setlength{\tabcolsep}{2.5pt}
\suppfit{\begin{tabular}{llccccc}
\toprule
Task & Event & No event & Visible & Time only & Content only & Hidden \\
\midrule
\multicolumn{7}{l}{\emph{FrameSamp-Modul}} \\
Pick & pause $2{\times}30$ & 47 & 24 u & 34 (16e) & 37 (13e) & -- \\
Pick & pause $2{\times}100$ & 47 & 16 u & 17 (32e, 1x) & 25 (25e) & 45 (5e) \\
Pick & pause $2{\times}300$ & 47 & 16 u & 14 u & 24 u\,/\,16 (28e, 6x)$^\ddagger$ & 44 u \\
Swing & pause $2{\times}30$ & 44 & 21 u & 43 (3e, 4x) & 23 (12e, 14x) & -- \\
Swing & pause $2{\times}100$ & 44 & 10 u & 28 (17e, 5x) & 27 (13e, 10x) & 47 (1e, 2x) \\
Swing & pause $2{\times}300$ & 44 & 12 u & 23 u & 18 (25e, 7x) & 46 u \\
\midrule
\multicolumn{7}{l}{\emph{FrameSamp-Expert}} \\
Pick & pause $2{\times}300$ & 44 & 14 (35e, 1x) & 4 (46e) & 20 (18e, 12x) & 45 (5e) \\
\bottomrule
\end{tabular}}
\caption{\textbf{Route split} (successes of 50; failure types in brackets).
All non-hidden cells differ from no event at $p<.01$ except Swing $2{\times}30$ time only ($p{=}1$). No hidden cell differs significantly from no event.
$^\ddagger$Two runs of the same condition (original 24; stage-logged rerun 16; $-10/{+}2$, $p{=}.039$ between runs; cause not identified).}
\label{tab:supp-route}
\end{table*}

On Pick the stamps alone reproduce the full damage at $L{=}100$ and $300$ (17 and 14 against 16; for FrameSamp-Expert 4 against 14), the frames alone part of it; at $L{=}30$ each channel does about half.
On Swing the damage at $L{=}30$ goes entirely through the frames; at $L{=}100$ and $300$ neither channel alone reaches the visible damage, so the two add up.

\subsection{In-distribution manipulations of the clock}
\label{supp:clock:warp}

\paragraph{Graded rescaling.}
\Cref{tab:supp-warp} extends \cref{fig:clock} to all scales we ran (stamps of all memory frames multiplied by $s$; no event, no frame dropped).
FrameSamp-Expert, which reads the same memory through a different pathway, flips the same way on PickXtimes: under $\times0.5$ its failures are mostly extra repetitions (14 of 16), under $\times1.5$ almost all early presses (17 of 18; \cref{tab:supp_warp}).

\begin{table}[t]
\centering
\small
\setlength{\tabcolsep}{3pt}
\suppfit{\begin{tabular}{lcccccc}
\toprule
 & \multicolumn{3}{c}{PickXtimes} & \multicolumn{3}{c}{SwingXtimes} \\
\cmidrule(lr){2-4}\cmidrule(lr){5-7}
$s$ & Succ. & $p$ & e\,/\,x & Succ. & $p$ & e\,/\,x \\
\midrule
0.5 & 24 & $<10^{-6}$ & 3\,/\,23 & 20 & $<10^{-6}$ & 1\,/\,26$^\dagger$ \\
0.6 & 33 & .001 & 1\,/\,16 & 32 & .004 & 4\,/\,14 \\
0.7 & 40 & .065 & 2\,/\,8 & 40 & .29 & 3\,/\,7 \\
0.75 & 45 & .69 & u & 48 & .22 & u \\
1 & 47 & -- & -- & 44 & -- & -- \\
1.1 & 38 & .012 & 12\,/\,0 & 48 & .22 & 2\,/\,0 \\
1.25 & 30 & $<10^{-4}$ & 20\,/\,0 & 43 & 1.0 & 5\,/\,2 \\
1.5 & 18 & $<10^{-8}$ & 32\,/\,0 & 37 & .092 & 13\,/\,0 \\
\bottomrule
\end{tabular}}
\caption{\textbf{Graded clock rescaling} (FrameSamp-Modul, no event). $p$: McNemar against $s{=}1$.
Holm over the 59-test core family: Pick $\times$0.5, 0.6, 1.25, 1.5, Swing $\times$0.5 and the Pick direction flip survive; Pick $\times$1.1 (adjusted $p{=}.21$) and Swing $\times$0.6 ($.10$) do not; $\times$0.7 was not in the family.
$^\dagger$From the stage-logged rerun of the same condition (21/50).}
\label{tab:supp-warp}
\end{table}

\begin{table}[b]
\centering
\small
\setlength{\tabcolsep}{3pt}
\suppfit{%
\begin{tabular}{llcllr}
\toprule
Memory & Task & No event & Stamps & Succ. (early / extra) & $p$ \\
\midrule
FrameSamp-Modul & P & 47 & $\times0.5$ & 24 (3 / 23) & $<10^{-6}$ \\ %
 & & & $\times1.5$ & 18 (32 / 0) & $<10^{-8}$ \\ %
FrameSamp-Expert & P & 44 & $\times0.5$ & 34 (2 / 14) & .021 \\ %
 & & & $\times1.25$ & 36 & .021 \\ %
 & & & $\times1.5$ & 32 (17 / 0) & $<10^{-3}$ \\ %
 & S & 49 & $\times0.5$ & 31 (0 / 2) & $<10^{-4}$ \\ %
TokenDrop-Modul & P & 39 & $\times0.5$ & 35 (4 / 11) & .34 \\ %
 & & & $\times1.5$ & 35 (13 / 2) & .29 \\ %
\bottomrule
\end{tabular}}
\caption{\textbf{Rescaling the time stamps of released FrameSamp-Expert and TokenDrop}, no null event (successes of 50 paired episodes; $p$: exact McNemar against no event).
Failures not classified as early or extra are not listed.
TokenDrop's failure direction differs between the two scales (Fisher $p{=}.003$, $.063$ after Holm correction over the family of clock tests).}
\label{tab:supp_warp}
\end{table}

On Pick a slower clock causes extra repetitions and a faster one early presses (flip $\times$0.5 vs $\times$1.5: 23\,x/3\,e against 0\,x/32\,e, Fisher $p<10^{-4}$).
$\times$0.5 keeps every stamp inside the training range, so this is not an extrapolation artefact.
The effect is asymmetric: $\times$0.75 is harmless, $\times$1.1 is not, with its early presses at large $K$ ($K{=}2$: 1 of 10; $K{=}3$: 5/12; $K{=}4$: 2/6; $K{=}5$: 4/6), as expected if the policy presses once the clock passes a threshold near its usual press time.

\paragraph{Frozen clock.}
Once 32 frames are stored, we overwrite their stamps with $\mathrm{linspace}(0,T,32)$, the layout of a training episode of length $T$, whatever the true step.
The stamp pattern is then in distribution, but the clock always reads about $T/139$ repetitions on Pick (139 steps per repetition).
A policy that counts by the clock should press early when $K$ is below this reading and repeat too often when $K$ is above it; it does (\cref{tab:supp-fixedpos}).
On Swing the totals are 20, 32 and 17 at $T{=}150,300,600$, with the same flips.

\begin{wraptable}{r}{0.43\linewidth}
\typeout{WRAPLABEL:tab:supp-fixedpos}%
\centering
\small
\setlength{\tabcolsep}{3pt}
\suppfit{\begin{tabular}{lcccc}
\toprule
 & $n$ & $T{=}150$ & $T{=}300$ & $T{=}600$ \\
 & & ($\approx$1.1 rep.) & ($\approx$2.2) & ($\approx$4.3) \\
\midrule
Pick $K{=}1$ & 16 & 16 s & 15 s, 1e & 10 s, \textbf{6e} \\
Pick $K{=}2$ & 10 & 2 s, 1e, \textbf{7x} & 1 s, \textbf{9e} & 1 s, \textbf{9e} \\
Pick $K{=}3$ & 12 & \textbf{12x} & \textbf{12x} & \textbf{12e} \\
Pick $K{=}4$ & 6 & \textbf{6x} & \textbf{6x} & \textbf{6e} \\
Pick $K{=}5$ & 6 & \textbf{6x} & \textbf{6x} & 3e, 3x \\
Pick total & 50 & 18 & 16 & 11 \\
\bottomrule
\end{tabular}}
\caption{\textbf{Frozen clock}, PickXtimes (FrameSamp-Modul, no event; s = success). Unmanipulated success is 47 (Swing 44); every frozen cell on both tasks is worse ($p<.01$). Failures are extra repetitions where $K$ exceeds the clock reading and early presses where $K$ is below it.}
\label{tab:supp-fixedpos}
\end{wraptable}

\subsection{When does the policy press?}
\label{supp:clock:press}

\paragraph{Definition.}
We take the episodes that succeed without the event and fail early with it.
With $t_0$ the press step in the no-event run of the same episode, $t$ the press step with the event (null steps excluded) and $n_{\text{null}}{=}2L$, the \emph{work ratio} is $t/t_0$ and the \emph{elapsed ratio} $(t+n_{\text{null}})/t_0$.
\emph{Clock}: the policy presses when elapsed time reaches its usual press time (elapsed ratio ${\approx}1$, somewhat below if the threshold lies below the usual press time, as $\times$1.1 suggests).
\emph{Skip one}: the null events of an episode together count as one repetition whatever their duration (work ratio ${\approx}(K{-}1)/K$, elapsed ratio ${\approx}(K{-}1)/K + n_{\text{null}}/t_0$).
A press can only follow a completed repetition, so both also predict the repetitions $j$ completed at the press: skip one $j{=}K{-}1$; clock the first $j$ with $\bar t_0(j)+n_{\text{null}} \ge \bar t_0(K)$, where $\bar t_0(j)$ is the median no-event press time for $j$ repetitions (Pick: 279, 424, 579, 729, 852 steps for $j{=}1..5$).

\begin{wrapfigure}{r}{0.5\linewidth}
\typeout{WRAPLABEL:fig:press}%
\centering
\includegraphics[width=\linewidth]{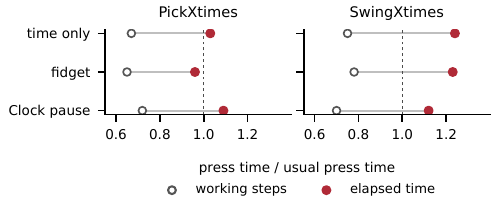}
\caption{\textbf{Press timing relative to the same episode without the event.} Median press time relative to the same episode without the event, counted in working steps (open) or in elapsed time including the null steps (filled), over episodes that succeed without the event and stop early with it. Time-stamped models only. On SwingXtimes the elapsed ratio exceeds 1 because a press can only follow a completed swing.}
\label{fig:press}
\end{wrapfigure}
\paragraph{At $L{=}100$ the two accounts nearly coincide.}
\Cref{tab:supp-press100} gives the ratios of \cref{fig:press} by $K$.
On Pick, skip one predicts 0.97--1.04 at every $K$, as does the clock.
At $K{=}2$ an early press can only follow the first repetition, which fixes the elapsed ratio at $(\bar t_0(1)+200)/\bar t_0(2)\approx1.13$ under any account.
Written with the measured press times, skip one predicts 1.13, 1.08, 1.07, 1.09 for $K{=}2..5$, as close to the data as the clock at $K{=}2,3$ and further only at $K{=}4,5$ ($n{=}2$--6).
The $L{=}100$ ratios on Pick thus do not discriminate between the accounts.

\begin{table*}[t]
\centering
\small
\setlength{\tabcolsep}{3pt}
\suppfit{\begin{tabular}{lcccccc}
\toprule
PickXtimes, $L{=}100$ & $n$ & work & elapsed & $K{=}2$ & $K{=}3$ & $K{=}4$ \,/\, $5$ \\
\midrule
FS time only & 29 & 0.67 & 1.03 & 1.13 (8) & 1.03 (11) & 0.88 (6) / 0.94 (4) \\
FS fidget & 31 & 0.65 & 0.96 & 1.14 (10) & 0.88 (11) & 0.90 (6) / 0.94 (4) \\
\clk{} pause & 24 & 0.72 & 1.09 & 1.14 (8) & 1.07 (12) & 1.01 (2) / 0.96 (2) \\
\midrule
\emph{Skip one}, $\frac{K-1}{K}+\frac{n_{\text{null}}}{t_0}$ & & & & 0.97--0.98 & 1.01 & 1.01--1.02 / 1.03--1.04 \\
\emph{Clock} & & & & 1 & 1 & 1 / 1 \\
\midrule
SwingXtimes, $L{=}100$ & $n$ & work & elapsed & $K{=}1$ & $K{=}2$ & $K{=}3$ \\
\midrule
FS time only$^\ast$ & 15 & 0.75 & 1.24 & -- & 1.25 (6) & 1.20 (9) \\
FS fidget$^\ast$ & 30 & 0.78 & 1.23 & -- & 1.30 (10) & 1.15 (20) \\
\clk{} pause & 35 & 0.70 & 1.12 & 1.80 (2) & 1.25 (12) & 1.08 (21) \\
\midrule
\emph{Skip one} & & & & 0.61--0.72 & 0.97--1.00 & 1.08--1.09 \\
\bottomrule
\end{tabular}}
\caption{\textbf{Press timing at $L{=}100$}: medians; per-$K$ medians with $n$ in brackets. FS = released FrameSamp-Modul. $^\ast$The released Swing no-event run has no client log; its press times are taken from the run with two 10-step pauses (outcomes agree in 47 of 50 episodes).}
\label{tab:supp-press100}
\end{table*}

\paragraph{Where the accounts differ, the stamps follow the clock.}
With 600 null steps on Pick, or 200 on Swing (a swing takes 70--100 steps), the clock is past the usual press time at the first decision and predicts a press after the first repetition for every $K$, whereas skip one predicts $K{-}1$.
\Cref{tab:supp-reps} lists every stage-logged cell where the predictions differ ($j$ read from the end-of-episode sub-goal).
Where the stamps carry the null interval (visible, fidget or time only on a time-stamped model), presses come at or before the clock prediction in 95 of 107 episodes; the exception is FrameSamp-Modul time only on Swing (4 of 9).
Where only the frames carry it (content only on the released model; \nope, \cref{supp:noclock:press}), presses on Swing skip exactly one repetition (17 of 20) and on Pick fall between the two predictions.

\begin{table*}[t]
\centering
\small
\setlength{\tabcolsep}{3pt}
\suppfit{\begin{tabular}{lllccc}
\toprule
Model & Cell & $K$ & $j$ at press ($j$:\,count) & Skip & Clock \\
\midrule
\multicolumn{6}{l}{\emph{Stamps carry the null interval}} \\
FS-Expert & Pick, visible $2{\times}300$ & 3\,/\,4\,/\,5 & 1:11 \,/\, 1:4 \,/\, 1:3 & 2\,/\,3\,/\,4 & 1\,/\,1\,/\,2 \\
FS-Expert & Pick, time only $2{\times}300$ & 3\,/\,4\,/\,5 & 0:3 1:8 \,/\, 0:2 1:1 \,/\, 0:2 1:1 & 2\,/\,3\,/\,4 & 1\,/\,1\,/\,2 \\
FS-Expert & Swing, visible $2{\times}300$ & 3 & 0:3 1:19 & 2 & 1 \\
\clk & Swing, visible $2{\times}100$ & 3 & 0:1 1:19 2:1 & 2 & 1 \\
FS-Modul & Swing, fidget $2{\times}100$ & 3 & 1:14 2:6 & 2 & 1 \\
FS-Modul & Swing, time only $2{\times}100$ & 3 & 1:4 2:5 & 2 & 1 \\
\midrule
\multicolumn{6}{l}{\emph{Only the frames carry the null interval}} \\
FS-Modul & Swing, content only $2{\times}100$ & 3 & 0:1 2:7 & 2 & 1 \\
FS-Modul & Swing, content only $2{\times}300$ & 3 & 1:2 2:10 & 2 & 1 \\
FS-Modul & Pick, content only $2{\times}300$ & 3\,/\,4\,/\,5 & 1:2 2:7 \,/\, 1:2 2:4 \,/\, 3:3 4:1 & 2\,/\,3\,/\,4 & 1\,/\,1\,/\,1 \\
\bottomrule
\end{tabular}}
\caption{\textbf{Repetitions completed before the early press} ($j$), in every stage-logged cell where the clock account (threshold at the usual press time) and the skip-one account predict different $j$. Episodes that succeed without the event and stop early with it. Post hoc analysis.}
\label{tab:supp-reps}
\end{table*}

On the released FrameSamp-Modul at $L{=}300$, which has no stage log, the median step of the failure (on Pick a press in 1486 of 1490 stage-logged cases) is 268--346 at every $K{=}2..5$ for visible and time-only pauses, near one repetition (279) rather than $K{-}1$ (579, 729 for $K{=}4,5$); this reading is post hoc.
At $L{=}30$, where the null interval is much shorter than a repetition, the early presses on Pick also come after $K{-}1$ repetitions (time only: 14 of 14 for $K\geq2$; post hoc). The clock account explains this only if the press threshold lies well before the usual press time, as the $\times1.1$ rescaling also implies: it advances the clock at the decision after two repetitions by only about 40 steps, yet causes early presses there (5 of 5 for $K{=}3$). The pre-registered dose test below decides between the two accounts.

\paragraph{Dose test on \clk{} (pre-registered).}
\clk{} runs on both tasks with $L{=}30$ and $300$, analysed as in \cref{tab:supp-press100}.
Criteria fixed before the run:
\emph{clock} if the median elapsed ratio at $L{=}30$ lies in 0.85--1.20 and the median work ratio at $L{=}300$ is $\le 0.55$;
\emph{skip one} if the median elapsed ratio is $<0.85$ at $L{=}30$ and $\ge1.5$ at $L{=}300$, with a work ratio near $(K{-}1)/K$.
Support requires both doses on the clock side; one each is partial; cells with fewer than 8 early failures are not interpreted.

\begin{table*}[t]\centering\small
\suppfit{%
\begin{tabular}{@{}llccccc@{}}\toprule
Task & $L$ & Succ. & $n$ early & Work & Elapsed & $j$ for $K{\geq}3$ \\\midrule
PickXtimes & 30 & 25 & 15 & 0.75 & 0.88 & $K{-}1$: 9 of 10 \\
 & 300 & 15 & 24 & 0.57 & 1.57 & $\leq1$: 13 of 15; $K{-}1$: 2 \\
SwingXtimes & 30 & 20 & 23 & 0.87 & 0.99 & $K{-}1$: 11 of 13 \\
 & 300 & 6 & 36 & 0.66 & 1.96 & $\leq1$: 21 of 21 \\\bottomrule
\end{tabular}}
\caption{\textbf{Dose test on \clk{}} (50 episodes paired with the no-event run; no-event success 41 and 48). Work and elapsed: median ratios over episodes that succeed without the event and stop early with it. $j$: repetitions completed before the press.}
\label{tab:supp-dose}
\end{table*}

By the registered rule the result is \emph{partial} (\cref{tab:supp-dose}): at $L{=}30$ both tasks fall on the clock side (elapsed 0.88 and 0.99), but at $L{=}300$ the work ratio misses the clock criterion (0.57 and 0.66) and matches the skip-one ratios.
We did not foresee that the ratios cannot separate the two accounts at $L{=}300$: no press can occur during a pause, and a 300-step pause outlasts the remaining work, so both accounts predict elapsed ratios well above 1; and the earliest press the clock account allows, after finishing the repetition in progress, lies above a work ratio of 0.55.
The repetitions completed before the press do separate them when $K{\geq}3$ (clock: at most one, since the clock has passed the usual press time by the end of the second pause; skip one: $K{-}1$).
34 of 36 such presses come after at most one repetition (all 21 on SwingXtimes), and 2 after $K{-}1$; \nope{} under the shorter pause, by contrast, presses after $K{-}1$ swings in all 14 cases (\cref{supp:noclock:press}).
At $L{=}30$ both accounts allow $K{-}1$, which is what we see (20 of 23), so this dose does not discriminate (see the $L{=}30$ paragraph above).
Because the repetition count was not the registered criterion, we repeat the test on held-out validation episodes with it as the only criterion, fixed before the run (clock if at least 75\% of presses with $K{\geq}3$ come after at most one repetition, skip one if at least 75\% come after $K{-}1$). On held-out episodes the outcome is clock: all 29 presses with $K{\geq}3$ come after one repetition (PickXtimes 15, SwingXtimes 14), none after $K{-}1$ (no-event success 38 and 44, with $2{\times}300$ pauses 10 and 14; 40 of 40 and 31 of 36 failures early).

\subsection{Further controls}
\label{supp:clock:falsify}

Hidden null events (the robot still pauses) leave Pick at 43--46 and Swing at 45--47, against 47 and 44 (all n.s.): the damage is not physical.
A single 100-step pause harms only where a stop decision follows it: after step 48, 16 of 16 $K{=}1$ episodes succeed; after step 300, 9 of 10 $K{=}2$ but 0 of 12 $K{=}3$.
Two 10-step pauses leave 43 and 45 (n.s.).

\section{What a model without time stamps makes of a null event}
\label{supp:noclock}
Without time stamps, a pause still hurts through the slots it fills.
The damage needs the null frames in memory but not what they show, grows with the number of null slots, and keeps the signature of early presses.

\begin{table}[t]
\centering
\small
\setlength{\tabcolsep}{4pt}
\suppfit{\begin{tabular}{lcccc}
\toprule
 & \multicolumn{2}{c}{PickXtimes} & \multicolumn{2}{c}{SwingXtimes} \\
\cmidrule(lr){2-3}\cmidrule(lr){4-5}
 & \clk & \nope & \clk & \nope \\
\midrule
No event & 41 & 42 & 48 & 47 \\
Time stamps $\times1.5$ & 19 & 44 & -- & -- \\
\midrule
Pause, visible & 17 & 21 & 10 & 14 \\
Pause, hidden & 38 & 44 & 38 & 46 \\
Pause, null frames swapped & -- & 17 & -- & 13 \\
\bottomrule
\end{tabular}}
\caption{\textbf{\nope{} read-out on both tasks} (PickXtimes panel of \cref{fig:twist}a).
Two models trained from scratch, identical except that \nope{} has no time encoding.
\nope{} ignores rescaled time stamps and counts as well as \clk, yet pauses hurt it as much.
Swapping in the last frame before the first pause leaves the damage unchanged; on SwingXtimes hiding does not fully restore \clk{} (\cref{supp:hidstatic}).
Successes of 50 paired episodes.}
\label{tab:noclock}
\end{table}

This section details the controls behind \cref{sec:noclock}.
\clk{} and \nope{} are trained from the $\pi_{0.5}$ base checkpoint for 30k steps with the released recipe; their configurations differ in one line (\texttt{use\_pos\_emb}).
Tests and failure types are as in \cref{supp:clock}.

\subsection{What \nope{} can see: slot order, not time stamps}
\label{supp:noclock:rope}

The memory has 32 slots of 16 tokens.
Slot $j$ holds stored frame $\lfloor jt/31\rfloor$, where $t$ is the newest stored frame; before step 32 the memory is shorter and padded on the right, in training and at test time.
Each token is encoded as $\mathrm{Linear}([\,\text{image token};\ \mathrm{silu}(W_p\,\tau)\,])$, where $\tau$ is the sinusoidal encoding of the frame's time step; \nope{} drops the $\tau$ branch.
The action expert reads memory through cross-attention with rotary position embeddings on both sides: memory keys get positions $0,\dots,511$ in slot order, action queries $512,\dots$; masked tokens leave the softmax without shifting the others' positions.
Both models thus know slot order, but under uniform sampling order carries no absolute time (slot $j$ is step $jt/31$ for any $t$).
Without $\tau$, elapsed time is available only through the frames themselves: how much the scene changes between neighbouring slots, and how many slots show a still scene.

\subsection{Slot layouts of the controls}
\label{supp:noclock:layout}

\begin{table}[t]
\centering
\small
\setlength{\tabcolsep}{2.5pt}
\suppfit{\begin{tabular}{p{2.2cm}p{4.2cm}c}
\toprule
\raggedright Control & \raggedright Slot contents (in slot order) & Real \tabularnewline
\midrule
\raggedright Hidden pause & \raggedright real frames, uniform over $[0,t]$ & 32\,/\,32\,/\,32 \tabularnewline
\raggedright Visible pause $2{\times}100$ & \raggedright all frames, uniform over $[0,t{+}200]$; slots falling in a pause show null frames & 20\,/\,22\,/\,25 \tabularnewline
\raggedright Null swap & \raggedright as visible; every null slot shows frame 48 (last frame before the first pause, gripper empty) & 20\,/\,22\,/\,25 \tabularnewline
\raggedright Duplicate fill (no pause) & \raggedright the real frames that the visible pause keeps, each repeated in place (about $1.5\times$) to fill all 32 slots & 20\,/\,22\,/\,25 \tabularnewline
\raggedright Null block $m$ (no pause) & \raggedright $32{-}2m$ real frames, uniform over $[0,t]$; $m$ consecutive copies of the frame at each event time (steps 48, 160), in time order & $32{-}2m$ \tabularnewline
\bottomrule
\end{tabular}}
\caption{\textbf{Memory layouts}, with distinct real frames at $t{=}300/450/600$ working steps. For \clk{}, frames keep their true stamps in the visible pause (the stamps advance over the pause) and their working-time stamps in duplicate fill (duplicates keep their own); null swap was run on \nope{} only.}
\label{tab:supp-layout}
\end{table}

Null swap keeps the number, position and stillness of the null slots and changes what they show.
Duplicate fill shows the same real frames, with the same spacing, as the visible pause, but no null frames.

\subsection{Results}
\label{supp:noclock:results}

\begin{table*}[t]
\centering
\small
\setlength{\tabcolsep}{2.5pt}
\suppfit{\begin{tabular}{lcccc}
\toprule
 & \multicolumn{2}{c}{PickXtimes} & \multicolumn{2}{c}{SwingXtimes} \\
\cmidrule(lr){2-3}\cmidrule(lr){4-5}
 & \clk & \nope & \clk & \nope \\
\midrule
No event & 41 (9e) & 42 (5e, 3x) & 48 (2x) & 47 (1e, 2x) \\
Stamps $\times1.5$ & 19 (28e, 2x) & 44 (3e, 3x) & -- & -- \\
\midrule
Pause $2{\times}100$, visible & 17 (33e) & 21 (21e, 8x) & 10 (37e, 3x) & 14 (26e, 10x) \\
Pause $2{\times}300$, visible & -- & 22 (19e, 9x) & -- & -- \\
Pause $2{\times}100$, hidden & 38 (12e) & 44 (4e, 2x) & 38 (3e, 9x) & 46 (1e, 3x) \\
Null swap & -- & 17 (21e, 12x) & -- & 13 (32e, 5x) \\
Fidget $2{\times}100$ & 16 (32e) & 23 (21e, 6x); hid.\ 45 & -- & -- \\
Duplicate fill & 39 (11e) & 38 (11e, 1x) & -- & 48 (1e, 1x) \\
\bottomrule
\end{tabular}}
\caption{\textbf{\clk{} and \nope{} under memory controls} (successes of 50; failure types in brackets; hid.: the same event hidden from memory). Against no event, every visible pause, swap and fidget cell is worse ($p<10^{-4}$); \nope{} $\times1.5$, \nope{} hidden and all duplicate-fill cells are not. \clk{} hidden is worse on Swing (38 vs 48, $p{=}.002$); two same-seed retrainings of \clk{} reach 43 and 41 there. --: not in this table; \clk{} under $2{\times}300$ is in \cref{tab:supp-dose} and \nope{} under $2{\times}300$ on SwingXtimes in \cref{supp:noclock:press}.}
\label{tab:supp-noclock}
\end{table*}

\paragraph{Paired contrasts.}
\nope{} ignores the stamps: $\times1.5$ gives 44 vs 42 ($p{=}.62$), and keeping the null frames but setting the stamps to working time gives 22 vs 21 visible.
What the null frames show does not matter: null swap vs visible, Pick 17 vs 21 ($-7/{+}3$, $p{=}.34$), Swing 13 vs 14 ($-4/{+}3$, $p{=}1$); both remain far below hidden ($-27/{+}0$ and $-33/{+}0$).
Sparser real history is harmless when the memory is full: duplicate fill does not differ from no event on either model ($p\ge.29$).

\subsection{When \nope{} presses}
\label{supp:noclock:press}

\begin{table}[t]
\centering
\small
\setlength{\tabcolsep}{3pt}
\suppfit{\begin{tabular}{lccccc}
\toprule
Cell & $n$ & work & elapsed & $j$ at press, $K{=}3$ & Clock\,/\,skip \\
\midrule
\multicolumn{6}{l}{\emph{\nope, visible pause or null swap}} \\
Pick, pause $2{\times}100$ & 17 & 0.74 & 1.13 & 2:10 & 2\,/\,2 \\
Pick, null swap & 16 & 0.76 & 1.15 & 2:8 & 2\,/\,2 \\
Pick, pause $2{\times}300$ & 15 & 0.74 & 1.74 & 1:4\; 2:3 & 1\,/\,2 \\
Swing, pause $2{\times}100$ & 23 & 0.84 & 1.28 & 2:14 & 1\,/\,2 \\
Swing, null swap & 29 & 0.85 & 1.28 & 1:2\; 2:16 & 1\,/\,2 \\
Swing, pause $2{\times}300$ & 16 & 0.85 & 2.26 & 1:4\; 2:1 & 1\,/\,2 \\
Pick, one pause of 100 & 13 & 0.76 & 0.98 & 2:5 & 3\,/\,2 \\
Swing, one pause of 100 & 20 & 0.81 & 1.03 & 2:9 & 2\,/\,2 \\
Pick, $2{\times}300$, held-out & 28 & 0.67 & 1.79 & 1:6\; 2:6 & 1\,/\,2 \\
Swing, $2{\times}300$, held-out & 11 & 0.92 & 2.21 & 0:1\; 1:2\; 2:4 & 1\,/\,2 \\
\multicolumn{6}{l}{\emph{\clk, for contrast}} \\
Swing, pause $2{\times}100$ & 35 & 0.70 & 1.12 & 0:1\; 1:19\; 2:1 & 1\,/\,2 \\
\bottomrule
\end{tabular}}
\caption{\textbf{Press timing of \nope{}} (definitions in \cref{supp:clock:press}). Medians over all $K$; $j$: repetitions completed at the press ($j$:\,count) for $K{=}3$, with the predictions of the clock account (from each model's no-event press times) and of skip one.}
\label{tab:supp-noclock-press}
\end{table}

After a visible pause on Pick, \nope{} presses at 0.74 of its usual working steps and 1.13 of its usual elapsed time.
This does not show that it reads elapsed time: at $L{=}100$ on Pick the clock and skip-one accounts predict the same repetition count (\cref{supp:clock:press}), and \nope{} matches both.
Where the two differ, \nope{} follows skip one: on Swing $K{=}3$ it presses after two swings in 14 of 14 episodes (16 of 18 with null swap), where the clock account predicts one and \clk{} presses after one in 19 of 21; with a second training seed the same contrast is weaker (\nope{} 5 of 8 after two, \clk{} 10 of 12 after one, same selection; \cref{supp:seed}).
To see how much one interval counts for, we ran one 5-second pause after step 48 with outcomes fixed in advance: one pause alone reads as about one repetition if on Swing at least 8 early presses with $K{=}3$ occur and at least 75\% come after two swings; only the pair does if success stays within 6 of no event or fewer than 8 such presses occur.
All 10 come after two swings (all $K{=}3$ early presses; 9 of 9 in the selection of the table), so one pause alone reads as about one repetition; on Pick all 5 do too, below the count the rule requires. Success falls from 42 to 27 and from 47 to 25 ($p{=}.001$ and $p<10^{-5}$).
On Swing the clock and skip-one accounts agree here (\cref{tab:supp-noclock-press}), so this cell measures the size of the effect, not which account holds.
A second pause adds failures (Pick 27 to 21, $p{=}.15$; Swing 25 to 14, $p{=}.03$), and on Swing makes the press more frequent (10 to 17 of the 23 episodes with $K{=}3$), but not earlier (all 17 after two swings).
With $2{\times}300$ pauses on test episodes its few $K{=}3$ early presses split on Pick (4 after one repetition, 3 after two; $n{=}7$) and lean to one swing on Swing (4 of 5).
That Swing run was pre-registered (length-independent if $\geq$75\% of $K{=}3$ early presses come after two swings, duration-carrying if $\geq$75\% after one, no verdict below $n{=}8$); with $n{=}5$ it gives no verdict.
We repeated it on the held-out episodes, Pick and Swing, with the same rule. All early presses with $K{=}3$: Pick $n{=}13$, 7 after two and 6 after one; Swing $n{=}9$, 6 after two, 2 after one, 1 before the first. Neither reaches 75\%, so the outcome is open on both tasks, with enough presses this time (in the selection of the table, 6 of 12 and 4 of 7 after two). Success falls from 46 to 16 and from 46 to 19 ($p<10^{-7}$ both); \clk{} on the same episodes presses after one in 29 of 29 with $K{\geq}3$.
What does change with length is the kind of failure, not the success rate (Pick 21 at $2{\times}100$, 22 at $2{\times}300$; Swing 14 and 13): on Swing, early presses fall from 26 to 16 of the failures and other failures rise from 10 to 21, whereas \clk{} stays early (37 of 40 failures at $2{\times}100$, 38 of 44 at $2{\times}300$).
Whether \nope{} reads the length of a pause is thus open: on test episodes longer pauses add no early presses, and at $2{\times}300$ a part of the presses, not all as with \clk{}, comes one repetition sooner than at $2{\times}100$. A plausible reading of that part, not tested, is that at $2{\times}300$ the null frames take most of the 32 slots and crowd out the frames of completed swings. The explicit clock, by contrast, carries the duration.
A pre-registered prediction that \nope{}'s press ratio falls with the length of the null interval missed (0.60 at two virtual 100-step pauses, 0.71 at 300); it was tested with the real frames of the visible pause packed against slot 31 and the null slots left empty, a left-padded layout never seen in training.

\subsection{Linear probe of the memory tokens}
\label{supp:noclock:probe}

A linear probe tests whether elapsed time or progress can be read from \nope{}'s memory tokens, and whether a pause inflates the reading.
\emph{Design}: \clk{} and \nope{}; training demonstrations of both tasks, one sample every 10 steps from step 64; memory of 32 frames uniform over $[0,s]$, encoded by each model's own memory encoder and averaged per slot.
Features: slot features in slot order (PCA to 32 dimensions per slot), or their mean and s.d.\ over slots.
Targets: elapsed steps, repetitions completed; ridge regression, 5-fold cross-validation by episode.
Pause test: two still 100-frame pauses inserted before execution frames 100 and 250, same working step.
\emph{Criteria fixed before the first run}: time is readable if held-out $R^2\ge0.8$ from \nope{}'s ordered features; pauses inflate it if \nope{}'s predicted elapsed time rises by at least 50\% of the 200 null steps and its predicted repetitions by more than 0.3 (\clk{} is the positive control); otherwise we report only the causal controls.
\begin{table*}[t]\centering\small
\suppfit{%
\begin{tabular}{@{}llcccc@{}}\toprule
 & & \multicolumn{2}{c}{PickXtimes} & \multicolumn{2}{c}{SwingXtimes} \\
 & & \clk & \nope & \clk & \nope \\\midrule
Held-out $R^2$, elapsed & ordered & 0.999 & 0.919 & 0.999 & 0.891 \\
 & set & 0.999 & 0.782 & 0.999 & 0.841 \\
Held-out $R^2$, repetitions & ordered & 0.958 & 0.927 & 0.955 & 0.920 \\
Pause shift, elapsed (\% of 200 null steps) & ordered & 99.7 & 56.4 & 98.6 & 27.3 \\
 & set & 99.1 & 38.1 & 98.0 & 15.3 \\
Pause shift, repetitions & ordered & $+1.38$ & $+0.94$ & $+1.31$ & $+0.53$ \\
 & set & $+1.31$ & $+0.50$ & $+1.29$ & $+0.47$ \\\bottomrule
\end{tabular}}
\caption{\textbf{Linear probe of the memory tokens}, 100 training demonstrations per task (4776 and 3747 samples; 2876 and 1847 late enough for the pause test). Shifts are medians over held-out samples of (reading with two 100-step pauses) $-$ (reading without), at the same working step.}
\label{tab:supp-probe}
\end{table*}

\emph{Result} (\cref{tab:supp-probe}).
Time stays linearly readable without time stamps: \nope{}'s ordered features give held-out $R^2$ 0.92 and 0.89 for elapsed steps (criterion met on both tasks), from the content and slot order of the memory alone.
\clk{}, the positive control, reads all 200 null steps as elapsed time and about 1.3 extra repetitions.
For \nope{} the pauses push the reading up on both tasks (repetitions $+0.94$ and $+0.53$, above the 0.3 criterion), but the elapsed-time shift meets the 50\% criterion only on PickXtimes (56\%); on SwingXtimes it is 27\%.
By the rule fixed in advance, the pause-inflation criterion is met on PickXtimes and not on SwingXtimes, so we draw no conclusion from the inflation test.
Two limits: the repetition target is correlated with elapsed time in demonstrations, so a probe for one partly reads the other; and a linear read-out shows what the memory encodes, not what the action expert uses, which the causal controls above address.

\subsection{Number of null slots}
\label{supp:noclock:dose}

Null block (\cref{tab:supp-layout}) varies the number of slots that show nothing happening while keeping the time span of the memory and the event times of the visible pause; no pause takes place.
It was pre-registered with two outcomes: an occupancy dose (success non-increasing over $m{=}1,3,6$ on both tasks, and $m{=}6$ at least 8 episodes below $m{=}1$ with McNemar $p<.05$ on one task and lower on the other) or a block read as one event ($m{=}1$ doing at least 75\% of the damage of $m{=}6$ on both tasks, with $m{=}6$ against $m{=}1$ not significant on either).
\Cref{tab:supp-nullblock} gives an occupancy dose on both tasks.
Two null slots cost nothing; twelve, about as many as a visible $2{\times}100$ pause leaves at $t{=}300$, cost about as much as the pause on PickXtimes (24 against 21) and most of it on SwingXtimes (23 against 14).
The failures keep the signature of the visible pause: at $m{=}6$, 15 of 26 failures on PickXtimes and 23 of 27 on SwingXtimes are early presses, and every early press at $K{=}3$ comes after two repetitions (7 of 7 and 16 of 16, in episodes that succeed without it), as under the visible pause (14 of 14 on SwingXtimes; \cref{supp:noclock:press}).

\begin{wraptable}{r}{0.59\linewidth}
\typeout{WRAPLABEL:tab:supp-nullblock}%
\centering
\small
\setlength{\tabcolsep}{3pt}
\suppfit{\begin{tabular}{@{}lcccc@{}}
\toprule
& \multicolumn{2}{c}{PickXtimes} & \multicolumn{2}{c}{SwingXtimes} \\
\cmidrule(lr){2-3}\cmidrule(lr){4-5}
Null slots & Succ. & vs none & Succ. & vs none \\
\midrule
0 (no event) & 42 & -- & 47 & -- \\
2 ($m{=}1$) & 45 & $-3/{+}6$, .51 & 47 & $-2/{+}2$, 1 \\
6 ($m{=}3$) & 37 & $-6/{+}1$, .13 & 43 & $-6/{+}2$, .29 \\
12 ($m{=}6$) & 24 & $-18/{+}0$, $<10^{-5}$ & 23 & $-24/{+}0$, $<10^{-6}$ \\
\midrule
Pause $2{\times}100$ & 21 & & 14 & \\
\bottomrule
\end{tabular}}
\caption{\textbf{Null-slot dose on \nope{}}, 50 paired episodes. $m{=}6$ against $m{=}1$: $-21/{+}0$ and $-24/{+}0$ (both $p<10^{-5}$).}
\label{tab:supp-nullblock}
\end{wraptable}
\subsection{What these controls do not establish}
\label{supp:noclock:limits}

The read-out is only partly identified: null swap rules out the phase shown in the null frames, duplicate fill rules out sparser real history, and null block shows that the damage grows with the number of null slots, but not which property of a block of identical slots \nope{} reads.
A second training seed of both models repeats the main cells (\cref{supp:seed}), with a less clean press-count contrast; same-seed retrainings of \clk{} differ by up to 9 episodes on hidden cells (\cref{supp:tab:reruns}).

\section{Other memory designs and an independent policy}
\label{supp:designs}
Among RoboMME's released memories, the way a memory selects its contents predicts what a null event does to it.
Two independent policies on other backbones, which both add to memory at every step, fit less neatly; we report them in full.

\subsection{SimpleMemVLA}
\label{supp:smv}

SimpleMemVLA~\cite{yin2026simplememvla} is an independent code base (Qwen3.5-4B with an action head of about 0.9B parameters) that shares only the benchmark with MME-VLA; we evaluate its released RoboMME checkpoint unchanged.
It keeps every frame at 2\,Hz in a fixed window of 120 frames, each preceded by a textual time stamp.
The window is padded at the start with copies of the first frame, so the stamps always read 0.2 to 59.2\,s; elapsed time is visible only as how many slots hold real frames.
On PickXtimes (\cref{tab:supp_smv}), \emph{history resampling}, the analogue of rescaling the time stamps of FrameSamp, spreads the real history over $\mathrm{round}(n s)$ slots (frames repeated for $s{=}1.5$, skipped for $s{=}0.5$; stamps and current frame unchanged).
\emph{Text-stamp rescaling} multiplies the stamps by $s$, which moves the current frame's stamp to values never seen in training; we treat it as exploratory.

\begin{table*}[t]
\centering
\small
\setlength{\tabcolsep}{3pt}
\suppfit{%
\begin{tabular}{llcllr}
\toprule
Split & Condition & Succ. & Failures & vs ref. & $p$ \\
\midrule
Test & No event (ref.) & 50 & -- & -- & -- \\ %
Test & History $\times1.5$ & 44 & 6 early, all $K{=}5$ & $-6/{+0}$ & .031 \\ %
Test & History $\times0.5$ & 45 & 4 extra, 1 timeout; all $K{=}5$ & $-5/{+0}$ & .063 \\ %
Test & Text stamps $\times1.5$ & 49 & 1 (stage 0) & $-1/{+0}$ & 1 \\ %
Test & Text stamps $\times0.5$ & 47 & 3 early, $K{=}5$ & $-3/{+0}$ & .25 \\ %
\midrule
Val & No event (ref.) & 49 & 1 & -- & -- \\ %
Val & History $\times1.5$ & 44 & 6 early, all $K{=}5$ & $-6/{+1}$ & .13 \\ %
Val & History $\times0.5$ & 45 & 4 extra ($K{=}5$), 1 ($K{=}1$) & $-5/{+1}$ & .22 \\ %
\midrule
Test & Pause, hidden (ref.) & 50 & -- & -- & -- \\ %
Test & Pause, visible & 47 & 3 & $-3/{+0}$ & .25 \\ %
Test & Fidget, hidden (ref.) & 50 & -- & -- & -- \\ %
Test & Fidget, visible & 38 & 8 early, 3 button, 1 stage 0 & $-12/{+0}$ & $<10^{-3}$ \\ %
\bottomrule
\end{tabular}}
\caption{\textbf{SimpleMemVLA on PickXtimes} (released checkpoint, successes of 50 paired episodes; exact McNemar against the reference of each block).
Pauses and fidgets are two events of 100 steps (5\,s each).
Early: button pressed before the $K$-th repetition. Extra: failure at the button stage after an extra repetition. Button: failure at the button stage.}
\label{tab:supp_smv}
\end{table*}

\paragraph{Resampling the history shifts the failure; rescaling the stamps does not.}
A history that looks 1.5 times longer causes only early stops, and one that looks half as long mostly extra repetitions, with the same counts on test and on held-out validation episodes.
Almost all of these failures fall among the 6 episodes per split with $K{=}5$ ($\times1.5$: 6 of 6 on both splits; $\times0.5$: 5 of 6 on test, 4 of 6 on validation); no episode with $K\leq4$ fails on test.
The cells are individually weak ($p{=}.031$ and $.063$ on test, $.13$ and $.22$ on validation), and the $\times1.5$ failures occur when the resampled history fills 98 to 119 of the 120 slots, more than any unperturbed PickXtimes episode does (at most about 88), so part of this effect may be out of distribution.
Rescaling the text stamps barely matters (49 and 47); the three failures at $\times0.5$ are early stops, opposite to the clock prediction, at stamps outside the training range, and we do not interpret them.

\paragraph{Null events on PickXtimes: fidgets hurt, pauses barely do.}
Two 5-second pauses add about 20 frames to a window that our episodes never fill and cost 3 episodes against the hidden condition (47 against 50, $p{=}.25$); we had predicted a loss of at least 5, so this prediction failed.
Fidgets of the same duration cost 12 (38 against 50, $p<10^{-3}$), mostly early stops, spread over all $K$ rather than confined to $K{=}5$.
We did not run 15-second pauses: 60 additional frames would push some episodes toward the end of the window and mix truncation into the comparison.

\begin{wraptable}{r}{0.59\linewidth}
\typeout{WRAPLABEL:tab:supp_smv_swing}%
\centering
\small
\setlength{\tabcolsep}{3pt}
\suppfit{%
\begin{tabular}{lcllr}
\toprule
Condition & Succ. & Failures & vs ref. & $p$ \\
\midrule
No event (ref.) & 49 & 1 early & -- & -- \\
History $\times1.5$ & 50 & -- & $-0/{+1}$ & 1 \\
History $\times0.5$ & 42 & 5 extra, 3 early & $-8/{+1}$ & .039 \\
\midrule
Pause, hidden (ref.) & 48 & 2 early & -- & -- \\
Pause, visible & 36 & 11 early, 3 extra & $-13/{+1}$ & .002 \\
\midrule
Fidget, hidden (ref.) & 49 & 1 early & -- & -- \\
Fidget, visible & 33 & 11 early, 6 extra & $-16/{+0}$ & $<10^{-4}$ \\
\bottomrule
\end{tabular}}
\caption{\textbf{SimpleMemVLA on SwingXtimes} (released checkpoint, 50 paired episodes, exact McNemar against the reference of each block; events of 100 steps). Extra: an additional swing, which ends the episode at the put-down stage.}
\label{tab:supp_smv_swing}
\end{wraptable}
\paragraph{SwingXtimes (pre-registered).}
We repeated the null-event and history-resampling cells on SwingXtimes (\cref{tab:supp_smv_swing}), with predictions written before any result: fidgets visible $\leq$ hidden $-6$ with mostly early failures (P1); pauses less harmful than fidgets (P2); history $\times1.5$ mostly early and $\times0.5$ mostly extra failures (P3).
P1 holds: fidgets cost 16 (33 against 49, $p<10^{-4}$), 11 of 17 failures early.
The fidget nudges the held cube by more than our validity tolerance in most episodes (37 visible, 34 hidden), but this physical disturbance is identical when the fidget frames are hidden, which costs a single episode; among the 31 episodes nudged in both conditions, 16 succeed with the frames visible and 30 with them hidden.
P2 holds, narrowly: pauses cost 12 (36 against 48, $p{=}.002$), 11 of 14 failures early, unlike on PickXtimes.
P3 holds only in part: history $\times0.5$ costs 7 (42 against 49, $p{=}.04$), mostly through extra swings, but $\times1.5$ costs nothing (50).
SimpleMemVLA's own spoken sub-goal often runs ahead of the environment's after a fidget (e.g.\ announcing the second swing while the first is unfinished), so its errors are visible in its language output as well.

\subsection{LTM-$\pi$}
\label{supp:ltm}

LTM-$\pi$~\cite{sun2026ltm} is a second independent code base (a modified OpenPI $\pi_{0.5}$ with a frozen latent-action model). For every pair of consecutive frames it appends one memory token, a projection of the latent transition between them, with ordinal positions and no explicit time stamp. It reports 100 of 100 on PickXtimes and SwingXtimes. We evaluate the released checkpoint (trained on ten times the RoboMME demonstrations) unchanged, through its own server, with our injection client.
Deviations from its own evaluator: CPU rendering (as for all our runs), noise seed 7, pauses after policy steps 50 and 160 (it replans every 10 steps), no retry on error.
The client checks that each visible frame adds one token, a visible event 100, and a hidden event none.

\emph{Predictions fixed before the run} (gate: no-event success $\geq45$ on both tasks).
If LTM-$\pi$ reads its memory length, visible pauses should cost at least 6 against hidden ones.
If it counts motion, as its tokens suggest, pauses should cost at most 3 and fidgets at least 6.
We expected the second.

\begin{table*}[t]\centering\small
\suppfit{%
\begin{tabular}{@{}llcllr@{}}\toprule
Task & Condition & Succ. & Failures & vs ref. & $p$ \\\midrule
PickXtimes & No event & 50 & -- & -- & -- \\
 & Pause, hidden & 50 & -- & $-0/{+0}$ (no event) & 1 \\
 & Pause, visible & 7 & 32 early, 11 other & $-43/{+0}$ (hidden) & $<10^{-12}$ \\
 & Fidget, hidden & 50 & -- & $-0/{+0}$ (no event) & 1 \\
 & Fidget, visible & 15 & 31 early, 4 other & $-35/{+0}$ (hidden) & $<10^{-10}$ \\\midrule
SwingXtimes & No event & 50 & -- & -- & -- \\
 & Pause, hidden & 31 & 19 extra & $-19/{+0}$ (no event) & $<10^{-5}$ \\
 & Pause, visible & 9 & 35 early, 6 other & $-41/{+0}$ (no event) & $<10^{-12}$ \\
 & Fidget, hidden & 32 & 18 extra & $-18/{+0}$ (no event) & $<10^{-5}$ \\
 & Fidget, visible & 10 & 36 early, 4 other & $-40/{+0}$ (no event) & $<10^{-11}$ \\\bottomrule
\end{tabular}}
\caption{\textbf{LTM-$\pi$} (released checkpoint, 50 paired episodes, exact McNemar; events of 100 steps). Early: button pressed before the $K$-th repetition. Extra: an additional swing, which ends the episode at the put-down stage.}
\label{tab:supp_ltm}
\end{table*}

The gate passes (50 and 50). The motion account fails: visible pauses cost LTM-$\pi$ as much as fidgets do, almost always through early presses, and on PickXtimes hiding either event leaves success at 50.
A still interval is not empty for this memory: it adds one token per step, each encoding a near-zero transition, and the policy reads the extra tokens as progress.
On SwingXtimes, hiding the events does not restore success fully (31 and 32): the hidden failures are all extra swings, on nearly the same episodes for pause and fidget, the opposite error to the visible one.
Keeping the hold frames until the arm is still does not close this gap (34; \cref{supp:hidstatic}), so the SwingXtimes comparisons above are against no event; visible remains below hidden ($p<10^{-4}$ for both events).

\subsection{Full numbers for the comparison of memory designs}
\label{supp:designs_full}

\Cref{tab:supp_designs} gives the cells behind \cref{tab:designs}, with hidden references where they exist, paired $p$ values, and the two released TokenDrop variants not shown in the main text.
$K{\geq}2$ counts successes without an event among episodes with at least two repetitions, the only ones in which counting requires memory.

\begin{table*}[t]
\centering
\small
\setlength{\tabcolsep}{2.5pt}
\suppfit{%
\begin{tabular}{llcccrcr}
\toprule
Memory & Task & No event & $K{\geq}2$ & Pause & $p$ & Fidget & $p$ \\
 & & & & vis.\,/\,hid. & & vis.\,/\,hid. & \\
\midrule
FrameSamp-Modul & P & 47 & 31/34 & 16\,/\,44 & $<10^{-9}$ & 16\,/\,46 & $<10^{-9}$ \\ %
 & S & 44 & 30/35 & 12\,/\,46 & $<10^{-9}$ & 10\,/\,45 & $<10^{-9}$ \\ %
FrameSamp-Expert & P & 44 & 28/34 & 14\,/\,45 & $<10^{-8}$ & 17\,/\,46 & $<10^{-7}$ \\ %
 & S & 49 & 34/35 & 11\,/\,-- & $<10^{-11}$ & -- & \\ %
FrameSamp-Context & P & 33 & 17/34 & 11\,/\,-- & $<10^{-6}$ & 20\,/\,-- & .004 \\ %
 & S & 43 & 29/35 & 10\,/\,-- & $<10^{-9}$ & -- & \\ %
\midrule
TokenDrop-Modul & P & 39 & 23/34 & 30\,/\,40 & .035 & 21\,/\,42 & $<10^{-5}$ \\ %
 & S & 45 & 30/35 & 27\,/\,-- & $<10^{-4}$ & 11\,/\,36 & $<10^{-9}$ \\ %
TokenDrop-Context & P & 37 & 21/34 & 32\,/\,-- & .18 & -- & \\ %
 & S & 47 & 32/35 & 26\,/\,-- & $<10^{-5}$ & -- & \\ %
TokenDrop-Expert & P & 46 & 30/34 & 32\,/\,-- & $<10^{-3}$ & -- & \\ %
 & S & 47 & 32/35 & 33\,/\,-- & $<10^{-3}$ & -- & \\ %
\midrule
MemER-style keyframes & P & 38 & 22/34 & 39\,/\,-- & 1 & 38\,/\,-- & 1 \\ %
 & S & 24 & 19/35 & 23\,/\,-- & 1 & -- & \\ %
Text sub-goals & P & 46 & 30/34 & 48\,/\,-- & .50 & 47\,/\,-- & 1 \\ %
 & S & 1 & 1/35 & 2\,/\,-- & 1 & -- & \\ %
\midrule
TTT (recency window) & P & 23 & 7/34 & 22\,/\,-- & 1 & 25\,/\,-- & .77 \\ %
 & S & 17 & 2/35 & 16\,/\,-- & 1 & -- & \\ %
$\pi_{0.5}$ (no memory) & P & 18 & 2/34 & 19\,/\,-- & 1 & 20\,/\,-- & .69 \\ %
 & S & 15 & 1/35 & 13\,/\,-- & .50 & -- & \\ %
\bottomrule
\end{tabular}}
\caption{\textbf{Released RoboMME memory policies under two null events} (P: PickXtimes, S: SwingXtimes; successes of 50 paired episodes).
Pauses: 300 steps; fidgets: 100 steps; vis./hid.: visible / hidden.
$p$: exact McNemar, visible against no event.
$K{\geq}2$: successes without an event among episodes with $K\geq2$ (34 on PickXtimes, 35 on SwingXtimes).
--: not run.}
\label{tab:supp_designs}
\end{table*}

\paragraph{Which ``unaffected'' rows are informative.}
TTT and $\pi_{0.5}$ succeed in nearly every episode with $K{=}1$ (31 of 31 and 30 of 31) but in almost none with $K\geq2$ (9 of 69 and 3 of 69), so their indifference to null events says nothing about their memory; the same holds for text sub-goals on SwingXtimes (1 of 50).
The text sub-goals are predicted online by RoboMME's released Qwen-VL sub-goal model from the current frame, the demonstration video and the sub-goals logged so far (no oracle sub-goals). The informative rows are MemER-style keyframes and text sub-goals on PickXtimes, which solve most episodes with $K\geq2$ (22 and 30 of 34, against 31 for FrameSamp-Modul) and are unchanged by pauses (39 and 48) and by fidgets (38 and 47).
Under one-way air-grasps MemER drops from 38 to 27, against 39 when the air-grasps are hidden ($-16/{+4}$, $p{=}.012$); text sub-goals are unaffected (44 against 45 hidden).

\paragraph{TokenDrop.}
Pauses cost TokenDrop less than FrameSamp but still significantly on PickXtimes (30 against 40 hidden, $-12/{+2}$, $p{=}.013$).
Fidgets pass the pixel filter and cost far more than pauses: 21 against 42 hidden on PickXtimes ($p<10^{-6}$) and 11 against 36 on SwingXtimes ($p<10^{-7}$).
On SwingXtimes the hidden fidget itself costs 9 episodes (36 against 45, $p{=}.022$); comparing against the hidden condition removes this physical part.
Rescaling TokenDrop's time stamps (\cref{tab:supp_warp}) shifts the failures in the same directions as for FrameSamp, but the drop in success is not significant, and two no-event runs on validation episodes differ by 5 successes, so we treat its clock evidence as exploratory.

\section{Data-level fixes}
\label{supp:data}
None of the augmentations we tested fixes the failure.
Pause augmentation removes the pause damage on PickXtimes only by lowering success without events, a weaker dose keeps more counting but helps no more under pauses, and the earlier augmentations become indifferent to null events because they no longer count.

All models in this section are trained from the $\pi_{0.5}$ base checkpoint with the recipe of \clk{} (same data, 30k steps, batch 64, clock on), changing only what the memory sees during training, and are evaluated on the same 50 paired episodes as \clk.
Earlier training arms are not reported: an implementation error left their data modifications inactive (applied in the main process but not in the data-loader workers). %
After the fix, the modified loader reproduces the original samples with all modifications off (60 of 60) and changes the memory inputs with them on (16 of 16). %

\subsection{Pause-only augmentation}
\label{supp:pauseonly}

\paragraph{Design.}
On 80\% of training samples we insert 1--3 synthetic pauses of 20--600 steps into the memory timeline by repeating a frame, with the time stamps advancing as in a real pause; nothing else changes (no replayed segments, no time warp). %
A held frame is a faithful pause, so if the damage were merely an unseen input, this model should become indifferent to pauses.

\paragraph{Pre-registered decision rules} (written before any result). %
\begin{itemize}
\item \emph{Pauses fixed}: pause 100 $\ge$ own hidden $-\,6$ \emph{and} no event $\ge$ \clk{} $-\,6$ (PickXtimes $\ge35$, SwingXtimes $\ge42$). \emph{Not fixed}: pause 100 within $\pm8$ of \clk{} (17 / 10). Otherwise partial.
\item \emph{Moving nulls} (predicted not fixed): PickXtimes fidget $\le24$ not fixed, $\ge$ own hidden $-\,6$ fixed; SwingXtimes fidget $\le18$ not fixed, same upper rule; SwingXtimes air-grasp-ret $\le21$ not fixed, $\ge35$ fixed.
\item \emph{Clock}: $\times1.5$ $\ge$ own no event $-\,6$: clock dependence removed; $\le27$: still reads the clock. $\times0.5$ $\le$ own no event $-\,10$ with mostly extra repetitions: still reads the clock.
\item \emph{Route}: time only $\le$ own no event $-\,10$: the stamp route is still used; $\ge$ own no event $-\,6$: removed. Content only: no rule, reported descriptively.
\end{itemize}

\begin{table*}[t]
\centering
\small
\setlength{\tabcolsep}{4pt}
\suppfit{\begin{tabular}{llcccc}
\toprule
Task & Condition & \clk{} & Retrain & + pause aug. & + weaker \\
\midrule
\multirow{10}{*}{PickXtimes}
 & No event & 41 & 32 & 24 & 39 \\ %
 & No event, held-out & 38 & 45 & 17 & 31 \\ %
 & Time stamps $\times1.5$ & 19 & -- & 24 & 30 \\
 & Time stamps $\times0.5$ & -- & -- & 21 & -- \\
 & Pause 100, visible & 17 & 16 & 25 & 19 \\
 & Pause 100, hidden & 38 & 35 & 22 & 34 \\
 & Pause 100, time only & -- & -- & 24 & -- \\
 & Pause 100, content only & -- & -- & 21 & -- \\
 & Pause 300, visible & 15 & -- & 21 & -- \\
 & Fidget 100 & 16 & -- & 21 & 18 \\
\midrule
\multirow{6}{*}{SwingXtimes}
 & No event & 48 & 46 & 19 & 48 \\
 & No event, held-out & 44 & 44 & 20 & 45 \\
 & Pause 100, visible & 10 & 11 & 10 & 12 \\
 & Pause 100, hidden & 38 & 40 & 16 & 43 \\
 & Fidget 100 & -- & -- & 14 & -- \\
 & Air-grasp-ret & 13 & -- & 17 & 24 \\
\bottomrule
\end{tabular}}
\caption{\textbf{Pause-only augmentation}, successes of 50 paired episodes. \clk: cluster. Retrain and both pause augmentations: same recipe, second machine. ``--'': not run.}
\label{tab:supp-pauseonly}
\end{table*}

\paragraph{Second machine and its control.}
The pause-only model was trained and evaluated on a second machine, with the recipe of \clk{} except FSDP on 2 instead of 4 GPUs; evaluation parity with the cluster is in \cref{supp:stats}. %
To separate the machine from the augmentation we retrained \clk{} there without augmentation (rule written before that training: no-event success at least 35 and 42 excludes the machine, within 6 of the pause-only model implicates it).
SwingXtimes passes (46).
PickXtimes on the test split falls between the two (32), so we ran held-out episodes on both models, again with the rules written first (control at least 32; pause-only at least 6 below it with $p<.05$).
Both were met: 45 against 17 ($-29/{+}1$, $p<10^{-6}$), and on SwingXtimes 44 against 20 ($-24/{+}0$, $p<10^{-6}$).
All comparisons below are therefore with the retrain on the same machine.
The retrain also reproduces the pause damage in a new training on a new machine: 32 to 16 and 46 to 11 under visible pauses, with hidden pauses at 35 and 40.
Its difference from \clk{} on PickXtimes without events goes both ways ($-9$ on test, $+7$ on held-out episodes; \cref{tab:supp-pauseonly}), so comparisons between trainings on this cell carry that uncertainty.

\paragraph{Result.}
By the rules the pauses are not fixed.
The no-event clause fails on both tasks: 24 and 19, against 35 and 42, and also against the retrain.
Pause 100 is not fixed on SwingXtimes (10; control 11).
On PickXtimes it is at the upper edge of the rule against \clk{} (25 against 17 $\pm$ 8) and just outside it against the retrain (25 against 16, $-0/{+}9$, $p{=}.004$), which we count as partial: the pause no longer costs anything (25 against its own 24), at a lower level.
On SwingXtimes visible pauses still cost it episodes (19 to 10, $-10/{+}1$, $p{=}.012$), including at $K{=}1$ (9 of 15).
The air-grasp is not fixed (17).
The stamp route is removed (time only 24 against its own 24, $-3/{+}3$); the $\times0.5$ rule for still reading the clock is not triggered (21).
Three cells meet both conditions of their rule, because the rules assumed that no-event success would hold: $\times1.5$ (24), PickXtimes fidget (21) and SwingXtimes fidget (14).
We give no verdict on them; against the model's own no-event cell they are $-6/{+}6$, $-6/{+}3$ and $-5/{+}0$, none significant.
The loss is in counting: $K{=}1$ is untouched (\cref{tab:supp-aug-k}, test split), and on held-out episodes $K{\ge}2$ falls from 35 to 7 of 40 and from 25 to 1 of 31.
On the test split, the same fall is 31 to 4 of 35 on SwingXtimes and 16 to 8 of 34 on PickXtimes ($-12/{+}4$, $p{=}.077$, not significant).

\paragraph{A weaker dose.}
This model adds about 270 steps of pause per sample on average. To test whether the strength alone causes the loss, we trained a weaker dose on the same machine with the same seed: pauses on 30\% of samples, at most 2, of 20--100 steps (about 22 null steps per sample), no noise.
Rules, written before training, against the retrain: \emph{counting kept} if no-event success is at least 26 and 40 on test and at least 39 and 38 on held-out episodes; \emph{pauses improved} if pause 100 is at least 25 and 20 and within 6 of its own hidden cell.
Both kept: a weaker dose works for pauses. Counting kept only: it keeps counting but does not help under pauses. Counting not kept: the cost is not only a matter of strength. %
SwingXtimes keeps counting (48 on test, 45 held-out; $K{\ge}2$ held-out 27 of 31, retrain 25) and pauses do not improve (12 against 20; its own hidden 43): it keeps counting but does not help under pauses.
PickXtimes does not keep counting on held-out episodes (31 against 39; retrain 45, $-15/{+}1$, $p<10^{-3}$; $K{\ge}2$ 21 of 40, retrain 35, strong dose 7), although on test it is above the retrain (39 against 32), and pauses do not improve (19 against 25; its own hidden 34).
By the rule, on PickXtimes the counting cost is not only a matter of strength; it is smaller at the weaker dose, and the gap to \clk{} on the cluster (38 held-out, $-10/{+}3$, $p{=}.09$) is within the spread between trainings on this cell.
The weaker dose also leaves the clock read: time stamps $\times1.5$ lower success from 39 to 30 ($-12/{+}3$, $p{=}.035$; 19 of 20 failures early).
At neither strength does pause augmentation move the pause damage.

\subsection{Augmentation with replayed segments, and training on event-indexed timelines}
\label{supp:paug}

Three earlier training interventions were run after the fix: \textbf{paug} adds to the pauses, on 50\% of samples, a \emph{replayed segment} (the most recent 15--60 frames inserted reversed and then forward, so the robot appears to go out and come back); %
\textbf{saug2} adds to paug a time warp of the history (0.5--2$\times$, log-uniform) on 80\% of samples; %
\textbf{eim2} is trained on memories passed through the simulator-contact event index of \cref{supp:eventindex}. %

\paragraph{They are indifferent to null events because they no longer count.}
saug2 and paug are flat across all their cells (PickXtimes 20--27, SwingXtimes 11--22): pauses, fidgets, air-grasps and even $\times1.5$ (24 and 23) no longer hurt, and within each model no null-event cell differs significantly from its own no-event cell. %
But their no-event success falls far below \clk{}: PickXtimes 25 and 22 against 41 ($-16/{+}0$, $-19/{+}0$), SwingXtimes 17 and 16 against 48 ($-31/{+}0$, $-32/{+}0$), all $p<10^{-4}$. %
\Cref{tab:supp-aug-k} shows why: $K{=}1$ is intact, $K\ge2$ almost always fails (on SwingXtimes, 23 of 33 saug2 failures occur during the swing itself, which we attribute to the time warp). %
eim2 counts better; its 9-episode gap on PickXtimes without any null event (32 against 41, $-11/{+}2$, $p{=}.022$, all at $K\ge3$) is within what an unaugmented retrain on a second machine also shows on this cell (\cref{supp:pauseonly}), so we do not read it as a cost. %
paug and saug2 therefore cannot serve as a fix, and eim2 shows no gain under null events.
We first attributed the loss in paug and saug2 to the replayed segment, which copies recorded frames, objects included, so that in memory it looks like a real repetition that the label says does not count. %
The pause-only model has no replayed segments and loses counting too (\cref{supp:pauseonly}), so the segment is not necessary for this loss, though it remains unusable as a moving null event for the same reason.
This holds for any moving null synthesised from recorded demonstrations, the only option on real-robot data (a simulator could render true moving nulls, which we did not do).
Applied to demonstrations, the event index in turn collapses real repetitions (116 of 191 PickXtimes loops cross a sub-goal change, \cref{tab:supp-le-audit}), so eim2 learns to count from memories with repetitions missing. %
Because both augmentations contain replayed segments, they do not test whether faithful pause augmentation alone removes the pause damage; the pause-only model does.

\begin{table*}[t]
\centering
\small
\setlength{\tabcolsep}{4pt}
\suppfit{\begin{tabular}{lccccc@{\hspace{10pt}}ccc}
\toprule
 & \multicolumn{5}{c}{PickXtimes, $K=$} & \multicolumn{3}{c}{SwingXtimes, $K=$} \\
\cmidrule(lr){2-6}\cmidrule(lr){7-9}
Model & 1 & 2 & 3 & 4 & 5 & 1 & 2 & 3 \\
\midrule
\clk{} & 16/16 & 9/10 & 12/12 & 2/6 & 2/6 & 15/15 & 12/12 & 21/23 \\
saug2 & 15/16 & 3/10 & 6/12 & 0/6 & 1/6 & 15/15 & 0/12 & 2/23 \\
paug & 16/16 & 1/10 & 4/12 & 1/6 & 0/6 & 15/15 & 0/12 & 1/23 \\
eim2 & 16/16 & 9/10 & 6/12 & 1/6 & 0/6 & -- & -- & -- \\
\midrule
Retrain (second machine) & 16/16 & 7/10 & 7/12 & 1/6 & 1/6 & 15/15 & 12/12 & 19/23 \\
Pause only (second machine) & 16/16 & 1/10 & 6/12 & 1/6 & 0/6 & 15/15 & 1/12 & 3/23 \\
Weaker pause only (second machine) & 16/16 & 7/10 & 12/12 & 2/6 & 2/6 & 15/15 & 12/12 & 21/23 \\
\bottomrule
\end{tabular}}
\caption{\textbf{No-event success by $K$} (successes / episodes, no event index at test). The augmented models keep $K{=}1$ and lose most $K\ge2$ episodes, except the weaker pause dose on this split.}
\label{tab:supp-aug-k}
\end{table*}

\section{The event index}
\label{supp:eventindex}

The event index of \cref{sec:fixes} is a test-time intervention that removes null frames from a time-indexed memory, to test the diagnosis rather than to propose a general null-event detector.
\Cref{tab:fixes} gives the full numbers behind \cref{fig:fixes}.

\begin{table}[t]
\centering
\small
\setlength{\tabcolsep}{3pt}
\suppfit{\begin{tabular}{lcccc}
\toprule
 & No event & Pause & Fidget & Air-grasp \\
 & P / S & P / S & P & S \\
\midrule
\clk{} (no fix) & 41 / 48 & 17 / 10 & 16 & 13 \\
+ event index (representation) & 36 / 48 & 34 / 30 & 33 & 43 \\
\midrule
\clk{} retrained, second machine & 32 / 46 & 16 / 11 & -- & -- \\
+ pause augmentation (data) & 24 / 19 & 25 / 10 & 21 & 17 \\
+ weaker pause augmentation & 39 / 48 & 19 / 12 & 18 & 24 \\
\midrule
Hidden (reference) & -- & 38 / 38 & -- & -- \\
\bottomrule
\end{tabular}}
\caption{\textbf{The two fixes the diagnosis predicts}, on from-scratch models of the same recipe, test split (on which the detector's tolerances were set; held-out replication in the text). Pause augmentation, at two strengths, was trained on a second machine and is compared with the retrain above it.
P / S: PickXtimes / SwingXtimes; air-grasps return to the starting pose. Event index: test-time detector on simulated proprioceptive, silhouette and contact signals (no object states).
Successes of 50 paired episodes.}
\label{tab:fixes}
\end{table}

\subsection{Definition}
\label{supp:le-def}

The detector processes the policy's execution frames online and maintains a \emph{kept} sequence; only kept frames enter memory, and their time positions are their indices in the kept sequence, so the memory clock advances only with kept frames.
A frame is a \emph{barrier} if the scene outside the robot changed with respect to the last kept frame, or if a grip or load switch (\cref{supp:le-signals}) occurred; a barrier is always kept and no stretch is collapsed across it.
Between barriers, two rules apply:
\begin{itemize}
\item \textbf{Loop closure.} If the robot configuration returns to within tolerance of a kept frame $p$ after having left it by more than the exit distance, and the scene outside the robot equals that at $p$, the kept frames after $p$ are collapsed to one frame (the kept frame farthest from $p$) and one clock tick.
\item \textbf{Stillness.} If the configuration stays within the stillness tolerance of the frame at which a stop began and the scene is unchanged, the first 5 frames of the stop are kept (grace) and later ones are dropped.
\end{itemize}
All tolerances (\cref{tab:supp-le-params}) were set during development on test-split episodes and then frozen, identical for all 16 RoboMME tasks and all models; the held-out replication (\cref{supp:le-released}) used environment seeds never used in development.

\begin{table*}[t]
\centering
\small
\setlength{\tabcolsep}{3pt}
\suppfit{\begin{tabular}{lll}
\toprule
Quantity & Value & Code name \\
\midrule
Joint return tolerance (arm) & 0.02\,rad (1 unit) & \texttt{LOOP\_TOL} \\ %
Finger return tolerance & 0.004\,m (1 unit) & \texttt{LOOP\_TOL\_GRIP} \\ %
Exit distance before a return counts & 2 units & \texttt{LOOP\_EXIT} \\ %
Stillness tolerance & 0.05 units & \texttt{LOOP\_STILL} \\ %
Scene unchanged at closure & $<5$ px differ by $>16/255$ & \texttt{LOOP\_SCENE\_PIX}, \texttt{DETECT\_DPIX} \\ %
Robot silhouette dilation & 3 px & \texttt{DETECT\_DIL} \\ %
Stillness grace & 5 frames & \texttt{LOOP\_GRACE} \\ %
Grip / load debounce & 3 frames & \texttt{CONTACT\_N}, \texttt{NP\_S} \\ %
Grip states: empty / open & $<0.004$\,m / $>w_{\max}-0.004$\,m & \texttt{NP\_W\_EMPTY}, \texttt{NP\_W\_OPEN} \\ %
Load on / off (net force) & $>0.02$ / $<0.01$ (sim.\ units) & \texttt{NP\_F\_ON}, \texttt{NP\_F\_OFF} \\ %
\bottomrule
\end{tabular}}
\caption{\textbf{Event-index parameters}, fixed during development and used unchanged for all tasks and models.}
\label{tab:supp-le-params}
\end{table*}

\subsection{Signals and how they are obtained}
\label{supp:le-signals}

The decision uses four signals, none of which is the simulator's object state.
\begin{itemize}
\item \textbf{Robot joint configuration} (7 arm joints, 2 finger joints): proprioception, available on any arm.
\item \textbf{Robot silhouette} in the front camera, used only to ignore the robot's own pixels when testing whether the scene changed. \emph{In all experiments it is the simulator's segmentation of the robot links.} On a robot it can be rendered from the joint readings and the camera calibration; we did not implement this rendering.
\item \textbf{Gripper width.} The gripper is \emph{holding} while the width has settled strictly between closed-on-nothing and open; a switch of this state (a grasp or a release) is an event. An air-grasp closes to empty and never holds.
\item \textbf{Net external force on the arm}: \emph{in all experiments the sum of the simulator's contact impulses between the robot and anything else}, with the robot links merged so that grip forces cancel. This is the quantity a wrist force/torque sensor measures; we did not model sensor noise. A switch between loaded and unloaded (lift-off, set-down, press, push) is an event.
\end{itemize}
The \emph{simulator-contact variant} replaces the last two signals by the simulator's list of contact pairs. Both variants use the simulator segmentation and noise-free renders; the 5-pixel scene tolerance would have to be revisited for a real camera.

\subsection{What the detector collapses}
\label{supp:le-collapse}

\paragraph{Recall on injected null intervals.}
\Cref{tab:supp-le-recall} gives the fraction of injected null frames that the detector removes.
Pauses are removed almost entirely; the missed frames are the deceleration at the start of each pause and the 5 grace frames.
A one-way air-grasp ends elsewhere, so its descent and approach never close a loop, and only about half of it is removed.
For returning events the closure is all-or-nothing: in 4 of the episodes lost on SwingXtimes air-grasp-ret, the return stopped 1.3--3 units short of the starting configuration and the whole excursion ($\approx$110 frames) was kept. %
On the \clk{} model, each 100-step pause on SwingXtimes leaves about 11 frames (recall $\approx$0.87). %

\begin{table*}[t]
\centering
\small
\setlength{\tabcolsep}{3pt}
\suppfit{\begin{tabular}{lccc}
\toprule
Cell (released FS-Modul) & Recall & \begin{tabular}[b]{@{}c@{}}Collapsed outside\\injected windows\end{tabular} & \begin{tabular}[b]{@{}c@{}}Collapsed frames with\\a moving object\end{tabular} \\
\midrule
PickXtimes, pause 300 $\times2$ & 0.96 & 22.7\% & all while held \\ %
RouteStick, pause 300 $\times2$ & 0.95 & 0.0\% & none \\ %
SwingXtimes, one-way air-grasp & 0.62 & 4.7\% & none \\ %
PickXtimes, no event & -- & 23.9\% & all while held \\ %
Swing / StopCube / VPB / RouteStick, no event & -- & 0--1.3\% & none (12 held frames, Swing) \\ %
\bottomrule
\end{tabular}}
\caption{\textbf{Detector recall and collapse outside the injected windows}, simulator-contact variant on the released policy, 50 episodes per cell (version with loop contraction; the version without contraction gives 0.96, 0.95 and 0.52 recall).
``All while held'': every collapsed frame in which the cube moved shows the closed gripper carrying it around a closed path; no placement or push was collapsed, and resting objects in collapsed frames moved $<1$\,mm.}
\label{tab:supp-le-recall}
\end{table*}

\paragraph{The policy's own loops.}
PickXtimes is the only task in which the policy makes many loops of its own: after releasing a cube it lifts and comes back down to it (26.6 loop closures per episode without null events; 23.9\% of frames collapsed, against 0.6\% on BinFill). %

\paragraph{Demonstration-replay audit.}
Replayed on the RoboMME training demonstrations (\cref{tab:supp-le-audit}), the detector collapses 19.0\% of PickXtimes frames, and 116 of 191 closed loops have ends carrying different recorded sub-goals, almost all ``place the cube onto the target'' $\to$ ``pick up the cube for the second time'': a real repetition returns the robot and the scene to a state they were in before. VideoRepick shows the same pattern (25.6\%, 144 of 205). %
These two tasks are where an event index can fail by construction (\cref{supp:le-cost}).

\begin{table}[t]
\centering
\small
\setlength{\tabcolsep}{4pt}
\suppfit{\begin{tabular}{lcccc}
\toprule
Task & Demos & Collapsed & Loops & \begin{tabular}[b]{@{}c@{}}Loops crossing\\a sub-goal\end{tabular} \\
\midrule
PickXtimes & 98 & 19.0\% & 191 & 116 \\
VideoRepick & 93 & 25.6\% & 205 & 144 \\
BinFill & 95 & 5.9\% & 75 & 10 \\
SwingXtimes & 96 & 5.5\% & 65 & 4 \\
PickHighlight & 97 & 4.8\% & 25 & 20 \\
PatternLock & 100 & 1.6\% & 15 & 12 \\
StopCube & 100 & 0.4\% & 11 & 0 \\
Other 9 tasks & 93--100 & 0.0--5.4\% & 0--7 & 0 \\
\bottomrule
\end{tabular}}
\caption{\textbf{Demonstration-replay audit} of the simulator-contact detector on the RoboMME training demonstrations. A loop crossing a sub-goal change is a real step of the task that the detector would collapse.}
\label{tab:supp-le-audit}
\end{table}

\subsection{Effect on the released policies}
\label{supp:le-released}

\Cref{tab:supp-le-grid} gives the full probe grid on the released FrameSamp-Modul policy (simulator-contact variant) with the hidden reference and two appearance-based baselines: \emph{g0}, which drops every frame in which neither the robot state nor the image changes (no grace) without advancing the clock over them, and TokenDrop, the released memory that selects tokens by pixel change.

\begin{table*}[t]
\centering
\small
\setlength{\tabcolsep}{4pt}
\suppfit{\begin{tabular}{lcccccccc}
\toprule
 & \multicolumn{5}{c}{Test split} & & \multicolumn{2}{c}{Held-out val split} \\
\cmidrule(lr){2-6}\cmidrule(lr){8-9}
Probe & Stock & + Event index & Hidden & g0 & TokenDrop & & Stock & + Event index \\
\midrule
PickXtimes pause 300 $\times2$ & 16 & 34 & 44 & 42$^{*}$ & 30 & & 12 & 31 \\ %
SwingXtimes one-way air-grasp & 28 & 40 & 48 & 33 & 34 & & 32 & 38$^{\dagger}$ \\ %
SwingXtimes air-grasp-ret & 23 & 42 & 47 & 25 & 27 & & 28 & 39 \\ %
PickXtimes fidget 100 $\times2$ & 16 & 36 & 46 & 16 & 21 & & 10 & 39 \\ %
SwingXtimes fidget 100 $\times2$ & 10 & 38 & 45 & 9 & 11 & & 18 & 33 \\ %
RouteStick pause 300 $\times2$ & 7 & 31 & 36 & 33 & 23 & & 7 & 30 \\ %
VideoPlaceButton pause 300 $\times2$ & 11 & 27 & 29 & 27 & 24 & & 11 & 33 \\ %
\midrule
Pooled (of 350) & 111 & 248 & 295 & -- & -- & & 118 & 243 \\ %
\midrule
No event, PickXtimes / SwingXtimes & 47 / 44 & 41 / 48 & -- & 42 / 45 & 39 / 45 & & \multicolumn{2}{c}{see \cref{tab:supp-le-cost}} \\ %
\bottomrule
\end{tabular}}
\caption{\textbf{Event index on the released FrameSamp-Modul policy} (simulator-contact variant), successes of 50 paired episodes.
Paired counts are episodes lost / gained. Test: every event-index cell differs from stock at $p\le.004$.
Held-out val (pre-registered, new environment seeds, tolerances unchanged): pooled $-14/{+}139$ over 350 pairs ($p<10^{-4}$, $+35.7$ points), 6 of 7 probes at $p<.05$; $^{\dagger}$the one-way air-grasp is not significant ($p{=}.11$).
g0 and TokenDrop are not significantly different from the event index on pauses ($^{*}$g0 is better on the PickXtimes pause, $p{=}.04$) and on the one-way air-grasp, and significantly worse on every moving null that returns ($p\le.003$).
The event index recovers 137 of the 184 episodes between stock and hidden on test (74\%); the remaining 47 are attributed in \cref{tab:supp-le-gap}.}
\label{tab:supp-le-grid}
\end{table*}

Dropping still frames (g0) recovers pauses as well as the event index does, or better (PickXtimes pause: 42 against 34), but recovers nothing under fidgets or a returning air-grasp, which change pixels without changing the world: the same dissociation as TokenDrop in \cref{tab:designs}.

\paragraph{Other released variants.}
The same detector, unchanged, raises FrameSamp-Context under pause 300 from 11 to 27 (PickXtimes) and from 10 to 28 (SwingXtimes), and FrameSamp-Expert from 14 to 35 and from 11 to 34 (Expert hidden, PickXtimes: 45); its cost without null events on both is in \cref{tab:supp-le-cost}. %

\subsection{Non-privileged versus simulator-contact signals}
\label{supp:le-np}

Replayed on 12 demonstrations of each of the 16 tasks, the non-privileged variant misses at most 2 frames per task that the simulator-contact variant collapses. %
It collapses \emph{more} in six tasks (BinFill +102 frames, +20\%; VideoRepick +71; SwingXtimes +45; ButtonUnmaskSwap +17; PickHighlight +14; ButtonUnmask +12; the other 10 tasks 0), because gripper width and net force miss some contact changes (event recall BinFill 0.80, ButtonUnmask 0.51, VideoUnmask 0.29, precision $\approx$1). %
\Cref{tab:supp-le-np} tests whether this over-collapse costs success: none of the 12 compared cells differs significantly.
On \clk{} the two variants also agree: non-privileged / simulator-contact PickXtimes no event 36 / 36, pause 34 / 33, fidget 33 / 31; SwingXtimes no event 48 / 46, pause 30 / 26, air-grasp-ret 43 / 42 (not tested pairwise). %
On held-out episodes (pre-registered: a gain of at least 8 with McNemar $p<.05$ on each task, tolerances unchanged), the non-privileged variant on \clk{} raises PickXtimes under pauses from 10 to 36 ($-0/{+}26$, $p<10^{-7}$; no event 38) and SwingXtimes from 14 to 32 ($-1/{+}19$, $p<10^{-4}$; no event 44); early presses fall from 40 to 9 and from 30 to 15 of the failures. %
These comparisons establish that the decision does not depend on the simulator's contact list, not that the detector is deployable: the silhouette and the force are simulator stand-ins (\cref{supp:le-signals}).

\begin{table}[t]
\centering
\small
\setlength{\tabcolsep}{4pt}
\suppfit{\begin{tabular}{lccc}
\toprule
Cell (released FS-Modul) & Sim.\ contact & Non-priv. & Paired \\
\midrule
PickXtimes, no event & 41 & 40 & $-6/{+}5$ \\ %
SwingXtimes, no event & 48 & 47 & $-1/{+}0$ \\
PickXtimes pause 300 & 34 & 36 & $-2/{+}4$ \\
SwingXtimes one-way air-grasp & 40 & 39 & $-5/{+}4$ \\
SwingXtimes air-grasp-ret & 42 & 43 & $-0/{+}1$ \\
PickXtimes fidget 100 & 36 & 37 & $-5/{+}6$ \\
SwingXtimes fidget 100 & 38 & 38 & $-0/{+}0$ \\
\midrule
BinFill, no event & 22 & 22 & $-3/{+}3$ \\ %
VideoRepick, no event & 13 & 11 & $-3/{+}1$ \\
ButtonUnmaskSwap, no event & 9 & 6 & $-3/{+}0$ \\
ButtonUnmask, no event & 10 & 11 & $-0/{+}1$ \\
PickHighlight, no event & 12 & 11 & $-1/{+}0$ \\
\bottomrule
\end{tabular}}
\caption{\textbf{Non-privileged (gripper width + net force) versus simulator contact pairs}, 50 paired episodes; paired = episodes lost / gained by the non-privileged variant. No difference is significant (smallest $p{=}.25$, ButtonUnmaskSwap). The five no-event tasks are those in which the replay showed extra collapse; the other nine tasks showed none and were not evaluated.}
\label{tab:supp-le-np}
\end{table}

\subsection{Costs, residual gap and boundaries}
\label{supp:le-cost}

\paragraph{Cost without null events.}
On PickXtimes (\cref{tab:supp-le-cost}) the cost was first observed on the test split; a held-out replication was then pre-registered (a net loss with episode-aggregated $p<.05$, three policy seeds) and confirmed: 139 to 120 of 150 (17 episodes worse, 4 better; sign-flip permutation $p{=}.005$). %
Under a Holm correction over our 59-test family neither split survives alone ($p_\text{Holm}=.10$ test, $.15$ val); the pooled test+val comparison does (31 worse, 6 better, $p_\text{Holm}=.001$). %
With the event index the policy performs an extra repetition, which the stock policy never does (stock 0 of 250 episodes, event index 32 of 300 over test and val; val alone 0 against 18 of 150): collapsing the loop back to a released cube gives the policy back the memory that made it leave, so it repeats the step. %
Across the other 15 tasks (SwingXtimes included; without it 272 against 269 of 700) the event index costs nothing overall (316 against 317 of 750, $-22/{+}23$, $p{=}1.0$; a 95\% interval of about $\pm1.8$ points on the net). %

\begin{table*}[t]
\centering
\small
\setlength{\tabcolsep}{3pt}
\suppfit{\begin{tabular}{llccc}
\toprule
Model & Cells & Stock & + Event index & Paired \\
\midrule
FS-Modul & PickXtimes test, 3 seeds & -- & net $-16$ & 14 worse / 2 better \\ %
FS-Modul & PickXtimes val, 3 seeds (pre-reg.) & 139/150 & 120/150 & $p{=}.005$ (perm.) \\ %
FS-Modul & SwingXtimes test & 44 & 48 & $-0/{+}4$ \\ %
FS-Modul & 15 tasks other than PickXtimes & 316/750 & 317/750 & $-22/{+}23$ \\ %
FS-Context & PickXtimes / SwingXtimes & 33 / 43 & 34 / 43 & n.s. \\ %
FS-Expert & PickXtimes / SwingXtimes & 44 / 49 & 39 / 48 & $-6/{+}1$, $p{=}.125$ / n.s. \\ %
\clk{} & PickXtimes / SwingXtimes (non-priv.) & 41 / 48 & 36 / 48 & $-6/{+}1$, $p{=}.12$ / $-1/{+}1$ \\ %
\clk{} & PickXtimes val (non-priv., pre-reg.) & 38 & 40 & $-3/{+}5$, $p{=}.73$ \\ %
\bottomrule
\end{tabular}}
\caption{\textbf{Cost without null events.} Simulator-contact variant unless stated. Per-task values for the 15 tasks: BinFill 23/22, StopCube 24/23, RouteStick 33/33, VideoPlaceButton 29/28, VideoUnmaskSwap 14/13, ButtonUnmaskSwap 6/9, PatternLock 28/28, MoveCube 39/38, PickHighlight 12/12, InsertPeg 5/3, ButtonUnmask 11/10, VideoPlaceOrder 18/20, VideoRepick 13/13, VideoUnmask 17/17, SwingXtimes 44/48.}
\label{tab:supp-le-cost}
\end{table*}

\paragraph{Residual gap to the hidden reference.}
On the 7 test probes the event index stays 47 episodes below the hidden reference (248 against 295 of 350).
\Cref{tab:supp-le-gap} attributes each of the 47 from the detector's per-frame log, paired with the same episode under the hidden condition (rules hard-coded in the analysis script; counts are net of chance expectation and of reverse pairs, so approximate). %
About a quarter of the gap is false collapse, entirely on PickXtimes and in the same episodes that carry the clean-episode cost; about an eighth is the one-way boundary; slightly more than a quarter is missed null frames; the rest is within run-to-run noise. We did not tune the detector to close this gap. %

\begin{table}[t]
\centering
\small
\setlength{\tabcolsep}{3pt}
\suppfit{\begin{tabular}{lccccc}
\toprule
Probe & Gap & \begin{tabular}[b]{@{}c@{}}False\\collapse\end{tabular} & \begin{tabular}[b]{@{}c@{}}One-way\\boundary\end{tabular} & \begin{tabular}[b]{@{}c@{}}Missed null\\frames\end{tabular} & Noise \\
\midrule
PickXtimes pause 300 & 10 & $\approx$7 & -- & $\approx$0 & $\approx$3 \\
SwingXtimes one-way air-grasp & 8 & -- & $\approx$6 & -- & $\approx$2 \\
SwingXtimes air-grasp-ret & 5 & -- & -- & $\approx$4 & $\approx$1 \\
PickXtimes fidget 100 & 10 & $\approx$4--5 & -- & $\approx$2 & $\approx$3 \\
SwingXtimes fidget 100 & 7 & 0 & -- & $\approx$5 & $\approx$2 \\
RouteStick pause 300 & 5 & 0 & -- & $\approx$2 & $\approx$3 \\
VideoPlaceButton pause 300 & 2 & 0 & -- & 0 & 2 \\
\midrule
Total & 47 & $\approx$11 & $\approx$6 & $\approx$13 & $\approx$17 \\
\bottomrule
\end{tabular}}
\caption{\textbf{Attribution of the gap between the event index and the hidden reference} (released FS-Modul, test).
False collapse: the episode also fails with the event index and no null event (PickXtimes pause: 6 of 12 lost episodes are among the 8 episodes lost on clean episodes with the event index, chance expectation 2.2, hypergeometric $p{=}.003$), or the policy repeats a step.
One-way boundary: the null excursion never returns, so it cannot be collapsed.
Missed null frames: returns that stop 1.3--3 units short (all-or-nothing closure), transition frames at the start of each event, and the last incomplete cycle of a fidget.}
\label{tab:supp-le-gap}
\end{table}

\paragraph{Randomised micro-stops.}
Brief stops at random action-chunk boundaries (probability 0.3 per boundary, 5--40 steps each, until a 300-step budget; on average 7.2 stops and $\approx$157 steps per episode; same schedule in every condition of an episode) harm through memory nearly as much as long pauses: hiding their frames restores success (\cref{tab:supp-le-jitter}). %
The event index recovers little and not significantly.
Each stop begins with a deceleration and settling phase that is neither still nor a closed loop, and the event index keeps the first 5 frames of every stop by design (the grace); summed over many short stops, it keeps 35--50\% of the stop frames (66--71 per episode) and g0 36--39\% (55--60), whereas the hidden condition keeps none. %
This form of null event, the one a deployed robot is most likely to produce, is not solved here.

\begin{wraptable}{r}{0.61\linewidth}
\typeout{WRAPLABEL:tab:supp-le-jitter}%
\centering
\small
\setlength{\tabcolsep}{4pt}
\suppfit{\begin{tabular}{lccccc}
\toprule
Released FS-Modul & No event & Visible & Hidden & + Ev.\ index & g0 \\
\midrule
PickXtimes & 47 & 20 & 46 & 27 & 30 \\ %
SwingXtimes & 44 & 17 & 45 & 24 & 33 \\ %
\bottomrule
\end{tabular}}
\caption{\textbf{Randomised micro-stops}, 50 paired episodes (event index: simulator-contact variant). Visible versus hidden: $p<10^{-4}$ on both tasks. Event index versus visible: $p{=}.09$ and $.14$ (not significant). g0 versus visible on PickXtimes: $p{=}.006$.}
\label{tab:supp-le-jitter}
\end{wraptable}
\paragraph{Summary of the boundaries.}
The event index removes the damage of null events that return the world to a previous state; it partly fails on null events that do not return (one-way air-grasp, randomised micro-stops); it can fail where a real repetition also returns the world to a previous state (PickXtimes, and in demonstrations VideoRepick); and it is wrong by construction where elapsed time is the task (TimedTransfer, \cref{supp:mikasa}).
These limits follow from a state-return criterion, though we observed them before explaining them, and the miss on micro-stops owes as much to the detector's mechanics.

\section{An exploratory study on MIKASA-Robo}
\label{supp:mikasa}
\label{supp:mikasa-robo}

\paragraph{Summary.}
On MIKASA-Robo~\cite{cherepanov2025mikasa} we trained the same architecture (64 memory slots) from the $\pi_{0.5}$ base checkpoint on \emph{BlinkCount}, in which a light blinks $K$ times and the robot must then press a button $K$ times and submit.
The policy solves only 24 of 50 episodes, below the competence level we set in advance, so this evidence is exploratory.
In the demonstrations the cue lasts longer the more blinks it contains, with non-overlapping ranges, so the time at which the cue ends identifies $K$.
This is consistent with reading the cue's duration: with the cue fixed to 50 steps for every $K$, success falls from 24 to 6, and almost all remaining successes are for the one count whose usual cue lasts about that long ($K{=}5$: 5 of 9; other $K$: 1 of 41); seeing the light does not help (9 with the light hidden as well).
Faster time stamps make it over-press (38 of 42 failures), and a visible pause after the cue reduces success to 8, against 23 when hidden, here through the stored frames rather than the stamps.

\paragraph{Model and data.}
We trained FrameSamp-Modul (clock on, 64 memory slots) from the $\pi_{0.5}$ base checkpoint with the recipe of \clk{} (30k steps, batch 64) on 1250 MIKASA-Robo demonstrations (about 183k frames) of BlinkCount (Medium and Hard) and TimedTransfer (Easy, Medium and Hard), and evaluate BlinkCount-Medium ($K\le5$) and TimedTransfer-Medium. %
With 32 slots, uniform time sampling loses blinks (3 steps each): replayed on the real episode lengths, blinks remain countable at the last step in 54\% of Medium episodes, against 100\% with 64. %
The loader does not use the hook affected by the error of \cref{supp:data}. %

\paragraph{Task structure.}
In BlinkCount-Medium the cue lasts a waiting time of 5--10 steps plus $K$ blinks of 3 steps separated by 3--5 dark steps, and ends when a phase light turns green.
The step at which the cue ends identifies $K$: the ranges do not overlap ($K{=}1$: 12--18, $K{=}2$: 20--23, $K{=}3$: 25--31, $K{=}4$: 32--38, $K{=}5$: 39--47). %
The robot must then press a button $K$ times and submit. Failures are \emph{over-press} (a $(K{+}1)$-th press, which ends the episode), \emph{early submit}, or timeout.
In TimedTransfer the robot must move an object while a timer light is in its target window, so reading time is correct.

\paragraph{Pre-registered gate.}
Before any result we required, for interpreting any other cell, BlinkCount no-event success $\ge60\%$ and $\ge30$ points above an uninformative control (blank cue: blink light hidden throughout), and TimedTransfer no-event success $\ge50\%$. %
\textbf{The gate failed}: BlinkCount reached 48\% (24/50), and the blank cue gave 23. %
The blank cue is informative, since its duration still identifies $K$; against a cue padded to 50 steps with the light hidden, the margin is 30 points (24 against 9, $+8/{-}23$, $p{=}.011$), but the first criterion is still not met. %
A retraining with cue duration decoupled from $K$ was pre-registered to repair the gate but never run. %
All MIKASA results are therefore exploratory; each cell added after the gate had a written prediction before it was run. %

\begin{table*}[t]
\centering
\small
\setlength{\tabcolsep}{4pt}
\suppfit{%
\begin{tabular}{lcccccccccc}
\toprule
 & & \multicolumn{3}{c}{Failures} & \multicolumn{5}{c}{Success by $K$} & \\
\cmidrule(lr){3-5}\cmidrule(lr){6-10}
BlinkCount-Medium & Succ. & Over-press & Early & Timeout & 1 & 2 & 3 & 4 & 5 & vs no event \\
\midrule
No event & 24 & 15 & 11 & 0 & 4/8 & 5/7 & 7/10 & 4/16 & 4/9 & \\ %
Light hidden (blank cue), natural duration & 23 & 9 & 18 & 0 & 5/8 & 5/7 & 6/10 & 4/16 & 3/9 & $+5/{-}6$, $p{=}1$ \\ %
\midrule
Cue fixed to 50 steps, light visible & 6 & 36 & 8 & 0 & 0/8 & 0/7 & 1/10 & 0/16 & 5/9 & $+4/{-}22$, $p{=}.0005$ \\ %
Cue fixed to 50 steps, light hidden & 9 & 31 & 10 & 0 & 0/8 & 0/7 & 0/10 & 7/16 & 2/9 & $+8/{-}23$, $p{=}.011$ \\ %
\midrule
Time stamps $\times1.5$ from step 0 & 8 & 38 & 3 & 1 & 3/8 & 1/7 & 1/10 & 0/16 & 3/9 & $+3/{-}19$, $p{<}.001$ \\ %
Time stamps $\times0.67$ from step 0 & 18 & 13 & 19 & 0 & 1/8 & 1/7 & 8/10 & 8/16 & 0/9 & $+10/{-}16$, $p{=}.33$ \\ %
Time stamps $\times1.5$ from step 50 & 15 & 26 & 8 & 1 & 4/8 & 2/7 & 3/10 & 3/16 & 3/9 & $+4/{-}13$, $p{=}.049$ \\ %
Time stamps $\times0.5$ from step 50 & 26 & 9 & 14 & 1 & 4/8 & 5/7 & 5/10 & 10/16 & 2/9 & $+9/{-}7$, $p{=}.80$ \\ %
\midrule
Pause 50 after first press, visible & 8 & 1 & 41 & 0 & 7/8 & 1/7 & 0/10 & 0/16 & 0/9 & $+4/{-}20$, $p{=}.0015$ \\ %
Pause 50, hidden & 23 & 18 & 9 & 0 & 4/8 & 3/7 & 5/10 & 5/16 & 6/9 & $+5/{-}6$, $p{=}1$ \\ %
Pause 50, time only & 19 & 26 & 5 & 0 & 4/8 & 1/7 & 4/10 & 4/16 & 6/9 & $+6/{-}11$, $p{=}.33$ \\ %
Pause 50, content only & 5 & 3 & 42 & 0 & 5/8 & 0/7 & 0/10 & 0/16 & 0/9 & $+3/{-}22$, $p{<}.001$ \\ %
\bottomrule
\end{tabular}}
\caption{\textbf{All BlinkCount-Medium cells} of the 64-slot MIKASA model, 50 paired episodes (same environment seeds and policy noise); vs no event: episodes gained / lost, exact McNemar.
``From step 50'': only stamps of frames after the cue (for every $K$) are rescaled.
Pause: 50 steps immediately after the first counting press; time only: pause frames withheld, stamps advance; content only: pause frames stored, stamps frozen.}
\label{tab:supp-mikasa-blink}
\end{table*}

\paragraph{TimedTransfer, the positive control.}
Without events the policy succeeds in all 50 TimedTransfer-Medium episodes. %
Rescaling the stamps fails every episode in the predicted direction: $\times1.5$ gives 0 (all 50 too early), $\times0.5$ gives 0 (all 50 late). %
A 50-step pause at step 40 gives 35 when visible (15 late, $+0/{-}15$, $p{<}.001$) and 0 when hidden (all 50 late): hiding a pause removes real elapsed time from memory, so this failure holds by construction. %

\paragraph{Reading.}
\begin{itemize}
\item \emph{$K$ is read from the cue duration.} With the cue fixed to 50 steps, success concentrates on $K{=}5$, whose natural cue end (39--47) is closest to 50 (5 of 9, against 1 of 41 for other $K$), and failures are over-presses (36). Hiding the light as well gives a similar result (9), though it changes which $K$ succeed ($K{=}4$: 0/16 visible, 7/16 hidden; $K{=}5$: 5/9 and 2/9). %
\item \emph{The duration is read from the time stamps.} Rescaled from step 0, they shift the inferred count both ways: over-presses rise from 15 to 38 at $\times1.5$ ($p{<}.001$); early submissions rise from 11 to 19 at $\times0.67$, with no significant change in success ($p{=}.33$) because the shift also corrects over-counted episodes ($K{=}3,4$ improve, $K{=}1,2,5$ drop). %
\item \emph{Pre-registered miss (H1).} We predicted that stamps rescaled after the cue would cause early submissions at $\times1.5$ and extra presses at $\times0.5$. Instead, $\times1.5$ from step 50 costs 9 episodes ($p{=}.049$) through over-presses, and $\times0.5$ has no effect (26, $p{=}.80$). %
\item \emph{The pause acts through memory content.} The pre-registered hypothesis H2 (visible pause $\ge10$ points below no event with mostly early submissions, hidden within 6) held: a visible pause after the first press causes early submissions (41 of 42 failures), a hidden one does not (23). Under pre-registered split thresholds ($\le14$ with mostly early submissions: route used; $\ge18$: not used), content only reproduces the damage (5, 42 early) and time only does not (19, mostly over-presses), as for \nope{} (\cref{sec:noclock}). %
\end{itemize}
Run-to-run noise: a duplicate no-event evaluation differs from no event by $+7/{-}4$. %

\paragraph{What MIKASA does and does not show.}
On a second benchmark, with a different action space and demonstration source, a time-indexed memory reads a task variable ($K$) from time when time predicts it in the demonstrations, and the stamp manipulation acts on time (TimedTransfer).
It does not show that a competent BlinkCount policy is vulnerable: base success is 48\%, the gate failed, and the decoupled retraining was not run.

\section{Pre-registered predictions and their outcomes}
\label{supp:prereg}
\begin{table}[t]
\centering\small\setlength{\tabcolsep}{2pt}
\suppfit{\begin{tabular}{llccccc}
\toprule
Model / task & Condition & $K{=}1$ & 2 & 3 & 4 & 5 \\
\midrule
Released FS-Modul, Pick & no event & 16/16 & 10/10 & 11/12 & 6/6 & 4/6 \\
Released FS-Modul, Pick & pause & 16/16 & 0/10 & 0/12 & 0/6 & 0/6 \\
Released FS-Modul, Pick & pause, hidden & 16/16 & 10/10 & 11/12 & 5/6 & 3/6 \\
Released FS-Modul, Pick & fidget & 16/16 & 0/10 & 0/12 & 0/6 & 0/6 \\
Released FS-Modul, Pick & air-grasp & 2/16 & 10/10 & 0/12 & 0/6 & 0/6 \\
Released FS-Modul, Swing & no event & 14/15 & 10/12 & 20/23 & -- & -- \\
Released FS-Modul, Swing & pause & 10/15 & 0/12 & 0/23 & -- & -- \\
Released FS-Modul, Swing & pause, hidden & 13/15 & 12/12 & 22/23 & -- & -- \\
Released FS-Modul, Swing & fidget & 10/15 & 0/12 & 0/23 & -- & -- \\
Released FS-Modul, Swing & air-grasp & 13/15 & 4/12 & 6/23 & -- & -- \\
\clk{} (from scratch), Pick & no event & 16/16 & 9/10 & 12/12 & 2/6 & 2/6 \\
\clk{} (from scratch), Pick & pause & 16/16 & 1/10 & 0/12 & 0/6 & 0/6 \\
\clk{} (from scratch), Pick & pause, hidden & 16/16 & 9/10 & 9/12 & 3/6 & 1/6 \\
\clk{} (from scratch), Pick & fidget & 15/16 & 1/10 & 0/12 & 0/6 & 0/6 \\
\clk{} (from scratch), Swing & no event & 15/15 & 12/12 & 21/23 & -- & -- \\
\clk{} (from scratch), Swing & pause & 10/15 & 0/12 & 0/23 & -- & -- \\
\clk{} (from scratch), Swing & pause, hidden & 11/15 & 12/12 & 15/23 & -- & -- \\
\clk{} (from scratch), Swing & air-grasp & 13/15 & 0/12 & 0/23 & -- & -- \\
\nope{} (from scratch), Pick & no event & 16/16 & 10/10 & 11/12 & 3/6 & 2/6 \\
\nope{} (from scratch), Pick & pause & 16/16 & 3/10 & 1/12 & 1/6 & 0/6 \\
\nope{} (from scratch), Pick & pause, hidden & 16/16 & 9/10 & 12/12 & 5/6 & 2/6 \\
\nope{} (from scratch), Swing & no event & 15/15 & 12/12 & 20/23 & -- & -- \\
\nope{} (from scratch), Swing & pause & 9/15 & 4/12 & 1/23 & -- & -- \\
\nope{} (from scratch), Swing & pause, hidden & 14/15 & 12/12 & 20/23 & -- & -- \\
\bottomrule
\end{tabular}
}
\caption{\textbf{Success by the required number of repetitions $K$} (successes / episodes with that $K$). SwingXtimes has $K\leq3$.}
\label{supp:tab:byK}
\end{table}

For the key tests we recorded the prediction and the decision rule, including thresholds, in a dated log before the run.
Below are the predictions that bear on main-text claims, with their outcomes, including those that missed.
\begin{itemize}\itemsep0pt
\item \nope{} (registered before training): manipulation check, rescaling its (absent) stamps changes nothing (44 against 42, met); pauses still hurt by at least 8 episodes on PickXtimes (21 against 42, met), with mostly early failures (21 of 29, met).
\item SimpleMemVLA: history stretched 1.5 times adds at least 5 early stops and compressed 0.5 times adds at least 3 extra repetitions (met on test and on validation episodes); visible pauses cost at least 5 episodes against hidden (\emph{not} met: 47 against 50); visible fidgets do (met: 38 against 50).
\item SimpleMemVLA on SwingXtimes (\cref{tab:supp_smv_swing}): fidgets visible $\leq$ hidden $-6$, mostly early (met: 33 against 49, 11 of 17 early); pauses less harmful than fidgets (met: $-12$ against $-16$); history $\times1.5$ mostly early and $\times0.5$ mostly extra (half met: $\times0.5$ 42 against 49, 5 of 8 extra; $\times1.5$ no loss).
\item Event index, held-out replication on new environment seeds: gains on the seven probes (met in 6 of 7; the one-way air-grasp was not significant); on \clk{} under pauses, a gain of at least 8 on each task (met: 10 to 36, 14 to 32).
\item Pause-only augmentation (\cref{supp:pauseonly}): pauses fixed with no-event success kept (\emph{not} met on either task); second-machine control and held-out comparison, rules written before each run (met, PickXtimes only on held-out episodes). Weaker dose: counting kept (met on SwingXtimes, not on PickXtimes held-out, 31, below the threshold of 39; retrain 45); pauses improved (\emph{not} met on either task).
\item MIKASA route split: if the stamp route carried the pause effect, time only would fall below 18 (it did not: 19); content only reproduced the effect (5).
\item MIKASA: the pre-registered gate that the policy reads the light failed (\cref{supp:mikasa}).
\item Linear probe of the memory tokens (\cref{supp:noclock:probe}): time readable from \nope{} (met on both tasks); pause inflation met on PickXtimes, not on SwingXtimes (elapsed shift 27\% against 50\%).
\item Null-slot dose on \nope{} (\cref{supp:noclock:dose}): occupancy dose rather than one block read as one event (met on both tasks: $m{=}6$ against $m{=}1$, $-21$ and $-24$).
\item Second training seed of \clk{} and \nope{} (\cref{supp:seed}): five rules, all met.
\item One pause on \nope{} (\cref{supp:noclock:press}): one 5-second pause alone reads as about one repetition (met: 10 of 10 early presses after two swings at $K{=}3$, SwingXtimes). $2{\times}300$ on held-out episodes, length-independent or duration-carrying: neither met on either task (7 of 13 and 6 of 9 after two repetitions); open.
\item Press timing over pause length on \clk{} (\cref{tab:supp-dose}): partial (clock side at $L{=}30$, skip-one ratios at $L{=}300$); the repetition count, not registered, favours the clock (34 of 36 at most one repetition, both tasks); the held-out replication with that criterion, registered, gives clock (29 of 29).
\item Fidgets on four further released designs (\cref{tab:supp_designs}), PickXtimes: FrameSamp-Context falls to at most 18 (\emph{not} met: 20 against 33, $p{=}.004$, all 30 failures early presses); TTT, $\pi_{0.5}$ and text sub-goals stay within 6 of their success without events (met: 25 against 23, 20 against 18, 47 against 46).
\item Seven further RoboMME tasks (\cref{tab:supp_tasks}): visible $<$ hidden where information is shown briefly, shrinking with video length (\emph{not} met: VideoUnmask $-3$, VideoPlaceOrder $-6$, both n.s.); StopCube hidden $\leq$ visible (met at floor, 0 against 2; not interpreted).
\end{itemize}

\section{Every cell of the main text}
\label{supp:cells}
\begin{table*}[t]\centering\small\setlength{\tabcolsep}{2.5pt}
\suppfit{\begin{tabular}{llcccccccc}\toprule
Task & Condition & Succ. & 95\% CI & vs ref. & $p$ & $p_\text{Holm}$ & Early & Extra & Other/Uncl. \\ \midrule
Pick & no event & 47/50 & 84--98 & -- & -- & -- & 0 & 0 & 0/3 \\
Pick & pause 1.5\,s & 24/50 & 35--61 & $-$23/+0 & $<10^{-6}$ & $<10^{-5}$ & 0 & 0 & 0/26 \\
Pick & pause$^\ast$ & 16/50 & 21--46 & $-$31/+0 & $<10^{-9}$ & $<10^{-7}$ & 34 & 0 & 0 \\
Pick & pause, hidden & 45/50 & 79--96 & $-$3/+1 & 0.62 & 1 & 5 & 0 & 0 \\
Pick & pause 15\,s & 16/50 & 21--46 & $-$31/+0 & $<10^{-9}$ & $<10^{-7}$ & 0 & 0 & 0/34 \\
Pick & pause 15\,s, hidden & 44/50 & 76--94 & $-$4/+1 & 0.38 & 1 & 0 & 0 & 0/6 \\
Pick & fidget & 16/50 & 21--46 & $-$31/+0 & $<10^{-9}$ & $<10^{-7}$ & 34 & 0 & 0 \\
Pick & fidget, hidden & 46/50 & 81--97 & $-$2/+1 & 1 & 1 & 4 & 0 & 0 \\
Pick & air-grasp & 12/50 & 14--37 & $-$35/+0 & $<10^{-9}$ & $<10^{-8}$ & 0 & 0 & 0/38 \\
Pick & air-grasp, hidden & 43/50 & 74--93 & $-$5/+1 & 0.22 & 1 & 7 & 0 & 0 \\
Pick & stamps $\times0.5$ & 24/50 & 35--61 & $-$23/+0 & $<10^{-6}$ & $<10^{-5}$ & 3 & 23 & 0 \\
Pick & stamps $\times1.5$ & 18/50 & 24--50 & $-$29/+0 & $<10^{-8}$ & $<10^{-6}$ & 32 & 0 & 0 \\
Swing & no event & 44/50 & 76--94 & -- & -- & -- & 0 & 0 & 0/6 \\
Swing & pause 1.5\,s & 21/50 & 29--56 & $-$25/+2 & $<10^{-5}$ & $<10^{-3}$ & 0 & 0 & 0/29 \\
Swing & pause$^\ast$ & 10/50 & 11--33 & $-$34/+0 & $<10^{-9}$ & $<10^{-8}$ & 36 & 0 & 4 \\
Swing & pause, hidden & 47/50 & 84--98 & $-$2/+5 & 0.45 & 1 & 1 & 1 & 1 \\
Swing & pause 15\,s & 12/50 & 14--37 & $-$32/+0 & $<10^{-9}$ & $<10^{-7}$ & 0 & 0 & 0/38 \\
Swing & pause 15\,s, hidden & 46/50 & 81--97 & $-$2/+4 & 0.69 & 1 & 0 & 0 & 0/4 \\
Swing & fidget & 10/50 & 11--33 & $-$34/+0 & $<10^{-9}$ & $<10^{-8}$ & 35 & 0 & 5 \\
Swing & fidget, hidden & 45/50 & 79--96 & $-$2/+3 & 1 & 1 & 1 & 1 & 3 \\
Swing & air-grasp & 23/50 & 33--60 & $-$22/+1 & $<10^{-5}$ & $<10^{-3}$ & 0 & 0 & 0/27 \\
Swing & air-grasp, hidden & 47/50 & 84--98 & $-$0/+3 & 0.25 & 1 & 0 & 0 & 0/3 \\
Swing & stamps $\times0.5$ & 20/50 & 28--54 & $-$25/+1 & $<10^{-6}$ & $<10^{-4}$ & 0 & 0 & 0/30 \\
Swing & stamps $\times1.5$ & 37/50 & 60--84 & $-$10/+3 & 0.092 & 1 & 13 & 0 & 0 \\
\bottomrule\end{tabular}}
\caption{\textbf{Released FrameSamp-Modul}: every cell (50 paired episodes). 95\% CI: Wilson, in \%. vs ref.: episodes lost/gained relative to no event. $p$: exact McNemar; $p_\text{Holm}$: Holm-corrected over all comparisons in this section. Early/Extra: failures before the $K$-th repetition / after it; Uncl.: runs that did not log the stage. $^\ast$Classes from the stage-logged rerun, which reproduces the cell episode by episode (\cref{supp:robust}).}
\label{supp:tab:cells_released}\end{table*}

\begin{table*}[t]\centering\small\setlength{\tabcolsep}{2.5pt}
\suppfit{\begin{tabular}{llcccccccc}\toprule
Task & Condition & Succ. & 95\% CI & vs ref. & $p$ & $p_\text{Holm}$ & Early & Extra & Other/Uncl. \\ \midrule
Pick & no event & 41/50 & 69--90 & -- & -- & -- & 9 & 0 & 0 \\
Pick & pause & 17/50 & 22--48 & $-$24/+0 & $<10^{-6}$ & $<10^{-5}$ & 33 & 0 & 0 \\
Pick & pause, hidden & 38/50 & 63--86 & $-$4/+1 & 0.38 & 1 & 12 & 0 & 0 \\
Pick & fidget & 16/50 & 21--46 & $-$25/+0 & $<10^{-7}$ & $<10^{-5}$ & 32 & 1 & 1 \\
Pick & stamps $\times1.5$ & 19/50 & 26--52 & $-$22/+0 & $<10^{-6}$ & $<10^{-4}$ & 28 & 2 & 1 \\
Swing & no event & 48/50 & 87--99 & -- & -- & -- & 0 & 0 & 2 \\
Swing & pause & 10/50 & 11--33 & $-$38/+0 & $<10^{-9}$ & $<10^{-9}$ & 37 & 0 & 3 \\
Swing & pause, hidden & 38/50 & 63--86 & $-$10/+0 & 0.002 & 0.031 & 3 & 0 & 9 \\
Swing & air-grasp & 13/50 & 16--40 & $-$35/+0 & $<10^{-9}$ & $<10^{-8}$ & 37 & 0 & 0 \\
\bottomrule\end{tabular}}
\caption{\textbf{\clk{} (from scratch)}: every cell (50 paired episodes). 95\% CI: Wilson, in \%. vs ref.: episodes lost/gained relative to no event. $p$: exact McNemar; $p_\text{Holm}$: Holm-corrected over all comparisons in this section. Early/Extra: failures before the $K$-th repetition / after it; Uncl.: runs that did not log the stage.}
\label{supp:tab:cells_clock}\end{table*}

\begin{table*}[t]\centering\small\setlength{\tabcolsep}{2.5pt}
\suppfit{\begin{tabular}{llcccccccc}\toprule
Task & Condition & Succ. & 95\% CI & vs ref. & $p$ & $p_\text{Holm}$ & Early & Extra & Other/Uncl. \\ \midrule
Pick & no event + event index & 36/50 & 58--83 & $-$6/+1 & 0.12 & 1 & 10 & 4 & 0 \\
Pick & pause + event index & 34/50 & 54--79 & $-$1/+18 & $<10^{-4}$ & 0.0014 & 12 & 3 & 1 \\
Pick & fidget + event index & 33/50 & 52--78 & $-$2/+19 & $<10^{-3}$ & 0.0038 & 12 & 5 & 0 \\
Swing & no event + event index & 48/50 & 87--99 & $-$1/+1 & 1 & 1 & 0 & 0 & 2 \\
Swing & pause + event index & 30/50 & 46--72 & $-$3/+23 & $<10^{-4}$ & 0.0016 & 13 & 0 & 7 \\
Swing & air-grasp + event index & 43/50 & 74--93 & $-$0/+30 & $<10^{-8}$ & $<10^{-7}$ & 5 & 0 & 2 \\
\bottomrule\end{tabular}}
\caption{\textbf{\clk{} (from scratch) + event index (test time)}: every cell (50 paired episodes). 95\% CI: Wilson, in \%. vs ref.: episodes lost/gained relative to the same event without the event index. $p$: exact McNemar; $p_\text{Holm}$: Holm-corrected over all comparisons in this section. Early/Extra: failures before the $K$-th repetition / after it; Uncl.: runs that did not log the stage.}
\label{supp:tab:cells_clock_le}\end{table*}

\begin{table*}[t]\centering\small\setlength{\tabcolsep}{2.5pt}
\suppfit{\begin{tabular}{llcccccccc}\toprule
Task & Condition & Succ. & 95\% CI & vs ref. & $p$ & $p_\text{Holm}$ & Early & Extra & Other/Uncl. \\ \midrule
Pick & no event & 42/50 & 71--92 & -- & -- & -- & 5 & 3 & 0 \\
Pick & pause & 21/50 & 29--56 & $-$21/+0 & $<10^{-6}$ & $<10^{-4}$ & 21 & 8 & 0 \\
Pick & pause, hidden & 44/50 & 76--94 & $-$2/+4 & 0.69 & 1 & 4 & 2 & 0 \\
Pick & pause, null frames swapped & 17/50 & 22--48 & $-$25/+0 & $<10^{-7}$ & $<10^{-5}$ & 21 & 12 & 0 \\
Pick & stamps $\times1.5$ & 44/50 & 76--94 & $-$1/+3 & 0.62 & 1 & 3 & 3 & 0 \\
Swing & no event & 47/50 & 84--98 & -- & -- & -- & 1 & 1 & 1 \\
Swing & pause & 14/50 & 17--42 & $-$33/+0 & $<10^{-9}$ & $<10^{-8}$ & 26 & 0 & 10 \\
Swing & pause, hidden & 46/50 & 81--97 & $-$4/+3 & 1 & 1 & 1 & 0 & 3 \\
Swing & pause, null frames swapped & 13/50 & 16--40 & $-$34/+0 & $<10^{-9}$ & $<10^{-8}$ & 32 & 0 & 5 \\
\bottomrule\end{tabular}}
\caption{\textbf{\nope{} (from scratch)}: every cell (50 paired episodes). 95\% CI: Wilson, in \%. vs ref.: episodes lost/gained relative to no event. $p$: exact McNemar; $p_\text{Holm}$: Holm-corrected over all comparisons in this section. Early/Extra: failures before the $K$-th repetition / after it; Uncl.: runs that did not log the stage.}
\label{supp:tab:cells_nope}\end{table*}

\Cref{supp:tab:cells_released,supp:tab:cells_clock,supp:tab:cells_clock_le,supp:tab:cells_nope} list every cell of the four models analysed throughout, one per model; other designs and tasks are in their own sections. They give Wilson 95\% intervals, paired changes, exact McNemar $p$-values with Holm correction over the 42 comparisons, and failure classes (\cref{supp:stats}); \cref{supp:tab:byK} splits the same cells by $K$.

\paragraph{Success by $K$.}
On PickXtimes, episodes with $K=1$ are almost untouched by pauses and fidgets in every model (16 of 16 under pauses), where there is no decision to stop early; on SwingXtimes pauses do cost some $K=1$ episodes (10, 10 and 9 of 15 for the released model, \clk{} and \nope{}), for reasons we have not resolved. For $K\geq2$ success collapses (released model, PickXtimes: 0 of 34 under pauses).
The PickXtimes air-grasp cell does not follow this pattern (2 of 16 at $K=1$): the returning gripper often re-grasps the cube it has just released, so the event is not null for this task, and we do not use the cell (\cref{supp:robust}).

\end{document}